%% file: survey.tex
\documentclass{article} % For LaTeX2e
\usepackage{iclr2024_conference,times}

\input{math_commands.tex}

\usepackage{hyperref}
\usepackage{url}
\usepackage{csquotes}
\usepackage{placeins}

\usepackage{amsfonts}
\usepackage{amsmath}
\usepackage{amssymb}
\usepackage{amsthm}

\usepackage{cleveref} % for cref command
\usepackage{graphicx}
\usepackage{tcolorbox}

\usepackage{tikz}
\usepackage{booktabs}    % For better table lines
\usepackage{array}       % For improved table formatting
\usepackage{geometry}    % To adjust page margins if needed
\usepackage{caption}     % For better caption control
\usepackage{multirow}    % If multi-row cells are needed in future
\usepackage{adjustbox}   % To scale the table if needed
\usepackage{tabularx}
\usepackage{lscape}
\usetikzlibrary{trees,arrows.meta}

\newcommand{\elliot}[1]{\textcolor{olive}{Elliot: #1}}

\definecolor{lightblue}{RGB}{98, 163, 217}
\definecolor{lightred}{RGB}{227, 89, 110}

\newtcolorbox{takeaways}[1][]{
  colback=yellow!15,
  colframe=lightred,
  boxrule=2pt,
  title=\textbf{Takeaways},
  #1 % This allows you to pass optional arguments
}

\newtcolorbox{openquestions}[1][]{
  colback=yellow!15,
  colframe=lightblue,
  boxrule=2pt,
  title=\textbf{Open Questions},
  #1 % This allows you to pass optional arguments
}

\title{Surveying the Effects of Quality, Diversity, and Complexity in Synthetic Data From Large Language Models}

\author{Alex Havrilla$^1$, Andrew Dai$^{4,5}$, Laura O'Mahony$^3$, Koen Oostermeijer$^4$, Vera Zisler$^4$ \AND Alon Albalak$^6$, Fabrizio Milo$^7$, Sharath Chandra Raparthy$^8$, Kanishk Gandhi$^9$, Baber Abbasi$^{10}$
\AND Duy Phung$^6$, Maia Iyer$^{11}$, Dakota Mahan$^6$, Chase Blagden$^6$, Srishti Gureja$^{12}$, Mohammed Hamdy$^{12}$ \AND
Wen-Ding Li$^2$, Giovanni Paolini$^{13}$, Pawan Sasanka Ammanamanchi$^{7}$, Elliot Meyerson$^{14}$}

\newcommand{\commentout}[1]{}

\iclrfinalcopy % Uncomment for camera-ready version, but NOT for submission.
\begin{document}

% Fabrizio comments: make a grid/taxonomy of synthetic data generation methods
% how does cost factor in as a function of the number of inference steps

% Other comments: dive deeper on some aspects of synthetic data generation. 
% how to encourage more exploration/diversity by combining with exterior

\maketitle

\vspace{-0.5cm}

\begin{center}
    Georgia Tech$^1$, Cornell University$^2$, University of Limerick$^3$, Aleph Alpha @ IPAI$^4$, Sakana AI$^5$, SynthLabs$^6$, Independent$^{7}$, Reka AI$^8$, Stanford University$^9$, Eleuther AI$^{10}$, IBM$^{11}$, Cohere for AI Community$^{12}$, University of Bologna$^{13}$, Cognizant AI Labs$^{14}$
\end{center}

\vspace{0.5cm}

\begin{abstract}
    Synthetic data generation with Large Language Models (\textbf{LLMs}) has emerged as a promising paradigm for augmenting natural data over a nearly infinite range of tasks. However, most existing methods are fairly ad-hoc, utilizing a wide range of seed-datasets, LLMs, prompts, filters, and task-specific generation strategies. Given this variety, direct comparisons among synthetic data generation algorithms are scarce, making it difficult to understand where improvement comes from and what bottlenecks exist. To address this, we propose to evaluate algorithms via the makeup of synthetic data generated by each algorithm. In particular, we propose to examine the \textit{quality}, \textit{diversity}, and \textit{complexity} (\textbf{QDC}) of resulting synthetic data. We choose these three data characteristics due to their significance in open-ended processes and the impact each has on the capabilities of downstream models. We find quality  to be essential for \textit{in-distribution} model generalization, diversity to be essential for \textit{out-of-distribution} generalization, and complexity to be beneficial for both. Further, we emphasize the existence of Quality-Diversity trade-offs in training data and the downstream effects on model performance. We then examine the effect of various components in the synthetic data pipeline on each data characteristic. This examination allows us to taxonomize and compare synthetic data generation algorithms through the components they utilize and the resulting effects on data QDC composition. This analysis extends into a discussion on the importance of balancing QDC in synthetic data for efficient reinforcement learning and self-improvement algorithms. Analogous to the QD trade-offs in training data, often there exist trade-offs between model output quality and output diversity which impact the composition of synthetic data. We observe that many models are currently evaluated and optimized only for output quality, thereby limiting output diversity and the potential for self-improvement. We argue that balancing these trade-offs is essential to the development of future self-improvement algorithms and highlight a number of works making progress in this direction. 
\end{abstract}

\tableofcontents
\newpage

\input{sections/intro}

\input{sections/qdc_measures}

\input{sections/qdc_effects}

\input{sections/qdc_algos}

\input{sections/alternative_tasks}

\input{sections/conclusion}

\bibliography{manual,auto}
\bibliographystyle{iclr2024_conference}

\input{sections/appendix}

\end{document}

%% file: math_commands.tex
\usepackage{amsmath,amsfonts,bm}

\def\eqref#1{equation~\ref{#1}}
\def\1{\bm{1}}

\DeclareMathAlphabet{\mathsfit}{\encodingdefault}{\sfdefault}{m}{sl}
\SetMathAlphabet{\mathsfit}{bold}{\encodingdefault}{\sfdefault}{bx}{n}

\newcommand{\R}{\mathbb{R}}

%% file: sections/intro.tex
\section{Introduction}

% Laura posted this for taxonomy/contents overview visual aid: https://openreview.net/pdf?id=8C5zt-0Utdn

Synthetic data generation has emerged as a promising approach to enhance the capabilities of large language models beyond traditional supervised fine-tuning datasets.
% Alon - this is a very interesting citation, but probably makes more sense to include farther down
% \hamdy{It can also be used as an intervention to personalize or reduce undesired behaviors in language models \citep{wei2024simplesyntheticdatareduces}.}
This development has led to the creation of a diverse set of synthetic data generation algorithms for a variety of tasks and domains. The majority of these algorithms follow a two-step process: First, leverage existing large language models to gather a large set of task prompts and sample continuations. Second, filter the generated dataset to eliminate ``low-quality'' samples.
Their main goal is to maximize the ``quality'' and quantity of synthetically generated data. Relatively little effort is spent seeking to carefully understand what intrinsic characteristics of the data most impact downstream generalization. While these algorithms are a natural place to start, this type of approach is inefficient, leading to most synthetically generated data being discarded \citep{zhou2023limaalignment}. 

%\alexh{Complexity is a third essential characteristic of data that intuitively measures some notion of the ``difficulty'' or ``compositionality'' of a sample. ... This notion of complex systems could be important for future developments in dataset creation: learning increasingly complex knowledge is important for the development of more powerful models, and the evolution of more complex data is important for models to be able to have more learning opportunities in the first place.}

This survey aims to clarify the impact of synthetic data generation on downstream model generalization by analyzing three key data characteristics: \textit{quality}, \textit{diversity}, and \textit{complexity}. Informally, quality measures the ``noisiness'', ``correctness'', or ``alignment'' of data with a desired target distribution $Q$. Diversity measures the ``self-similarity'' or ``coverage'' of data. Complexity intuitively captures some notion of the ``difficulty'' or ``compositionality'' of data. We choose these characteristics for their importance so far in assessing and building artificial open-ended systems, an emerging paradigm that can be applied to iterative self-improvement of models \citep{hughes2024openendedness}. The field of Quality-Diversity has established quality and diversity measures as effective proxies in encouraging increasingly novel and interesting/learnable/valuable synthetic artifacts, often of increasing complexity, and synthetic data generation is natural application of this framework \citep{pugh2016quality,cully2017quality,chatzilygeroudis2021quality}. The importance of data quality, diversity, and complexity is also reflected in many prominent synthetic data generation methods, which explicitly or implicitly aim to maximize at least one of the above (though rarely all three together) \citep{xu2023wizardlmempoweringlargelanguage,gunasekar2023textbooks,wang2023selfinstructaligninglanguagemodels}.

Through this quality-diversity-complexity (\textbf{QDC}) lens we investigate three closely related research questions:

\begin{itemize}
    \item \label{q:rq1}\textbf{RQ1:} How should quality, diversity, and complexity be defined? How are these quantities measured in LLM literature?
    \item \label{q:rq2}\textbf{RQ2:} How do quality, diversity, and complexity in training data impact model generalization?
    \item \label{q:rq3}\textbf{RQ3:} How do existing synthetic data generation algorithms promote quality, diversity, and complexity?
\end{itemize} The answers to these questions can enable the design of more sample-efficient synthetic data generation algorithms with better model generalization and self-improvement abilities.

In Section \ref{sec:qdc_measures} we investigate \textbf{RQ1}. We begin by providing abstract, high-level definitions of quality, diversity and complexity in data. Informally, each characteristic is fairly intuitive: quality measures the ``noisiness'' or ``correctness'' of data, diversity measures the ``coverage'' and ``self-similarity'' of data, and complexity measures the ``difficulty'' or ``compositionality'' of data. Yet, despite these intuitive informal definitions, many different practical measures for each characteristic exist in the literature, and these practical measures vary in their utility. Some are generally applicable, others domain-specific. Some correlate with downstream metrics of interest, while others do not (depending on the task).
 
%After presenting a taxonomy for each set of practical measures, we discuss those that are most useful in practice. While there is not a single best choice for QDC measurement, in general we find the best performing measures are also the most domain-specific. Often these methods are implemented by prompting LLMs with domain/problem specific structure.

Armed with a better understanding of how data quality, diversity, and complexity are measured in practice, in Section \ref{sec:qd_effects} we survey the effects of each characteristic on model performance. We come away with three key takeaways in answer to \textbf{RQ2}: 
\begin{itemize}
    \item Data quality is essential for \textit{in-distribution} generalization.
    \item Data diversity is essential for \textit{out-of-distribution} (\textbf{OOD}) generalization.
    \item Appropriate levels of data complexity benefit both in-distribution and OOD generalization.
\end{itemize} Further, trade-offs often arise between the quality and diversity of training data. In such situations, decisions must be made on how to prioritize the three characteristics. 
This gives rise to a potential QDC \textit{generalization frontier} as different mixtures of quality, diversity, and complexity change how downstream models generalize.

%\alon{Is the ``QDC generalization frontier'' a candidate for a conceptual diagram?}

\begin{figure*}[t]
\hspace{-0.8cm}
\includegraphics[scale=0.75,keepaspectratio]{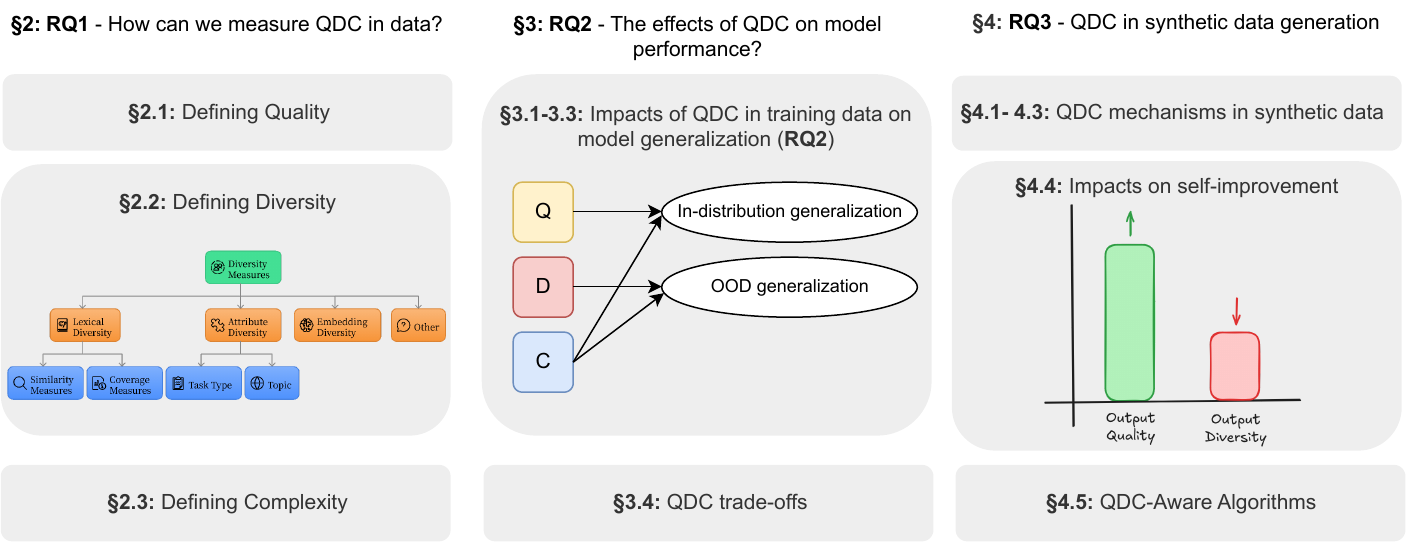}  
\caption{A summary of our research questions, and key findings discussed in greater detail in the relevant sections and subsections.}
\label{fig:overview}
\vspace{-10pt}
\end{figure*}
% %\laura{\href{https://drive.google.com/file/d/1B5_tzII5YzCP4h5JZPIjqPQ9RdHfMoZK/view?usp=sharing}{Link to edit draw.io diagram}}
% \alexh{remove weird vertical spacing}
% %TODO: add caveat in data complexity explaning that the model needs to be able to learn the complexity (e.g. not something totally random)

Finally, we move to investigate \textbf{RQ3} in Section \ref{sec:qd_algo}. We begin by taxonomizing existing synthetic data generation approaches through the QDC perspective. This is done by classifying common components of synthetic data pipelines as ``quality-promoting'', ``diversity-promoting'', or ``complexity-promoting''. This results in a continuum of methods which mix and match various components, resulting in synthetic data with varying degrees of quality, diversity, and complexity. We find the majority of algorithms employ relatively simple methods of promoting quality, often by sampling from a large SOTA model. Similarly, many methods promote diversity simply by initializing sampling using a large seed-dataset. Often complexity is not explicitly considered at all. We then discuss the \textbf{impact of QDC data characteristics on the synthetic data generation process itself} with applications to model self-improvement. Analogous to the QDC trade-offs found in Section \ref{sec:qd_effects}, we find several works suggesting a trade-off between models generating high-quality data and models generating highly diverse data, i.e., a trade-off between model output quality and model output diversity. We argue that, due to this trade-off, future algorithms for synthetic data generation must carefully balance QDC mixtures of synthetic training data for optimal self-improvement. However, the majority of algorithms and benchmarks today optimize for quality alone. As a result, model output diversity and the potential for bigger self-improvement gains suffer. Finally, we highlight a few approaches directly inspired by more classical quality diversity (QD) search algorithms \citep{lehman2011evolving,mouret2015illuminatingsearchspacesmapping} (cf. QD paragraph in \cref{sec:related_surveys}), which attempt to more explicitly control the quality and diversity of generated data.
These \textit{QD synthetic data generation algorithms} explicitly aim to generate data which has both maximal quality and diversity in a sample efficient way, thus receiving the benefits of both characteristics.

In Section \ref{sec:alternative_data_generation} we survey evolutionary/quality-diversity algorithms for synthetic data generation with LLMs outside of common benchmarks tasks. We then conclude the survey in Section \ref{sec:conclusions} by reviewing the key takeaways highlighted at the end of earlier sections. Notable takeaways include:

\vspace{0.3cm}
\begin{takeaways}
\setlength{\leftmargini}{0pt}
    \begin{itemize}
    \item Quality largely improves in-distribution generalization and diversity largely improves OOD generalization. Appropriate levels of complexity can improve both.
    \item Quality and diversity often trade-off in training data.
    \item Many existing models/methods are heavily optimized and evaluated for model output quality, thereby limiting synthetic data diversity.
\end{itemize}
\end{takeaways} We also summarize the list of collected open problems highlighted at the end of earlier sections. Notable open questions include:

\vspace{0.3cm}
\begin{openquestions}
\setlength{\leftmargini}{0pt}
    \begin{itemize}
    \item Establishing benchmarks jointly measuring the quality \textit{and diversity} of model output and synthetic data.
    \item Designing better algorithms explicitly controlling for trade-offs between model output quality and output diversity.
    \item Better understanding of the trade-offs between complexity and the other two characteristics.
\end{itemize}
\end{openquestions} See Figure \ref{fig:overview} for a visual outline of the survey's organization and key-takeaways.

% TODO: more explicitly state in paper that quality and diversity and optimizing for two different background distributions

% TODO: possibly change generalization -> performance. Generalization implies OOD in many people's mind

%\andrew{Would a single statement analogy fit here to further motivate the potential of synthetic data (e.g. humans can master skills in mathematics by synthesizing diverse variations in ``practice" problems (== synthetic data) as novel but learnable opportunities to plug the gaps in their learning that would otherwise be limited by the number of learning from a fixed set of textbooks...?)}

\subsection{Related Topics and Surveys}
\label{sec:related_surveys}

\textbf{Synthetic Data Generation}\quad Synthetic data generation algorithms utilize generative models to create ``synthetic'' data points which can be used downstream for training, benchmarking, etc. A few recent surveys have investigated synthetic data generation \citep{bauer2024comprehensiveexplorationsyntheticdata, guo2024generativeaisyntheticdata, liu2024bestpracticeslessonslearned,long2024llmsdrivensyntheticdatageneration}. \citet{bauer2024comprehensiveexplorationsyntheticdata} provide a broad overview of synthetic data generation across both vision and language throughout the past decade. Specifically, they highlight the difficulty of benchmarking existing algorithms. \citet{guo2024generativeaisyntheticdata} and \citet{liu2024bestpracticeslessonslearned} focus their surveys on synthetic data generation practices that have developed more recently, with a primary spotlight on LLMs. Discussion is centered around applications in various domains (e.g., reasoning and multi-modality). Less emphasis is put on comparing characteristics of the data generated by different algorithms in the same domain. \citet{long2024llmsdrivensyntheticdatageneration} look at LLM-driven synthetic data generation, curation, and evaluation of synthetic data without as much emphasis on downstream impacts.
%\elliot{How are these existing surveys limited?}

%\andrew{Once the survey is near-completion, and we get a sense of the state of the literature, it would be very nice to motivate the importance of this survey by identifying which gap we fill (i.e. a survey to understand the unified relationship between quality and diversity on training outcomes (section 2), and techniques highlighting generation that can lead to both high-quality and diverse data (or varying mixtures of both quantities). A very short related survey works paragraph could do, to reinforce the value of the takeaways from the story already presented in the intro and contents.}

\textbf{Data Selection}\quad Data selection is the task of selecting a susbet of desirable training samples from a larger training dataset $\mathcal{D}$. It plays an important role in a number of synthetic data generation pipelines, and is an important topic that has been previously surveyed~\citep{albalak2024surveydataselectionlanguage, qin2024unleashingpowerdatatsunami,wang2024surveydataselectionllm}.
~\citet{albalak2024surveydataselectionlanguage} present a systematic survey of data selection methods focused on language model pre-training, and of particular importance to the current work, they point out that data selection methods generally fall under one of two categories: \emph{distribution matching} and \emph{distribution diversification} methods, which are closely related to quality and diversity, respectively.
~\citet{qin2024unleashingpowerdatatsunami} present a survey of data selection methods for instruction tuning, finding that methods can be categorized into three groups: quality-based, diversity-based, and importance-based.
%\elliot{Why do we do complexity instead of importance? Is importance limited in some way?}
%\alexh{It's not the same thing as complexity and less directly related to open-endedness literature.}
~\citet{wang2024surveydataselectionllm} also present a survey on data selection for instruction tuning; however, their work focuses on describing how a sample of popular datasets were created.

\paragraph{Quality-diversity (QD) and open-endedness}
% \andrew{TODO: QD in evolution, OE as paragraphs from surveys}
% Open-ended learning: \citep{openendedlearningteam2021openendedlearningleadsgenerally}, \citep{wu2024evolutionarycomputationeralarge} + Andrew's survey \andrew{OELM survey pending, will worry about this later - OEL paper here isn't a survey so maybe we can use a different survey ref}
Quality-diversity (QD) algorithms \citep{pugh2016quality, cully2017quality, chatzilygeroudis2021quality} are a class of search algorithms originating from evolutionary computation \citep{lehman2011evolving,mouret2015illuminatingsearchspacesmapping} that aim for both quality and diversity among a population of solutions and artifacts, two of the key dataset attributes covered in this survey.
Such methods are inspired by the creativity of natural evolution in discovering diverse solutions (e.g. species) that also excel within the diverse niches they fill in the environment, and evolve into increasingly diverse and fit species in the population.
QD combines traditional objective optimization with insights from novelty search \citep{lehman2011abandoning}, an open-ended algorithm that overcomes local optimality through the continual accumulation of novel solutions.
By generating and maintaining a collection of diverse solutions, and then selecting for the next generation of solutions that are either increasingly novel, or optimized improvements to existing solutions within similar niches, QD leverages this growing collection to discover more diverse, high-quality solutions without having to trade off between quality and diversity.
QD methods have been applied recently for their illuminating search capabilities towards generating diverse, high-quality synthetic data for training models (cf. \Cref{sec:qd_algo}).
QD research is aligned with the study of open-ended systems, or open-endedness (OE) \citep{soros2017open,song2022little}, a broader term for the field that emerged from the study of open-ended evolution \citep{packard2019overview}. OE studies systems designed to generate and discover continually ``novel" and ``interesting" outcomes, and takes inspiration from real-world open-ended processes such as natural evolution, and human collective innovation. OE has become a key subject for providing new ways to tackle challenges in AI research, for example, towards generating open-ended synthetic data from which models can learn \citep{jiang2023general,sigaud2023definition,hughes2024openendedness,samvelyan2024rainbowteamingopenendedgeneration}.
LLM-based tools may provide new opportunities to advance surveyed methods in synthetic data generation, as OE and evolutionary methods are becoming more integrated with LLM components \citep{lehman2022evolutionlargemodels,meyerson2023language,zhang2023omni,wu2024evolutionarycomputationeralarge,chao2024match}.
%\elliot{Per the title of this section, would be good to clarify which of the above QD papers are surveys.}
%\alexh{I don't think any of them are surveys. But it's still nice to have a section introducing open-endedness as a related field.}
%\andrew{\citep{jiang2023general,sigaud2023definition,hughes2024openendedness,soros2017open,song2022little} are position papers, \citep{wu2024evolutionarycomputationeralarge,chao2024match} are surveys, and I think \citep{pugh2016quality, cully2017quality, chatzilygeroudis2021quality} are surveys, reviews or chapters to describe the emergence of QD as a paradigm in evolution}

\textbf{This survey}\quad Our survey complements the above perspectives on synthetic data generation, open-endedness, and quality diversity (QD).
We unify these findings to form a broader view of how future work in data generation and selection can emerge from distinct fields. This is done by providing a taxonomy for synthetic data through the lens of quality, diversity, and complexity, thus providing a better framework for understanding trade-offs and inefficiencies in the synthetic data generation process. We ground this framework with concrete takeaways and best practices in popular domains, including pre-training, instruction-tuning, and reasoning. We conclude by offering a list of open-problems and promising future research directions to better understand the intersection of synthetic data generation and QDC. The above summary of existing works and surveys highlights the important gaps that this survey fills.

%% file: sections/qdc_measures.tex
\section{Defining Data Quality, Diversity, and Complexity}
\label{sec:qdc_measures}

\begin{figure}
    \centering
    \includegraphics[width=0.7\textwidth]{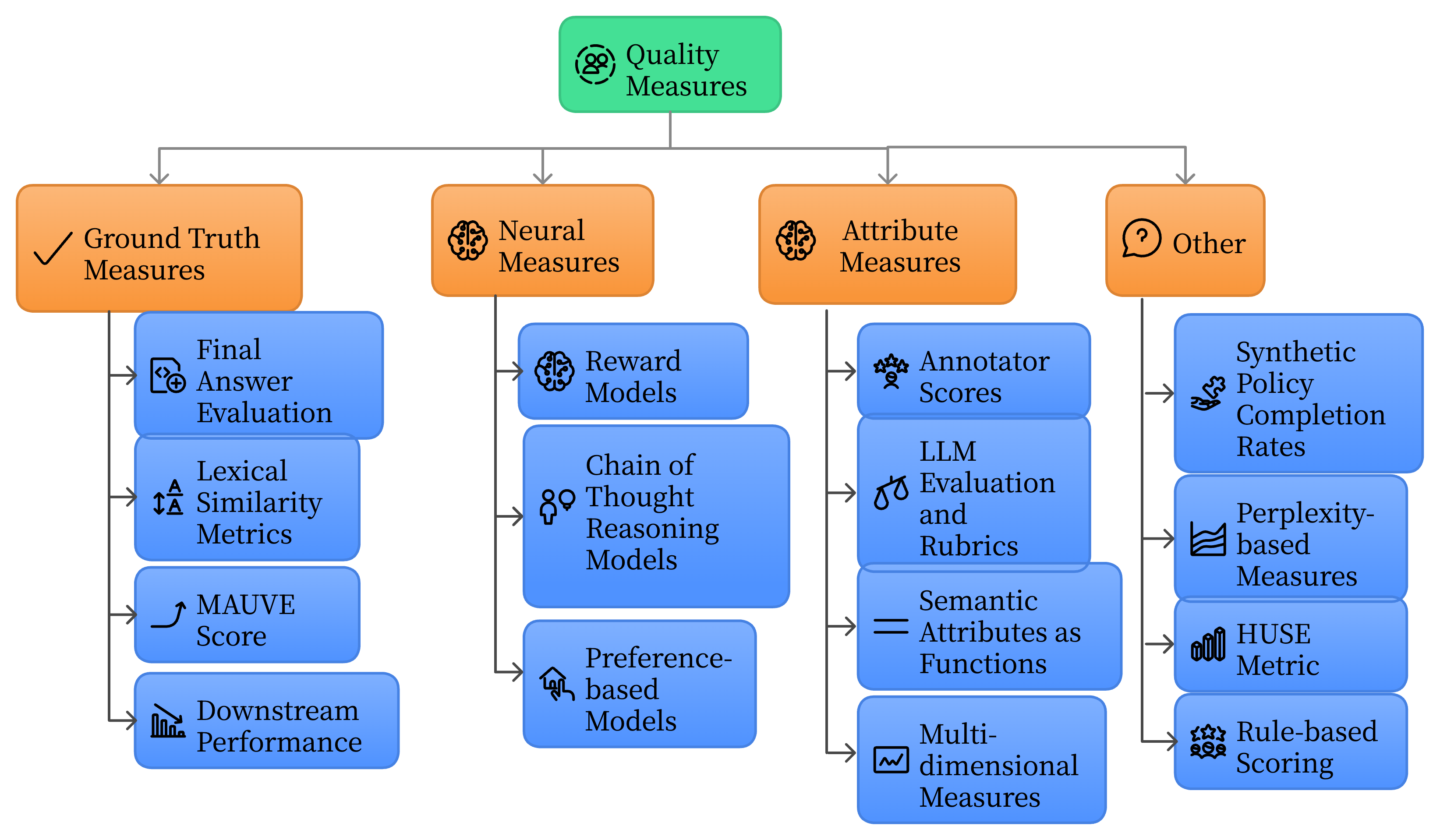}
    \caption{Quality Metrics}
    \label{fig:quality_metrics}
\end{figure}

Suppose we have some sample space $\Omega = \mathcal{X} \times \mathcal{Y}$ where each $\omega = (x,y) \in \Omega$ is an input-output sample pair. Further suppose we have a set of tasks $\tau_1,...,\tau_k$ defined as probability measures on $\Omega$. Finally, suppose we are given a large $n$-sample training dataset $\mathcal{D} \in \Omega^n$ and a model $\textbf{M}_\theta$. Note that $\mathcal{D}$ need not necessarily be sampled depending on the tasks $\tau_1,...,\tau_k$. Given some objective/loss $l$, it is often of interest to find characteristics of $\mathcal{D}$ which can be used to predict the downstream performance of $\textbf{M}_\theta$ on tasks $\tau_1,...,\tau_k$. In this survey we are interested in understanding the impact of three intuitive characteristics: dataset quality $Q(\mathcal{D})$, dataset diversity $D(\mathcal{D})$, and dataset complexity $C(\mathcal{D})$. 

%\andrew{How should we incorporate/handle "quantity" as a consideration of (maybe) "diversity"? Readers (especially from Twitter/seeing Karpathy's tweet) might be interested in this, we could decide on the scope to deal with the gap for the reader (even later), or state if it's beyond the scope/not interesting depending on the narrative. For me, I do think quality and diversity are more interesting/impactful measures}

%\alexh{I think quantity is is somewhat orthogonal to QDC. But definitely also relevant from the scaling law perspective.}

%\alon{I think we can make a nice table that summarizes the measures of quality and diversity. This may also help us to compare/categorize methods while writing.}

%\alexh{Yes!}

Despite being intuitive, widely used terms in ML, defining exactly what is meant by dataset quality, diversity, and complexity can be a surprisingly difficult task. Numerous implementations of proxy measures exist which attempt to capture our intuitive notions of these characteristics. To further complicate matters, different measures correlate better with downstream metrics of interest (such as model performance) in different settings. This makes choosing a single definitive measure practically impossible. Instead, we attempt to define abstract notions of quality, diversity, and complexity that line up with our intuitions as best as possible. With these formalisms in place, we survey the many different practical implementations of each characteristic found in the literature. Along the way, we discuss trade-offs of different types of implementations and their utility in predicting and improving downstream metrics of interest such as model performance. We determine the overall utility of a particular metric by assessing three key qualities: (1) \textit{applicability}, how widely applicable the metric is across domains, (2) \textit{cost}, how expensive is the metric to compute, and (3) \textit{performance}, how closely does the metric correlate with downstream model performance. Note: all the metrics we discuss are collected in tables in Appendix Section \ref{sec:qdc_metrics_table}.

%Define an ambient task space $\Omega = \mathcal{X} \times  \mathcal{Y}$ with a probability measure $P$ on $\Omega$ and let $\mathcal{D} \in \Omega^n$ be a training dataset with a fixed number of samples $n \in \mathbb{N}$. The quality, diversity, and complexity measures $Q,D,C$ are then scalar functions $Q, D, C: \Omega^n \to \mathbb{R}$. 

%(\alon{I'm not sure how to fit this in, but shouldn't quality also dependent on the desired behaviors? For example, toxic language is low quality or undesirable for training a generative model, but high quality for a toxicity detection model.}
%\alexh{What do you think about my reframing in terms of a target distribution $P$?}) 
%\elliot{I see what you're getting at, but it's not clear here exactly what $P$ is. Is it defined w.r.t. the training distribution or reality? I.e., is $\mathcal{D}$ sampled from $\Omega^n$ w.r.t. $P$?}

%Often instead the diversity of $\mathcal{D}$ is 

% \alexh{Still not 100\% happy with this framing.}

\subsection{Defining Dataset Quality} 
\label{subsec:quality_measures}

Fix a target task $\tau$ as a probability distribution on the sample space $\Omega$. Informally, we say the quality $Q(\mathcal{D})$ of dataset $\mathcal{D}$ aims to measure the ``noisiness'' or ``correctness'' of samples in $\mathcal{D}$ with respect to $\tau$. High-quality datasets with respect to $\tau$ should be entirely contained inside the support of $\tau$ in the sample space $\Omega$. Low-quality datasets will contain many samples far out of distribution of $\tau$. Often, quality measures are defined at the sample level $Q_\Omega: \Omega \to \mathbb{R}$. For example, the ``quality'' of a piece of code could be determined by how many unit tests it passes. In these common cases we could regard the corresponding task distribution $\tau$ which concentrates on the set of code samples passing all unit tests. A quality measure for the entire dataset can then be extracted by averaging the sample level quality: $Q(\mathcal{D}) = \frac{1}{n}\sum_{\omega \in \mathcal{D}} Q_\Omega(\omega)$.

Implementations of $Q$ can be categorized into four groups: ground truth measures, neural measures, attribute measures, and hybrid / other.

\textbf{Ground truth reference measures:}\quad Most often the data quality of a sample $\omega$ is measured by comparing to a corresponding ground truth sample $\omega^*$. The sample level quality $Q_\Omega(\omega)$ is then the similarity of $\omega$ to $\omega^*$. In domains where there is an intended final answer $A^*$, e.g., math \citep{yu2024metamathbootstrapmathematicalquestions, toshniwal2024openmathinstruct118millionmath, singh2024humandatascalingselftraining}, coding (HumanEval, \citet{pourcel2024acesgeneratingdiverseprogramming}), and other reasoning settings, the quality of a sample $\omega$ can be easily measured as

\begin{equation*}
    Q(\omega) = \begin{cases}
                    1 & A_\omega = A^* \\
                    0 & \text{otherwise}.
                \end{cases}
\end{equation*}

However, this comes with the obvious drawback of ignoring any intermediate steps in the chain of thought used to produce the final answer $A_\omega$. Further, more complex, open-ended tasks such as theorem proving and creative writing do not have a final answer. In these cases, another measure must be used to assess quality. When a ground truth sample $\omega^*$ is available, lexical measures of similarity to $\omega^*$, such as Rouge or Bleu/SacreBleu \citep{samvelyan2024rainbowteamingopenendedgeneration} can be used. While these measures are popular for less complex tasks such as paragraph-level summarization \citep{stiennon2022learningsummarizehumanfeedback}, lexical similarity is often insufficient for evaluating correctness on more complex tasks such as instruction following \citep{liu2024makesgooddataalignment}. In such cases, alternative metrics have been proposed. MAUVE \citep{pillutla2021mauve, ye2022progenprogressivezeroshotdataset} proposes to approximate the area under the divergence curve between the given data and a reference dataset. Other works simply define quality as performance on a downstream benchmark: \citet{zhou2023limaalignment, viswanathan2023prompt2modelgeneratingdeployablemodels,gandhi2024streamsearchsoslearning}.

\textbf{Neural measures:}\quad When training data is available, \textit{reward models} can be trained and generalize to unseen samples. The simplest example of a reward model is a classifier trained using labeled data (e.g., a toxicity classifier \citep{chakrabarty2019machinelearningapproachcomment}). A number of works have trained binary classification models using data that is known to be high-quality as the positive class examples, and unfiltered data as the negative class examples \citep{brown2020languagemodelsfewshotlearners,gao2020pile800gbdatasetdiverse, du2022glamefficientscalinglanguage, xie2023dataselectionlanguagemodels,li2024datacomplmsearchgenerationtraining}. In domains such as reasoning, specialized reward models such as Outcome-based reward models (ORMs) \citep{uesato2022solvingmathwordproblems} are used, which are trained with a classification objective at the step level supervised by the the final answer \citep{pang2024iterativereasoningpreferenceoptimization, tian2024selfimprovementllmsimaginationsearching, havrilla2024glorewhenwhereimprove}. The resulting reward model outperforms final answer classifiers and (somewhat) generalizes to intermediate steps. Process-based reward models (PRMs) \citep{lightman2023letsverifystepstep} further improve on ORMs by training a step-level classifier with supervised labels at each step. The PRM can then be used as a quality measure of each step of a solution $\omega$ in addition to the full trace. However, collecting human annotations for training PRMs is expensive. Recent work \citep{havrilla2024glorewhenwhereimprove,wang2024mathshepherdverifyreinforcellms} has investigated automating this process with synthetic data.
%\elliot{It could be useful to add equations for measures when possible, so they can be compared formally and visually by the reader. One might see some interesting patterns.}
\citet{ye2024improvingrewardmodelssynthetic} use synthetic natural language critiques generated by LLMs to provide additional feedback and then train a reward model that predicts a scalar reward on top of them.

A related line of work trains so called Bradley-Terry reward models contrastively on ordered preference data of the form $(\omega_-, \omega_+)$ where $\omega_+$ is the accepted sample and $\omega_-$ is the rejected sample \citep{ziegler2020finetuninglanguagemodelshuman,ouyang2022traininglanguagemodelsfollow,bai2022traininghelpfulharmlessassistant,kirk2024understandingeffectsrlhfllm, bukharin2024datadiversitymattersrobust,havrilla-etal-2023-trlx,bai2022traininghelpfulharmlessassistant, dubey2024llama}. Classifiers can also be used in this setting by conditioning on a ground truth reference, as in the Stanford Human Preferences Dataset \citep{pmlr-v162-ethayarajh22a}. A quickly growing line of work \citep{ankner2024critiqueoutloudrewardmodels,zhang2024generativeverifiersrewardmodeling,mahan2024generativerewardmodels} trains reward models with the ubiquitous next-token prediction objective instead of a discriminative objective to generate critiques or explanations before providing a quality score. Such approaches appear to generalize further out of distribution than their purely discriminative counterparts due to their CoT reasoning capability and ability to effectively utilize more inference-time compute. Also related is the zero-shot or few-shot use of LLMs-as-a-judge \citep{zheng2023judgingllmasajudgemtbenchchatbot} to perform in-context preference selection over multiple candidates.

\textbf{Attribute measures:}\quad When ground truth data is not available to assess data quality another option is to rely on data annotators to assess sample \textit{attributes} relevant to quality. Abstractly, a sample attribute is a generic function $T: \Omega \to [0, 1]$ measuring some semantic property of the sample. $T(\omega) = 1$ signifies that $\omega$ has the semantic property, $T(\omega) = 0$ signifies that $\omega$ does not have the semantic property, and $0 < T(\omega) < 1$ signifies that $\omega$ somewhat has the property. A quality score $Q$ for a sample can then be recovered via some combination $f$ of the attributes $Q(\omega) = f(T_1(\omega),...,T_k(\omega))$. 

Each attribute function $T$ is often implemented as a discrete Likert score \citep{alma991033023919703276} over the integers 1-N (which can then be normalized into the range $[0,1]$). %\elliot{This appears to contradict $T(\omega) \in [0,1]$ above.} 
For example, \citet{bai2022traininghelpfulharmlessassistant} recruit annotators to measure helpfulness, harmlessness, and honesty of AI assistant responses on a scale of 1-5, and \citet{stiennon2022learningsummarizehumanfeedback} recruit annotators to measure the coverage, clarity, and fidelity of a summary. LLMs themselves are increasingly being used as annotators to judge how well a sample adheres to a written constitution \citep{bai2022constitutionalaiharmlessnessai,samvelyan2024rainbowteamingopenendedgeneration}), to measure the difficulty of a problem \citep{gandhi2024bettersyntheticdataretrieving}, or to compare data examples across various dimensions of quality \citep{wettig2024quratingselectinghighqualitydata}.

A related approach to evaluating quality, uses sample \textit{rubrics} to assess the correctness of open-ended reasoning and language tasks. \citet{sawada2023arbadvancedreasoningbenchmark} use GPT-4 to write solution rubrics for hard STEM problems given example ground truth answers. While previous reference solution metrics were based on Bleu/Rouge, rubric based approaches generalize better to measuring important attribute based sample features. For example, EvolQuality evolved solutions of increasing quality in a manner similar to WizardLM \citep{luo2023wizardcoderempoweringcodelarge} and then applies GPT-4 as a judge to assess the quality of each sample of the sequence in the same shared context \citep{liu2024makesgooddataalignment}.
\citet{zhong2022towards} propose UniEval, a method that assigns multidimensional quality measures $T_{i}\in[0,1],   i\in\{1,...,k\}$, where $T_{i}$ is defined as 
\begin{equation}
    T_{i}(\omega) = \frac{P(\text{Yes}|\omega, q_{i})}{P(\text{Yes}|\omega, q_{i})+P(\text{No}|\omega, q_{i})},
\end{equation}
where $P(.)$ is the probability of the model to generate a specific word, and the input $q_{i}$ is a question specific to the quality being measured. For instance, the question to measure coherence could be ``Is this a coherent extraction of the document?". The same process can be repeated for other dimensions including fluency, understandability, naturalness, consistency, groundedness, engagingness, and others.
% For this type of quality measurements the authors perform continual learning on top of T5 with a specific order of the measures, and using appropriate datasets for which they generate positive and negative pseudo samples.
\citep{li2024rule} introduce an automated, rule-based framework for filtering for high-quality training data. They generate a diverse set of rules using an LLM, before rating a batch of data based on these rules. Following this, they use the determinantal point process
(DPP) from random matrix theory to select the most orthogonal score vectors, thereby identifying a
set of independent rules used to evaluate all data to select for training.
This rule evaluation metric is designed to promote low correlation and high diversity among rules. 
% Their automated pipeline generates the rules,
% selects the rules, and then chooses data according to rule-based ratings.

\textbf{Hybrid/other measures:}\quad Several other miscellaneous measures of dataset quality can be found in the literature. \citet{fontaine2021illuminatingmariosceneslatent} generates synthetic game levels and uses the completion rate of an expert policy as a measure of quality. A similar idea is explored in \citet{havrilla2024teachinglargelanguagemodels} by evaluating the quality of synthetically generated questions by the solve rate of a fixed student policy. Other works use perplexity as a measure for quality \citep{wenzek-etal-2020-ccnet,sharma2024textqualitybasedpruningefficient}. \citet{hashimoto2019unifyinghumanstatisticalevaluation} introduces the HUSE metric for jointly measuring the quality and diversity of machine generated text. HUSE utilizes averaged annotator quality ratings for a per-sample quality measure and length normalized model log-likelihood to measure diversity.

%\elliot{It could be possible/cleaner to fit all these hybrid/other measures under either Neural or Attribute. E.g., HUSE is an attribute measure where each annotator defines an attribute.}

%However, HUSE requires human intervention and thus is not an automatic step as the others.

\subsection{Defining Dataset Diversity}
\label{subsesc:diversity_measures}

\begin{figure}
    \centering
    \includegraphics[width=0.7\textwidth]{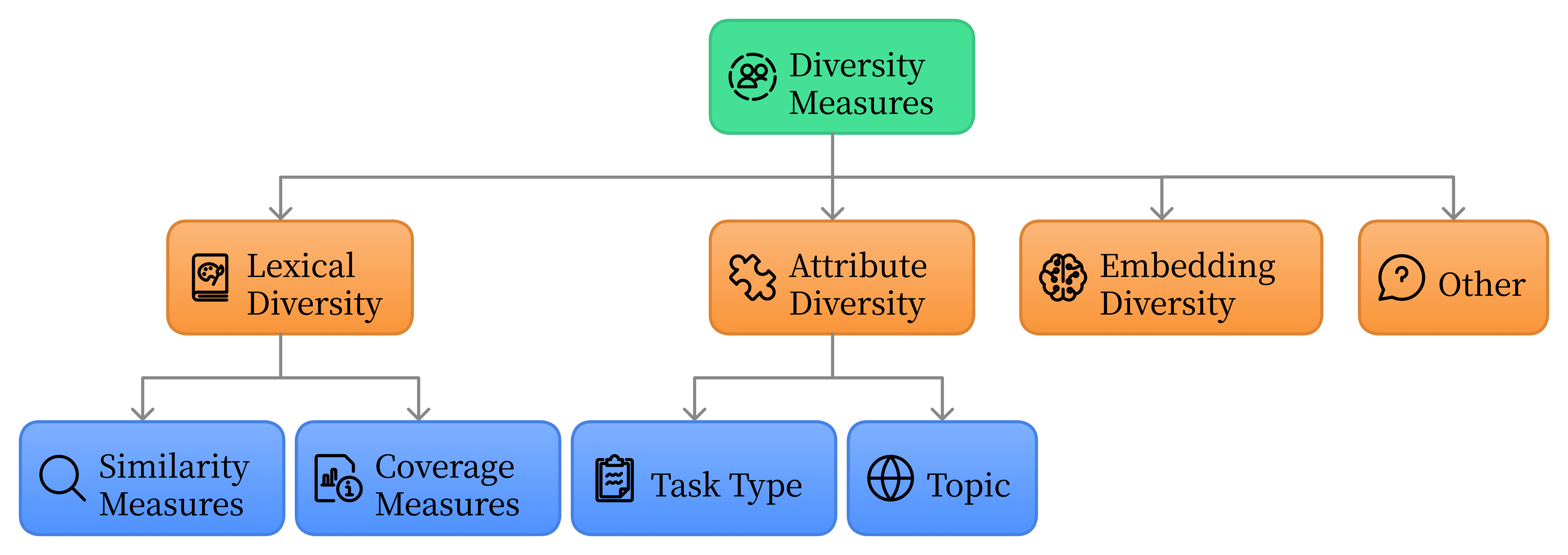}
    \caption{Diversity Metrics}
    \label{fig:diversity_metrics}
\end{figure}

A diversity measure $D(\mathcal{D})$ aims to capture the ``self-similarity'' and ``coverage'' of $\mathcal{D}$ over the entire sample space $\Omega$. A highly diverse dataset should be uniformly distributed on $\Omega$. This motivates us to consider formally defining the diversity of $\mathcal{D}$ as the distance between $\mathcal{D}$ and the uniform measure on $\Omega$ under some chosen probability metric. Importantly, this means the diversity of $\mathcal{D}$ is not necessarily dependent on a target task $\tau$. However, in some settings where we have a good prior over the sample space $\Omega$ it may make more sense to adapt our notions of diversity to be uniform with respect to the prior instead of uniform over the entire sample space. For example, in the language setting a more sensible uniform prior may be the uniform distribution over all natural language sentences instead of uniform over all possible strings up to a fixed finite length.

Unlike quality, it is difficult to define diversity at the sample level. We categorize practical diversity measures into four groups: lexical, attribute, embedding, and miscellaneous as summarized in Figure~\ref{fig:diversity_metrics}.

\textbf{Ground truth measures:}\quad \textbf{Lexical similarity measures} can be used to capture a relatively simple type of diversity in text data by comparing n-grams between source and target datasets. 
Scores like ROUGE \citep{lin-2004-rouge,gandhi2024streamsearchsoslearning}, Bleu \citep{Papineni2002BleuAM, samvelyan2024rainbowteamingopenendedgeneration, ye2022zerogenefficientzeroshotlearning,zhu2018texygenbenchmarkingplatformtext, oattributing}, and INGF \citep{yu2023largelanguagemodelattributed, kirk2024understandingeffectsrlhfllm}
measure pairwise similarity between samples in a dataset $\mathcal{D}$ by pairing n-gram overlap between samples. A diversity measure for $\mathcal{D}$ is then assigned by averaging all 
pairwise similarity measures. We refer to these kinds of diversity measures as \textit{similarity measures}. Dataset vocabulary size, i.e., the number of unique words in a dataset, can also be used as a simple diversity measure which does not rely on pairwise
lexical comparisons between samples \citep{yu2023largelanguagemodelattributed}. We refer these types of diversity measures as \textit{coverage measures}. Measure of Textual Lexicial Diversity (\textbf{MTLD}) \citep{mtld, cao2024instructionmininginstructiondata} is another measure of lexical diversity which analyzes the average number of words in a row that maintain a certain Type-Token Ratio (TTR). The ``distinct'' metric \citep{li-etal-2016-diversity} measures diversity of a collection of texts as the ratio of distinct tokens to the total number of tokens in the text. The expectation adjusted distinct metrict (\textbf{EAD}) additionally normalizes by the expected number of distinct tokens under some prior. 

Lexical diversity measures are general-purpose, requiring no special domain knowledge for application. However, for the same reason
they often do not capture more relevant/interesting notions of diversity in domain specific settings: e.g., the type of skill used to solve a programming problem \citep{pourcel2024acesgeneratingdiverseprogramming}.

%\elliot{At a higher level: How do we know which of these measures are `good'. E.g., are there some ideal desiderata that can be given for each characteristic and check which each measure satisfies? E.g., general, open-ended, efficiently-computable, non-reliant on ground truth.}

%(Lexical similarity to of dataset to another ground truth dataset?) 

\textbf{Attribute diversity:}\quad Similar to attribute-based measures of quality, attribute-based measures of diversity 
also require domain-specific knowledge to define attribute functions $T_i: \Omega \to [0, 1]$. Each attribute function $T_i$ 
measures where a given sample $\omega$ falls in the semantic attribute $T_i$. The collection of attribute measures 
$T(\omega) = (T_1(\omega),...,T_k(\omega))$ defines an attribute description for each sample $\omega$. 
Concretely, the attributes computed for a dataset vary from task to task. Many instruction tuning papers \citep{wang2022supernaturalinstructionsgeneralizationdeclarativeinstructions, li2024selfalignmentinstructionbacktranslation,zhang2024instructiondiversitydrivesgeneralization, lu2023instaginstructiontagginganalyzing}
categorize instruction by task type (summarization, question answering, translation, etc.) to measure instruction diversity. 
Some papers \citep{zhou2023limaalignment, wang2022supernaturalinstructionsgeneralizationdeclarativeinstructions} also annotate the topic, e.g., politics, health, sports, etc., of each instruction. 
\citet{wang2022supernaturalinstructionsgeneralizationdeclarativeinstructions} additionally classifies instructions by language to measure instruction diversity along three axes: 
type, topic, and language. Other works describe attributes using \textit{behavioral descriptors}. \citet{samvelyan2024rainbowteamingopenendedgeneration}
classifies red-teaming prompts by what constitutional rules are violated. \citet{fontaine2021illuminatingmariosceneslatent} counts the number of attributes contained in a description of a tile-based Mario level. 
\citet{pourcel2024acesgeneratingdiverseprogramming} categorizes solutions to programming puzzles by tracking what skills are used in the solution (e.g., arrays,
graphs, dynamic programming, etc.). \citet{tian2024selfimprovementllmsimaginationsearching, havrilla2024teachinglargelanguagemodels} identify solutions to math problems by their orders of operations.
\citet{gandhi2024streamsearchsoslearning} measures the diversity of learned search algorithms playing the Game of 24 by the number of unique states they visit 
while finding a solution.

The main difficulty with measuring attribute-based diversity is in accurately measuring the desired attributes of an
unstructured text sample $\omega$. Implementations of these measures fall largely into two approaches: 
(1) relying on human annotators or (2) using LLMs as judges to automatically annotate samples.
Earlier works \citep{zhou2023limaalignment, wang2022supernaturalinstructionsgeneralizationdeclarativeinstructions} utilize attribute labels coming exclusively from
human annotators. The marked improvement in LLM capability made automatic annotation of attributes only recently possible. 
Several works have already experimented with this approach \citep{samvelyan2024rainbowteamingopenendedgeneration, pourcel2024acesgeneratingdiverseprogramming, bradley2023qualitydiversityaifeedback, bai2022constitutionalaiharmlessnessai}, 
with results varying depending on the complexity of the attribute. \citet{yu2023largelanguagemodelattributed} experimented with GPT for automated attribute discovery. Instag \citep{lu2023instaginstructiontagginganalyzing} uses GPT-4 to generate sets of skill/topic labels for instruction tuning data.

Many different diversity measures can be implemented once a set of attributes $\{T(\omega) : \omega \in \mathcal{D}\}$ 
has been measured for every sample in the dataset. Similarity-based measures can be implemented by computing the pairwise
similarities bewteen attribute profiles. \citet{pourcel2024acesgeneratingdiverseprogramming} implement a similarity-based measure for python puzzle solutions
by computing binary indicator attributes for programming skills. Hamming distance can then be used to measure 
distance between the binary attribute profiles. A coverage-based approach can also be taken by measuring the total number 
cells in the boolean grid occupied by a sample. \citet{samvelyan2024rainbowteamingopenendedgeneration} use a similar coverage-based approach to measure 
diversity of red-teamed LLM responses based on what constitutional rules are violated. QD-score \citep{tjanaka2022quantifying} jointly measures the quality and diversity of solutions in a grid by summing the fittest sample in each cell: $\sum_{c \in \mathcal{G}} f(c)$.
%\elliot{Worth mentioning QD-score here.}

In general, attribute-based measures of diversity are effective at measuring interesting, domain-specific features
which capture useful notions of diversity. However, measuring such attributes can be difficult and expensive. 
LLM-as-a-judge approaches have made automating attribute labeling more feasible, but struggle on more complex tasks/attributes.

%\andrew{For structure, would LLM-derived (e.g. labels from Rainbow Teaming) and non-LLM-derived attributes (i.e. manual/programmatic rules and counting of instances of attributes in implementation) be a useful division amongst paragraphs? It could also motivate future work considering more LLM-derived diversity measures, which could themselves be dynamically generated.}

\textbf{Embedding diversity:}\quad Another set of popular dataset diversity measures are computed by embedding the dataset 
$\mathcal{D}$ as latent space embeddings \citep{reimers2019sentencebertsentenceembeddingsusing, zhao-etal-2019-moverscore}. 
\citet{pourcel2024acesgeneratingdiverseprogramming} take a similar approach by embedding solutions to programming puzzles.
Many other works \citep{yu2023largelanguagemodelattributed, kirk2024understandingeffectsrlhfllm, yu2024metamathbootstrapmathematicalquestions} embed natural text using Sentence-BERT \citep{reimers2019sentencebertsentenceembeddingsusing} and measure average pairwise cosine similarity (\textbf{APS}) between the embeddings. \citet{cao2024instructionmininginstructiondata} measure diversity using $i$th-Nearest neighbor distance in SentenceBERT embedding space. Similarly, SemDeDup~\citep{abbas2023semdedupdataefficientlearningwebscale} and D4~\citep{tirumala2023d4improvingllmpretraining} measure the diversity of data points by embedding data, however they use a generative model, and their goal is to remove semantically duplicated data. \citet{ding2024qualitydiversityhumanfeedback} train an embedding model to measure sample diversity using a triplet-based contrastive objective. They do this by collecting triplet preferences of the form $y_1 \sim y_2$ or $y_1 \sim y_3$ where $y_1 \sim y_2$ indicates $y_1$ is more similar to $y_2$ than $y_3$. \citet{lee2023scalediversitycoefficientdata} measure the \textit{diversity coefficient} of pre-training data by computing the Fisher information matrix of 
gpt-2 gradients. \citet{bukharin2024datadiversitymattersrobust} is a type of coverage measure which measures the diversity of instruction following datasets 
using the \textit{facility location function} $d(A) = \sum_{v \in V} \max_{a \in A} sim(a,v)$ which relies on a background dataset $V$ to measure the diversity of $A$. Cosine similarity between embeddings is used as the similarity measure.

Measures of diversity using latent space embeddings do not require domain knowledge and thus serve as general purpose diversity measures
(provided a general enough embedding model). However, for more domain specific/complex tasks embeddings may fail to capture certain attribute-based features essential to comparing samples in the task. This may result in a weaker correlation with downstream metrics, when compared to attribute measures, depending on the domain.

\textbf{Other diversity measures:}\quad Many other miscellaneous measures of dataset diversity can be found in the literature. \citet{kirk2024understandingeffectsrlhfllm} measures the diversity of summaries for a fixed passage 
by computing the average number of pairwise contradictions via a natural language inference model. \citet{singh2024humandatascalingselftraining,havrilla2024teachinglargelanguagemodels,toshniwal2024openmathinstruct118millionmath}
measure the solution diversity of math problems via the pass@n score on the test set i.e., the number of problems that can be solved with an $n$-solution sample budget. \citet{zhao2024positionmeasuredatasetdiversity} propose using tools from measurement theory as measures of dataset diversity.

% Some papers \citep{?} measure the entropy of the dataset either in attribute space or embedding space. 

%\elliot{Similar to Quality measures, I think these other diversity measures can be grouped under the defined categories, e.g., Fisher under embeddings, flf under attributes (since its a coverage metric where each ground truth point defines an attribute).}
%\alexh{I moved all the ones I thought could be fit in elsewhere}

\subsection{Defining Dataset Complexity}
\label{subsec:complexity_measures}

\begin{figure}
    \centering
    \includegraphics[width=0.6\textwidth]{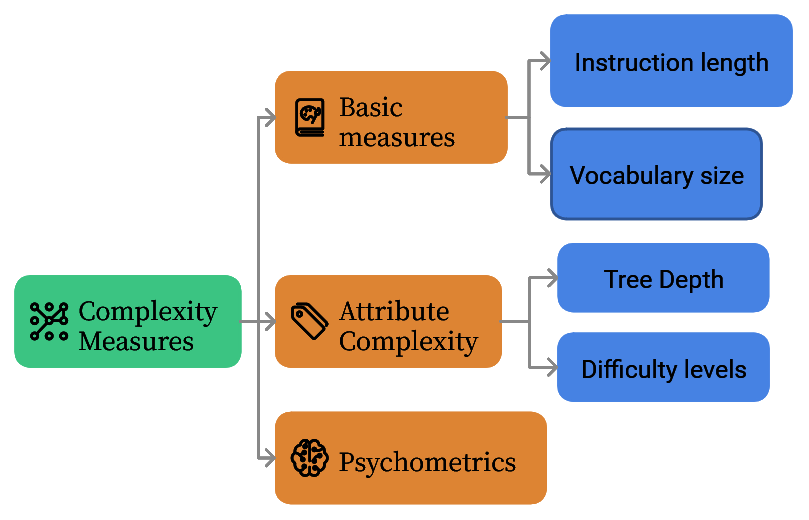}
    \caption{Complexity metrics.}
    \label{fig:complexity_metrics}
\end{figure}

Complexity is a third essential characteristic of data that intuitively measures some notion of the ``difficulty'' or ``compositionality'' of a sample. While less commonly considered than quality and diversity, improving output complexity is essential for any algorithm for recursive self-improvement through synthetic data. This is because the output complexity of a model reflects its capability to compose distinct concepts across domains. A model with high-quality and highly-diverse output may reliably recall the data it was trained on across a variety of tasks. However, if the model has low output complexity, it may completely fail to learn how to combine pieces of knowledge it has already learned. Learning how to do this type of composition is essential for intelligence.

%\andrew{At the dataset level, maybe complexity could also express the interconnectedness of knowledge that could be acquired from the collection of samples as a whole? Maybe, for example, having the model learn about creative writing in one document, then maths reasoning, and no more, may lead to fewer opportunities for the model to learn composable knowledge building blocks/foundations for more complex, newer knowledge, but having the corpus express learnable knowledge that somewhat intersects topics like biology and chemistry could give rise to enough information for models to learn emergent knowledge of increasing complexity into further sub-branches of fields like medicine/biochem, given the nature of how the base samples in the dataset could compose increasingly complex knowledge trees and semantics. Or imagining some visual, conceptual intuition, if the branches of the subsystem completely diverge, the number of opportunities for branches to mingle and intersect with each other for new, more complex units/components to emerge, would be less than if the branches were closer to each other, and more opportunities for intersections could give rise to more emerging sub-components, grouped under new labels for the simplest-to-explain units to describe them. Elliot also made some useful comments in the appendix below on complexity.}

Despite its intuitiveness, complexity is a notoriously difficult concept to formalize \citep{mitchell_complexity}. \textit{Kolmogorov complexity} \citep{li1997introduction} proposes to measure complexity of a sample as the length of the shortest program generating the sample $\omega$ with respect to a fixed programming language. \textit{Logical depth} \citep{Bennett1988LogicalDA} is another notion of complexity which instead measures the minimum amount of \textit{time} needed to compute the sample $\omega$ when minimizing over all programs in a fixed programming language. While such notions are quite general, they are intractable to compute in common situations, necessitating practical alternatives. However, they do point to the common theme that most measures of complexity implicitly depend on a fixed background set of primitives used to construct the sample $\omega$.

To help express the foundations of the nature of complexity that could be studied under different measures, we give a more actionable definition for complexity. We define complexity $C(\omega)$ of a sample $\omega \in \Omega$ as its size under a fixed representation scheme. This also aims to align closely with the broader definition of complex systems as described in \citet{mitchell_complexity}: 

\begin{quote}
    A system in which large networks of components with no central control and simple rules of operation give rise to complex collective behavior, sophisticated information processing, and adaptation via learning or evolution.
\end{quote} This notion of complex systems could be important for future developments in dataset creation: learning increasingly complex knowledge is important for the development of more powerful models, and the evolution of more complex data is important for models to be able to have more learning opportunities in the first place. 

Concretely, we denote an $n$-samples complexity measure as function $C: \Omega^n \to \mathbb{R}$, which intuitively measures data ``difficulty". Similarly to quality, complexity can also be defined on a sample level as $C_\Omega \to \R$ with $C$ being recovered as the average over samples. We now survey commonly used measures of complexity in text data and summarize them in Figure \ref{fig:complexity_metrics}.

\paragraph{Basic measures of complexity:} We start by reviewing some simple measures of textual data complexity. Token length \citep{liu2024makesgooddataalignment} is broadly applicable across all text but lacks any notions of semantic complexity. Sometimes human generated labels can also be used to form complexity hierarchies as in Hendryck's MATH which groups problems into increasing levels of difficulty \citep{hendrycks2021measuringmathematicalproblemsolving}. Some other works measure the complexity of data as its compressibility \citep{pandey2024gzippredictsdatadependentscaling} or intrinsic dimension \citep{sharma2020neuralscalinglawdimension,havrilla2024understandingscalinglawsstatistical}.

\paragraph{Attribute complexity:} Similar to attribute based measures of quality and diversity, one can define an attribute function $T$, which measures for each sample its complexity in this case. One such attribute complexity metric is Tree instruct \citet{zhao2024preliminarystudyintrinsicrelationship}, which is defined as the number of nodes in the semantic tree of an instruction. The authors show empirically that for this specific complexity metric, the number of added tree nodes in the prompt, with increasing complexity performance increases. Moreover using less, but more complex data over performs more, but less complex data. In this case, it is shown that the better performance does not come due to more tokens. Another complexity metric falling into this category is InsTag \citep{lu2023instaginstructiontagginganalyzing}, where $T(\omega) = $ \# semantic tags $(x)$ assigned to sample $\omega$. Each set of tags is assigned using a prompted LLM.\citep{sharma2024textqualitybasedpruningefficient} also propose to measure complexity based on the depth and structure of a parse tree. Other works directly prompt GPT-4 to score the complexity of samples on a 1-5 scale \citep{chen2024alpagasustrainingbetteralpaca}. EvolComplexity \citep{liu2024makesgooddataalignment} computes complexity scores by evolving instruction responses in a manner similar to WizardLM \citep{luo2023wizardcoderempoweringcodelarge} and then applying GPT-4 as a judge to assess the complexity of each sequence of evolved samples in-context.

\paragraph{Other model based measures:} Model perplexity \citep{liu2024bestpracticeslessonslearned} can also be regarded as a measure of complexity, with higher perplexity samples being more complex. Note however there is some overlap with using perplexity as a measure of data quality. The Instruction Following Difficulty (IFD) score \citep{li2024quantityqualityboostingllm} applies perplexity to measuring complexity of instruction following samples as
\begin{equation}
    IFD(x,y) = \frac{s(x|y)}{s(x)},
\end{equation}
where $s$ denotes the cross entropy
\begin{equation}
    s(x) = -\frac{1}{N}\sum_{i=1}^{N}\log P(x_{i}|x_{1},...,x_{i-1};\theta)])
\end{equation}
and $\theta$ denotes the weights of the pretrained base LLM.  \citet{albalak2023efficient} similarly measure complexity as the information gain of a sample.

%\elliot{Maybe instead of `Other' complexity measures the below could be called `Structured' or `Model-based', since each (as far as I can tell) assumes some structured model the whole dataset is fed in to.}

%\andrew{TODO: one/few sentences in the notion of complexity to motivate more open-ended synthetic data generation, WizardLM (covered in "a little taxonomy of open-endedness")}

% \kg{we might want to include psychometric measures here; the way difficulty of a sample is measured with human responses.}
\paragraph{Pschometric measures:} Psychometric approaches like item response theory (IRT)\citep{rasch1993probabilistic, DeMars2010ItemRT}, traditionally used to assess human cognitive abilities and the difficulty of questions, could be adapted to measure the difficulty or complexity of datasets for LLMs. IRT allows the difficulty of datasets to be captured through parameters in a probabilistic model aimed at explaining model performance by sampling from a diverse population of LLMs as respondents. This method could involve presenting samples from the dataset to a variety of LLMs with different sizes and strengths. Measures such as the perplexity of data points could then be analyzed using  IRT \citep{thissen1982marginal} to generate standardized difficulty scores for each sample. Such measures have been used to adaptively evaluate models with fewer examples \citep{polo2024tinybenchmarks,zhuang2023efficiently}, and to compare model responses to a population of humans \citep{he2024psychometric} --- but could be extended to measure the complexity of the training data. This  would provide a nuanced measure tailored specifically to LLMs, which could be valuable for curating datasets of varying difficulty levels for training and testing.

\paragraph{Recap} We now summarize the main takeaways from the last three sections covering various measures of quality, diversity, and complexity:

\vspace{0.3cm}
\begin{takeaways}
    \setlength{\leftmargini}{0pt}
    \begin{itemize}
        \item General purpose textual similarity measures are broadly applicable but often fail to capture domain-specific features of interest. 
        \item Attribute-based measures must be handcrafted for specific domains but can be chosen to capture important features of interest.
    \end{itemize}
\end{takeaways}
\vspace{0.3cm}

These observations are reinforced by findings in \citet{liu2024makesgooddataalignment} who compare several different measures of quality, diversity, and complexity in instruction tuning data. They found model-based measures relying on GPT-4 as a judge to outperformed purely lexical quality and complexity measures. This leads to the following question:

\vspace{0.3cm}
\begin{openquestions}
    \setlength{\leftmargini}{0pt}
    \begin{itemize}
        \item How can we develop better domain agnostic measures capable of adapting to domain specific features of interest?
    \end{itemize}
\end{openquestions}
\vspace{0.3cm}

%% file: sections/qdc_effects.tex
\section{The Effects of Data QDC on Model Performance}\label{sec:qd_effects}

In Section \ref{sec:qdc_measures} we defined notions of quality, diversity, and complexity and surveyed numerous practical implementations used in the literature. In this section we now seek to understand the effects of quality, diversity, and complexity in training data on model performance and generalization. Overall, we find that quality tends to most benefit in-distribution generalization, diversity most benefits OOD generalization, and \textit{appropriate levels} of complexity can benefit both. Additionally, we examine often occurring trade-offs between quality and diversity in training data. As a result, this often produces corresponding trade-offs between in-distribution and OOD generalization.

\paragraph{A note on in-distribution vs. OOD generalization} Let $P$ be probability measure on task space $(\mathcal{X}, \mathcal{Y})$ and $\mathcal{D}_{\text{train}} = \{(X_i, Y_i)_{i=1}^n\} \sim P$ a set of training data independently identically sampled (\textbf{i.i.d}) from $P$. In this survey, when we say a model $\textbf{M}_\theta$ has good \textbf{in-distribution generalization}, we mean that $\textbf{M}_\theta$ trained on $\mathcal{D}_{\text{train}}$ generalizes well to test data $\{(X_j, Y_j)_{j=1}^t\}$ sampled from $P$. We say $\textbf{M}_\theta$ has good OOD generalization if it generalizes well to a related but new distribution $Q$ on $(\mathcal{X}, \mathcal{Y})$. See \citet{miller2021accuracylinestrongcorrelation} for more discussion on in-distribution versus OOD generalization.

%\fabrizio{I feel the note should be added when first talking about in-distribution out distribution}
%\elliot{I agree this is an odd place for this. It could be moved to the beginning of Section~\ref{sec:qd_data} where the basic formalism is introduced.}

\subsection{The Effect of Quality}
\label{subsec:quality_effects}

Data quality on its own has been a huge topic of interest recently for both model pre-training and post-training \citep{albalak2024surveydataselectionlanguage, zhou2023limaalignment, chen2024alpagasustrainingbetteralpaca, cao2024instructionmininginstructiondata, sharma2024textqualitybasedpruningefficient}. For this reason we break our discussion into two parts: one for pre-training and one for post-training.

\paragraph{Data quality in pre-training} SOTA pre-training data pipelines include several iterative rounds of quality filtering \citep{ albalak2024surveydataselectionlanguage}. The initial round applies of set of heuristic filters aimed at removing noise coming from malformed webtext \citep{penedo2024finewebdatasetsdecantingweb} and maximizing ``educational content'' \citep{yue2023mammothbuildingmathgeneralist}. \citet{sharma2024textqualitybasedpruningefficient} apply over a dozen quality-based filters to OpenWebText to improve LLM pre-training efficiency. They find pruning 40\% of the available tokens improves model benchmark performance. \citet{zhang2024autonomous, shao2024deepseekmathpushinglimitsmathematical} both propose measuring quality of mathematical texts with an in-context LLM classifier. They see up to two times increase in pre-training efficiency gains on common reasoning tasks (GSM8K, BBH, MATH) when training on the filtered datasets. \citet{xie2023dataselectionlanguagemodels} samples high-quality data for pre-training via importance resampling with similarity to a target distribution computed using a bag-of-words n-gram model. \citet{maini2024rephrasingwebrecipecompute} demonstrated improved data quality (and efficiency) by pretraining on web data rephrased to be similar to high quality text (such as Wikipedia). The Phi series of models~\citep{gunasekar2023textbooks,li2023textbooks,javaheripi2023phi,abdin2024phi} explores the improvement from increasing data quality as opposed to scale. \citet{gunasekar2023textbooks} trains a model on high quality data filtered by ChatGPT and achieves benchmark performance comparable to significantly larger models. Subsequent iterations \citep{li2023textbooks,javaheripi2023phi,abdin2024phi} rely heavily on high-quality synthetic data, demonstrating impressive benchmark gains on targeted reasoning tasks. Several works \citep{penedo2024finewebdatasetsdecantingweb, abdin2024phi, dubey2024llama3herdmodels} show training synthetic data to train model quality classifiers works extremely well in filtering out low-quality data.

%TODO: make sure to touch on Supervised fine-tuning, RLHF, instruction tuning
%TODO: make sure to include discussion of llama-3.1 somewhere
\paragraph{Data quality in post-training} Much recent work also demonstrates the importance of high-quality fine-tuning data \citep{zhou2023limaalignment,chen2024alpagasustrainingbetteralpaca,cao2024instructionmininginstructiondata}. \citet{zhou2023limaalignment} finds fine-tuning even on just 1000 highest quality samples improves instruction following performance. However, they also note lower quality samples has a negative effect unless more diverse samples are also included. Quality is measured using a Likert scale with scores assigned by both humans and GPT. Diversity is measured via topic tagging. Most of their evaluation is done in-distribution. Alpagasus \citep{chen2024alpagasustrainingbetteralpaca} filters the Alpaca instruction tuning dataset from 52k samples to 9k samples via Likert-scale quality annotations from ChatGPT. Fine-tuning on this smaller, higher quality dataset results in a more preferred (by GPT-4) instruction following model. \citet{liu2024makesgooddataalignment} propose a number of quality-diversity metrics (EvolQuality/EvolComplexity) which they use to filter instruction-tuning data. They find the Evol quality scores and RM score correlate well with downstream performance. 

% Question: what is the interplay between quality during both pre-training and post-training. are both necessary?

% Takeaway: quality helps in-distribution generalization
% Takeaway: Training classifiers with synthetic data is extremely useful

%TODO: emphasize the importance of finding a good metric that correlates well with what you want

\subsection{The Effect of Diversity}

Of equal importance to the quality of training data is its diversity during both pre-training and post-training. In contrast to quality, post-training approaches often do not see a huge improvement in in-distribution performance with more diverse training data. Instead, diversity most benefits OOD performance. 

\paragraph{Data diversity in pre-training} Data diversity plays a crucial role in pre-training \citep{Raffel2019ExploringTL, NEURIPS2020_1457c0d6, li2023textbooks}. Many pre-training methods assume that a post-training process will occur, thereby making the goal of pre-training for the model to see as much data as possible.
% duplication is inefficient, and can lead to overfitting
However, simply training on all available data can lead to highly duplicated content, and training inefficiency, motivating significant efforts to improve data deduplication.
% hash-based deduplication
For example, \citet{wenzek-etal-2020-ccnet} and \citet{OrtizSuarezSagotRomary2019} utilize hashing based deduplication methods when developing CCNet and OSCAR, respectively. Additionally, \citet{lee-etal-2022-deduplicating} propose the \textsc{ExactSubstr} algorithm, which detects duplicates based on shared substrings.
% model-based deduplication
More recently, model-based deduplication has become popular, with \textsc{SemDeDup}~\citep{abbas2023semdedupdataefficientlearningwebscale} and D4~\citep{tirumala2023d4improvingllmpretraining} being common examples. Each of these methods utilize a neural network to embed the data, followed by some clustering and a removal of data within each cluster according to some similarity metric. In particular for pre-training where the downstream goals are not known, data deduplication reduces the risk of models overfitting to anything in particular, and has been shown to sometimes improve accuracy on downstream tasks~\citep{tirumala2023d4improvingllmpretraining}.

Additionally, a number of works have found that training on more diverse data can lead to improved performance \citep{lee-etal-2022-deduplicating, tirumala2023d4improvingllmpretraining}.
For example, \citet{ye2024datamixinglawsoptimizing} found that pretraining on datasets with higher diversity generally improves performance on downstream domains. The Phi model series studies the impact of training data diversity from the perspective of training on synthetically generated data \citep{li2023textbooks, javaheripi2023phi, abdin2024phi}. They find it challenging to ensure that the model-generated data is truly diverse, and find methods of injecting randomness. For example, \citet{javaheripi2023phi} seed the generation prompts for their synthetic data with 20K topics (e.g. science, daily activities) and include filtered web samples in the training data for additional diversity. \citet{ye2024datamixinglawsoptimizing} finds diverse mixture weights for data-sources from different topics is optimal pre-training performance. This aligns with \cite{zhang2024textbfonlyif}'s finding that even imbalanced distributions of diverse data can still drive effective generalization, as long as there is sufficient semantic coverage across domains. \cite{eldan2023tinystoriessmalllanguagemodels} introduced TinyStories, a diverse synthetic dataset of simple stories. The authors demonstrate that small models trained on this diversity-controlled dataset can generate coherent text and show basic reasoning--capabilities that typically require much larger models on standard corpora. The impact of diversity appears to benefit both specialist and generalist models in distinct ways: For specialist models (e.g., code language models), \cite{zhang2024textbfonlyif} demonstrate that extending data diversification beyond their core domain yields substantial performance improvements in instruction following (up to a limit) as compared to training solely on domain-specific data. For generalist models, using diverse data mixtures enhances their instruction-following capabilities across a broad range of domains more effectively than simply increasing the quantity of training data. \cite{chen2024diversitysyntheticdataimpact} studies the impact of diversity in synthetic data on the training of LLMs by proposing a diversity measure pipeline using LLMs to perform clustering of text corpus while generating data. They show that higher synthetic data diversity correlates positively with pre-training and supervised fine-tuning performance. 
%\elliot{The above two paragraphs are more about methods for filtering for diversity than the effects of diversity.}

\paragraph{Data diversity in fine-tuning} In contrast to quality,
diversity can improve the in-distribution generalization performance, but only up to a certain point \citep{albalak2024improving}. Instead, training dataset diversity appears to be essential for out-of-distribution 
(\textbf{OOD}) generalization \citep{wang2022supernaturalinstructionsgeneralizationdeclarativeinstructions, yu2023largelanguagemodelattributed, fan2023scalinglawssyntheticimages}. The Supernatural instructions dataset 
\citep{wang2022supernaturalinstructionsgeneralizationdeclarativeinstructions} assembled a large instruction dataset comprising over 1600+ instruction types spanning a diverse collection of 
domains and languages. Their evaluation procedure proceeds by training models on subsets of instructions and testing generalization 
to held out OOD instruction sets. They found scaling both the number of instruction classes and
the model size has a significantly improved test performance. However, increasing the number of instruction demonstrations 
per instruction type beyond 64 had little to no impact. A similar observation about scaling the number of instruction types 
versus the number of demonstrations per type was made in \citep{zhou2023limaalignment}.\citet{bukharin2024datadiversitymattersrobust} design an algorithm that selects the highest quality samples, while making sure selected data is sufficiently diverse, by properly adjusting a parameter trading off data quality and diversity. They find training instruction-following models on more diverse datasets does not improve average-case performance on 
in-distribution benchmarks (AlpacaEval) but does improve performance on the hardest questions. In the reasoning domain, \citet{yu2024metamathbootstrapmathematicalquestions} found training on more diverse data 
(as compared to higher quality data with the same level of diversity) improved the scaling laws of models on downstream tasks. \citet{zhang2024instructiondiversitydrivesgeneralization} experiment with varying levels of instruction diversity when training transformers to solve algorithmic string rewriting tasks. They find, when controlling for sample count, a more diverse task set with few samples per task generalizes to new tasks better than more samples concentrated on less-diverse tasks. \citep{wei2024simplesyntheticdatareduces} observed that instruction tuning increases sycophancy, a form of reward hacking where the model tends to repeat the user's opinion even if it is objectively incorrect. To mitigate and reduce this undesired behavior, they've used simple classification datasets and augmented them with diverse opinions to teach the model that a statement’s truthfulness is independent of a user’s opinion. \citet{zhao2024understandingsyntheticcontextextension} generate synthetic data improving the ability of LLMs to perform long-context retrieval in a diverse set of contexts.

Notably, in some cases diversity inversely correlates with better performance \citep{ye2022zerogenefficientzeroshotlearning, pourcel2024acesgeneratingdiverseprogramming, bukharin2024datadiversitymattersrobust, zhang2024instructiondiversitydrivesgeneralization}.\citet{zhang2024instructiondiversitydrivesgeneralization} notes instruction imbalance can be detrimental but addressed via more diversity. For example, \citet{yue2024mammoth2scalinginstructionsweb} find both kNN and MTLD measures of instruction diversity are not good predictors of generalization when evaluated on in-distribution chat tasks. FLAN trains models on a diverse set of instructions and sees great OOD generalization to new instructions \citep{wei2022finetunedlanguagemodelszeroshot}. \citet{dong2024abilitieslargelanguagemodels} ablates fine-tuning on different domain ratios and finds there is some positive transfer at low-data amounts but limited transfer at higher amounts. They suggest diverse data is most helpful when high-quality in-distribution data is not available.

\subsection{The Effect of Complexity}

Finally, we consider the effect of data complexity on model generalization during both pre-training and post-training. Unlike quality and diversity, complexity appears to often benefit both in-distribution and OOD generalization. However, this is only true to a certain point, with excessively complex data instead harming generalization. For example, recent work by \cite{kallini2024missionimpossible} demonstrate that while language models struggle with completely random patterns, they can learn unnatural but structured patterns, albeit less efficiently than natural ones. Hence an effective training data should maintain learnable structure while introducing sufficient challenge.

\paragraph{Data complexity in pre-training} Complexity measures can be used to filter pre-training datasets. \citet{albalak2023efficient} propose an approach that adapts data-mixing proportions in an online fashion by training a multi-armed bandit algorithm on domain perplexity. The higher the perplexity, the more complex the data is for the model and the larger the information gained by learning from it. This approach achieves lower text perplexity compared to DoReMi \citep{xie2024doremi} and an unweighted baseline. However, it requires a relatively clean dataset to ensure that high perplexity is not correlated with low-quality data. For instance, \citet{wenzek2019ccnet} train a language model on Wikipedia data to compute perplexity as a measure of distance from high-quality text, removing documents that deviate too much. 

Alternative approaches focus on gradually increasing complexity during training. \citet{dubey2024llama} show that using an annealing phase — where the learning rate is lowered to zero while incorporating more complex domain data like code and mathematics — enhances performance on key benchmarks. Others \citep{pandey2024gzippredictsdatadependentscaling,sharma2020neuralscalinglawdimension,havrilla2024understandingscalinglawsstatistical} investigate how data complexity impacts pre-training, using measures like intrinsic dimension or gzip compression. They demonstrate that more complex data takes longer to learn.

% Another way of increasing complexity during pre-training is by using an annealing phase, during which the learning rate is lowered to zero while the data mix is adjusted by incorporating more complex domain data, such as code and mathematical data. \citet{dubey2024llama} show that this method enhances performance on key benchmarks. Several papers \citep{pandey2024gzippredictsdatadependentscaling,sharma2020neuralscalinglawdimension} investigate the impact of data complexity, as measured by intrinsic dimension or gzip, on pre-training scaling laws. They show more complex data takes longer to learn.

\paragraph{Data complexity in fine-tuning} \citet{zhao2024preliminarystudyintrinsicrelationship} measures complexity of instruction following datasets by parsing instructions into a semantic tree. They find fine-tuning on more difficult instructions outperforms fine-tuning on more less-difficult instructions where the number of training tokens is equalized. INSTAG \citet{lu2023instaginstructiontagginganalyzing} measures complexity as the number of attribute tags assigned to a sample by a classifier LLM. They find models fine-tuned on more complex instruction samples perform better on MT-Bench. The WizardLM model series \citep{xu2023wizardlmempoweringlargelanguage, luo2023wizardcoderempoweringcodelarge, luo2023wizardmathempoweringmathematicalreasoning, zeng2024automaticinstructionevolvinglarge} employ LLMs to automatically evolve the complexity of seed data. They see significant improvement in models fine-tuned on the resulting synthetic data. \citet{li2024quantityqualityboostingllm} measures the complexity of instruction following data using a novel Instruction Following Difficulty metric based on the ratio of the instruction's perplexity given the response to instruction's unconditional perplexity. They also find higher complexity datasets yield better instruction following results. \citet{liu2024makesgooddataalignment} measures complexity of chat data by adopting similar evolutionary approach to WizardLM. They conduct a review of several baseline complexity measures, finding their best performing data split on MT-Bench also has high complexity as measured by their proposed metric. Similarly, \citet{zhao2024preliminarystudyintrinsicrelationship} measure complexity of instruction data by decomposing samples into a tree structure and counting the size of the tree. Controlling for token count, they find training on fewer, complex instructions outperforms training on more, simple instructions. However, it is also true that training on excessively difficult data can harm model generalization. Recent work on weak-to-strong generalization \citep{sun2024easytohardgeneralizationscalablealignment} demonstrates training on lower-complexity math problems can generalize as well as, and sometimes even better than, training directly on more complex data.

\paragraph{Recap} We'll now review the main takeaways from the last three sections covering the effects of quality, diversity, and complexity on model generalization: 

\vspace{0.3cm}
\begin{takeaways}
    \setlength{\leftmargini}{0pt}
    \begin{itemize}
        \item High-quality data improves in-distribution generalization more than OOD generalization.
        \item Highly diverse data improves OOD generalization more than in-distribution generalization.
        \item Complex data, at the right levels of difficulty, can improve both in-distribution and OOD generalization.
    \end{itemize}
\end{takeaways}
\vspace{0.3cm}

These takeaways summarize the qualitative impacts of QDC on model generalization. Further work in this direction could attempt to quantify these impacts in the form of \textit{quality-diversity-complexity} scaling laws on downstream tasks. Existing scaling laws \citep{kaplan2020scalinglawsneurallanguage,hoffmann2022trainingcomputeoptimallargelanguage} can typically be used to predict pre-training loss $L(N, D, C)$ as function of model size $N$ and data size $D$ and compute size $C$. Typically lower pre-training loss also correlates with better benchmark performance. Recently accurate quantitative predictions of model benchmark capabilities as a function of size, data, and pre-training loss have been made \citep{ruan2024observationalscalinglawspredictability,openai2024gpt4technicalreport}. These predictions of may be improved further by considering the quality, diversity, and complexity of training data. Already there exist several works demonstrating the complexity of pre-training data directly influences scaling behavior \citep{sharma2020neuralscalinglawdimension,havrilla2024understandingscalinglawsstatistical,pandey2024gzippredictsdatadependentscaling}. Additionally, many questions remain surrounding maximal levels of data complexity useful for improving performance. If samples are too complex then they will likely instead harm generalization.

\vspace{0.3cm}
\begin{openquestions}
    \setlength{\leftmargini}{0pt}
    \begin{itemize}
        \item Can we develop models predicting how down-stream metrics of interest will behave as a function of training data quality, diversity, and complexity?
        \item How can we determine the \textbf{right level} of complexity to most benefit model generalization.
    \end{itemize}
\end{openquestions}
\vspace{0.3cm}

\subsection{Trade-offs in QDC and Impacts on Performance}
\label{subsec:qd_tradeoffs}

In order to reap the benefits of quality, diversity, and complexity, the ideal training dataset should simultaneously be high-quality, highly-diverse, and highly-complex. However, in practice this is not always possible due to data availability and natural trade-offs between quality and diversity. For example, this happens often during data filtering, as more aggressive filtering can improve quality but limit the amount and diversity of data \citep{longpre2023pretrainersguidetrainingdata}. As a result, higher-quality datasets tend to be less diverse and vice-versa. \citet{fan2023scalinglawssyntheticimages}
study this phenomena in the image domain by prompting text-to-image models to generate large sets of synthetic data
to train image classifiers for the Imagenet classification task \citep{russakovsky2015imagenetlargescalevisual}. Quality is measured using a pre-trained image classifier and diversity is measured by extracting intermediate embeddings from the same pre-trained model and measuring standard deviation of the embeddings across all classes. They find synthetically generated Imagenet datasets with higher-quality tend 
to be less diverse. The models trained on these high-quality, low-diversity datasets achieve good in-distribution test performance. However, they struggle when generalizing out of distribution. Conversely, they find synthetically generated image datasets with high-diversity are often lower-quality. However, models trained on more diverse datasets generalize better to OOD benchmarks (MS-COCO). This demonstrates a trade-off between quality and diversity, \textbf{with higher-quality data benefitting in-distribution generalization and higher-diversity data benefitting OOD generalization}. \citet{hashimoto2019unifyinghumanstatisticalevaluation} and \citet{holtzman2020curiouscaseneuraltext} make similar observations about trade-offs between quality and diversity in text data.

\citet{bukharin2024datadiversitymattersrobust} conducts a similar study using quality and diversity metrics to filter publicly available instruction tuning datasets. To measure quality they use both a contrastively trained reward model and ChatGPT as a judge. To measure diversity of a training subset $S' \subseteq S$, they embed instruction, response pairs using Sentence Transformer \citep{reimers2019sentencebertsentenceembeddingsusing} and compute the \textit{facility location function} of $S'$ with respect to $S$. The quality and diversity measures are then used to construct an objective function $o(s) = \alpha q(s) + (1-\alpha)d(s)$ used to greedily select instruction samples with $0 \leq \alpha \leq 1$. When the number of samples is fixed, for low values of $\alpha$ the selected datasets is highly diverse but lower quality.For high values of $\alpha$ the resulting dataset is high-quality but lacks diversity. Further, the instruction models trained on high-quality, low-diversity datasets tend to perform well over-all on both AlpacaEval \citep{zheng2023judgingllmasajudgemtbenchchatbot}. Models trained on highly-diverse but lower-quality datasets perform worse overall. However, they generalize better than models trained only on high-quality samples to the most difficult benchmark instructions. \citet{liu2024makesgooddataalignment} report similar findings for chat data, in the process comparing multiple measures of data quality, diversity, and complexity. Alpagasus \citep{chen2024alpagasustrainingbetteralpaca} also notes that filtering out lower quality samples simultaneously simultaneously limits diversity. As a result, coding quality of the fine-tuned model suffers.

\vspace{0.3cm}
\begin{takeaways}
    \setlength{\leftmargini}{0pt}
    \begin{itemize}
        \item Levels of quality and diversity naturally trade-off in data, affecting model generalization.
    \end{itemize}
\end{takeaways}
\vspace{0.3cm}

Virtually no works have investigated the trade-offs between data \textbf{complexity} and quality and diversity. Understanding this relationship is essential for predicting the performance of models trained on data with varying mixtures of quality, diversity, and complexity.

\vspace{0.3cm}
\begin{openquestions}
    \setlength{\leftmargini}{0pt}
    \begin{itemize}
        \item What is the relationship between \textbf{complexity} and quality and diversity in natural data?
    \end{itemize}
\end{openquestions}
\vspace{0.3cm}

%% file: sections/qdc_algos.tex
\section{QDC Promoting Mechanisms in Synthetic Data Algorithms}\label{sec:qd_algo}

\begin{figure}[ht]
    \centering
    \includegraphics[scale=0.18]{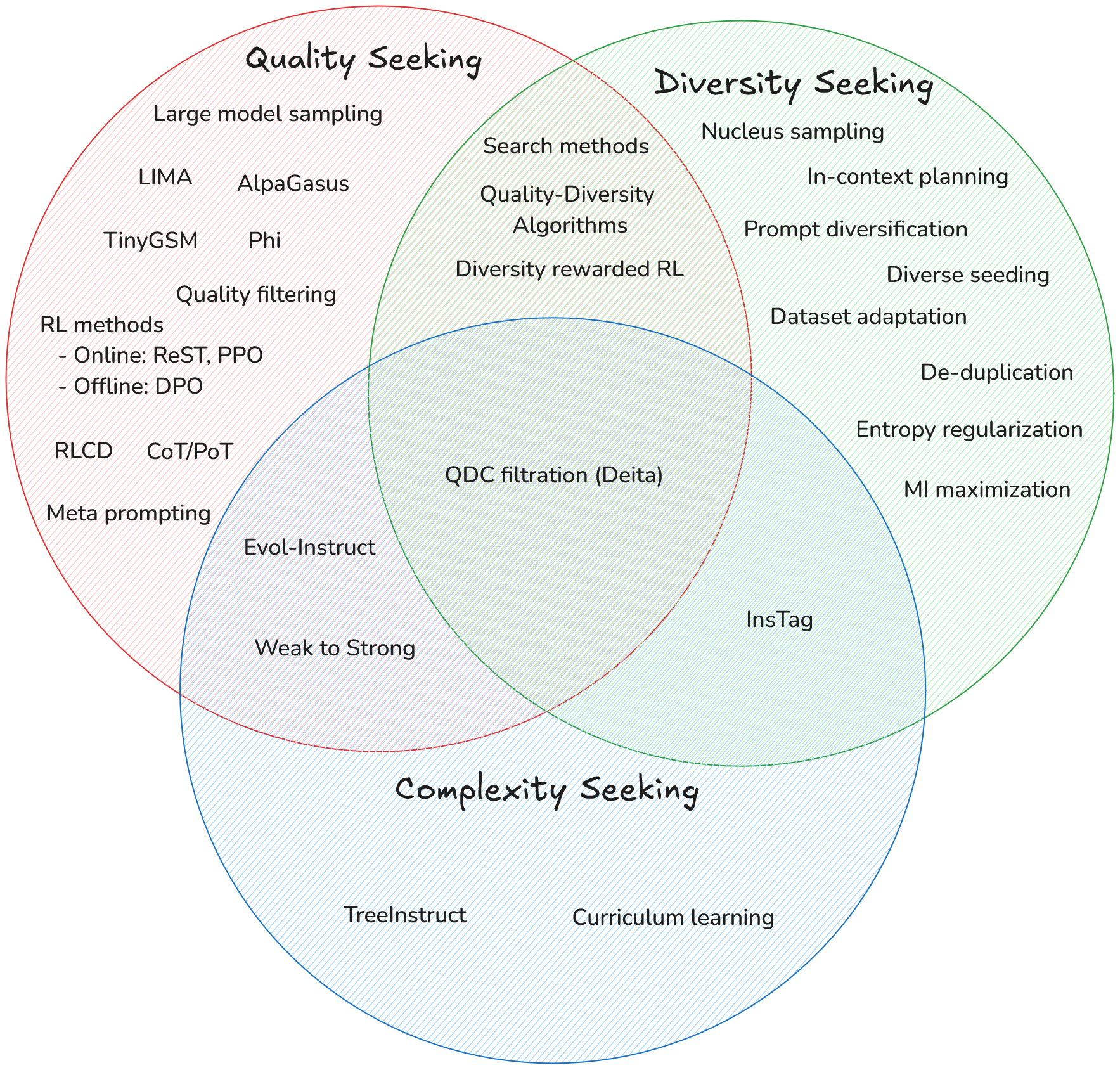}
    \caption{Venn diagram of quality, diversity, and complexity seeking  algorithms.}
    \label{fig:qdc_venn_diagram}
\end{figure}

In Section \ref{sec:qdc_measures} we defined notions of quality, diversity, and complexity and surveyed numerous implementations of each measure in the literature. In Section \ref{sec:qd_effects} we discussed how the quality, diversity, and complexity of a training dataset $\mathcal{D}$ 
impacts the generalization of models to downstream tasks. Now, in Section \ref{sec:qd_algo}, we will discuss existing methods for synthetic data generation through the lens of quality, diversity, and complexity. We do this by identifying and categorizing mechanisms for promoting the quality, diversity, and complexity of synthetic data created by synthetic data generation algorithms. This allows us to start categorizing synthetic data generation algorithms themselves as being ``quality-seeking'', ``diversity-seeking'', ``complexity-seeking'', or some mixture thereof. As in Section \ref{sec:qd_effects}, we also discuss the interplay between quality, diversity, and complexity promoting mechanisms in synthetic data generation algorithms and the resulting effects on downstream model capability. Of particular interest to us is \textbf{the impact of quality, diversity, and complexity in synthetic data when used for recursive model self-improvement}: either through reinforcement learning or other means. 

Most existing methods for synthetic data pipelines can be broken into three iterable phases: 

\begin{enumerate}
    \item Data generation
    \item Data filtration
    \item Data distillation
\end{enumerate} The dataset generation phase proceeds by sampling many responses to input prompts in training data, yielding synthetic dataset $\mathcal{D}_0$. During dataset filtration, the generated samples are filtered using a chosen objective function $O: \Omega \to \mathbb{R}$, yielding $\mathcal{D}_0' \subseteq \mathcal{D}_0$.
Finally, during dataset distillation the remaining desired synthetic data after filtration is distilled into a student model $M$, i.e., $M$ is trained on $\mathcal{D}_0'$ to  improve its performance. Note that $M$ can be either the same model used for data generation or an entirely different model.
We find the majority of approaches rely on a large, SOTA LLM to maximize answer quality during generation. 
The most common approach to maintaining diversity is through the use of a large seed-dataset of input prompts.
Even this simple combination can be effective without any filtering provided the SOTA LLM consistently produces enough high-quality solutions.
However, more sophisticated algorithms can be used to encourage better sample quality, diversity, and complexity during the entire process. Various sampling algorithms can be applied to improve the QDC of generated data. 
Complex filters can be applied to control and trade-off levels of QDC. Training procedures beyond the standard next token prediction loss can be used to improve model output quality, diversity, and complexity. 
Toward this end we highlight a small set of algorithms, largely inspired by techniques from QD literature \citep{mouret2015illuminatingsearchspacesmapping}, which encourage sampling directly maximizing both quality and diversity. Finally, we discuss the impact of the quality, diversity, and complexity on synthetic data generation capability itself resulting effects on recursive self-improvement algorithms. We highlight an apparent trade-off between model output quality and model output diversity and the importance of balancing both for optimal improvement. Further, we note that most existing models are heavily optimized for answer quality (via maximizing single response pass@1 scores) and discuss how this comes at the expense of better output diversity in synthetic data.
We finish with a discussion on how to sample and train models for simultaneously improving output quality, diversity and complexity. Note: all the mechanisms we discuss are collected in a table in Appendix Section \ref{sec:synthetic_data_tables}.

\subsection{Quality Promoting Mechanisms}

\begin{figure}
    \centering
    \includegraphics[scale=0.05]{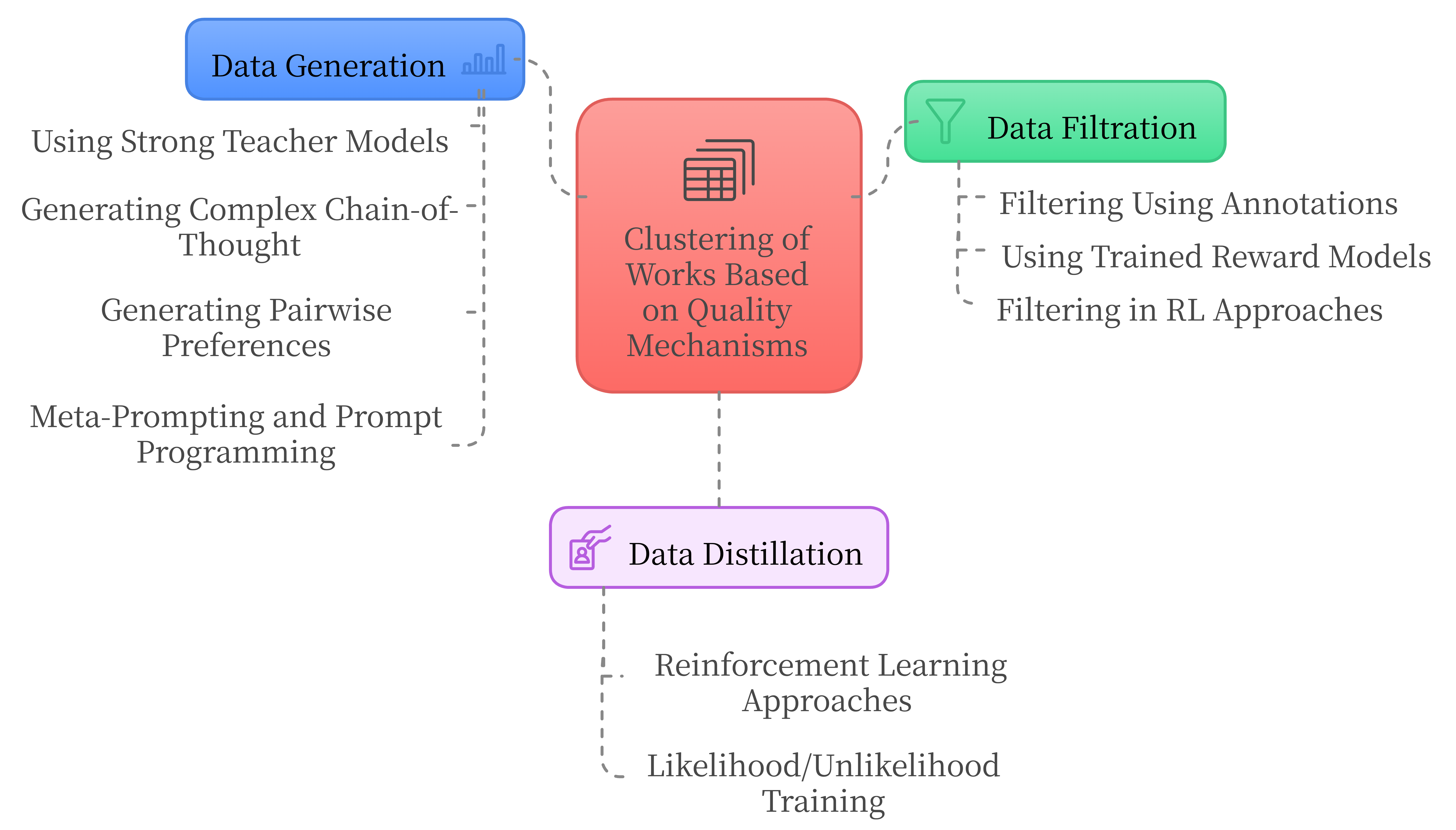}
    \caption{Mechanisms for promoting data quality in synthetic data generation.}
    \label{fig:quality_mechanisms}
\end{figure}

\vspace{0.25cm}

%\fabrizio{How to select an initial dataset of seed prompt? Vicuna used ShareGPT. How big?  How diverse?. Koala: To maintain data quality, we de-duplicated ShareGPT on the user-query level and removed any non-English conversations. This leaves approximately 30K examples from the initial and the human comparison corpus H3C dataset: https://arxiv.org/abs/2301.07597 87k question-answer examples.}
%\alexh{I moved some of these comments to more appropriate locations in the text. Can you clarify those that remain?}

\paragraph{1. Quality Mechanisms in Data Generation} One simple yet effective approach generates data using large SOTA LLMs (e.g. GPT-4), which guarantees a certain level of quality relative to weaker, less capable models \citep{mukherjee2023orcaprogressivelearningcomplex,gunasekar2023textbooks,li2023textbooks,javaheripi2023phi,abdin2024phi,vicuna2023,openai2024gpt4technicalreport, wang2022supernaturalinstructionsgeneralizationdeclarativeinstructions, honovich2022unnaturalinstructionstuninglanguage,dubey2024llama}. For example, Unnatural Instructions \citep{honovich2022unnaturalinstructionstuninglanguage} use 3 examples from Supernatural instructions and ask the pretrained model to generate a fourth. they repeat this process with 5 different seeds, generating 15 instruction examples to automatically produce 64k diverse triplets of instructions, inputs and outputs. This is an example of \textit{prompt generation} which generates both the an input prompt the LLM and the expected response. Distillation into the weaker student model can then be done via supervised fine-tuning, sometimes even with minimal data filtering. 
Alpaca, Vicuna \citep{vicuna2023}, and many derivatives take this approach by fine-tuning LLaMa \citep{touvron2023llamaopenefficientfoundation} on continuations generated with InstructGPT \citep{ouyang2022traininglanguagemodelsfollow}, ChatGPT and GPT-4. These are examples of \textit{response generation} where input prompts are sourced beforehand and only synthetic responses are generated.
The Phi model series \citep{gunasekar2023textbooks,li2023textbooks,javaheripi2023phi,abdin2024phi} and Orca \citep{mukherjee2023orcaprogressivelearningcomplex} generate complex chain of thought reasoning traces across a variety of domains using GPT-4. For example, from \citet{mukherjee2023orcaprogressivelearningcomplex}, ``Orca learns from rich signals from GPT-4 including explanation traces; step-by-step thought processes; and other complex instructions, guided by teacher assistance from ChatGPT.'' Many papers have also applied the same recipe to generate synthetic data for math and reasoning problems \citep{yu2024metamathbootstrapmathematicalquestions,liu2023tinygsmachieving80gsm8k, yue2023mammothbuildingmathgeneralist, toshniwal2024openmathinstruct118millionmath}. TinyGSM \citep{liu2023tinygsmachieving80gsm8k} generates millions of GSM8K \citep{cobbe2021trainingverifierssolvemath} style questions using GPT-4 to fine-tune a small generator and verifier. Even with minimal filtering only done on questions which have bad code syntax, the resulting models achieve nearly 70\% pass@1 accuracy on GSM8K. Using a verifier to rerank multiple solutions improves this by 12\%. 
\citet{kim2023aligninglargelanguagemodels} generate a dataset of pairwise preferences by prompting multiple models with varying parameter counts, quantities of in-context demonstrations, and qualities of in-context demonstrations. They make the assumption that larger models with higher quality and quantity of in-context examples \textit{should} generally be preferred to smaller models with fewer and lower quality demonstrations. 

Chain of thought prompting \citep{wei2023chainofthoughtpromptingelicitsreasoning} and all its derivatives \citep{chen2023programthoughtspromptingdisentangling} can be used to generate informative, higher-quality step by step responses to prompts. This type output is used in nearly all data generation schemes. \citet{reynolds2021promptprogramminglargelanguage} employ meta-prompting to generate better LLM prompts, improving quality on downstream tasks. Reinforcement learning from contrastive distillation  \citep{yang2024rlcdreinforcementlearningcontrastive} independently prompts an LLM to generate a ``good'' sample and a ``bad'' sample for a task and then trains a reward model using this synthetically generated data. Contrastive decoding \citep{li2023contrastivedecodingopenendedtext} samples text that maximizes the difference between expert log-probabilities and amateur log-probabilities, subject to plausibility constraints which restrict the search space to tokens with sufficiently high probability under the expert LM.
Another method for improving the quality of generated data is by training a critique model to generate natural language feedback on a model's response, which can then be followed by a refinement stage to improve quality~\citep{wang2023shepherdcriticlanguagemodel}.
West-of-N sampling~\citep{pace2024westofnsyntheticpreferencegeneration} draws inspiration from best-of-N sampling and defines the chosen and rejected continuations to be the best and worst N samples. One downside of this method is that it requires an initial reward model to make the initial ranking judgments.

%\fabrizio{There seems to be three subcategories of strategies here: prompt seed selection, prompt generation, and prompt completion}

%\alexh{TODO: write these categories in}

%\fabrizio{Structure filtering, the generated synthetic data does not respect the expected structure (i.e. instruction, input, output as in unnatural instructions). Syntactic filtering (i.e. using a neural  parser (i.e. berkley language parser) fail to deconstruct the sentence as in \citet(self-instruct paper.)). Semantic filtering, the content of the output does not meet a judge level information level standard. The judge can be a human or an automat.}
%\fabrizio{All these examples are semantic filtering which are the hardest to evaluate.}

\paragraph{2. Quality Mechanisms in Data Filtration} Aside from using a strong teacher model, the primary way quality is maintained during the synthetic data pipeline is through aggressive filtering with various quality metrics discussed in Section \ref{subsec:quality_measures}. Many works \citep{zhou2023limaalignment,chen2024alpagasustrainingbetteralpaca,cao2024instructionmininginstructiondata} demonstrate quality filtering synthetic data generated by strong base models can further improve performance of the student. Some filtering can be done at the \textit{syntactic level} to ensure generated outputs follow an expected structure. For example, \citet{honovich2022unnaturalinstructionstuninglanguage} filter out all samples not in prompt response format. More complex \textit{semantic filtering} can also be done assessing the content of generate samples. Several popular and effective semantic measures of quality can be used including data filtered via human ratings, contrastive reward models, outcome based reward models, or even process based reward models (see Sec \ref{subsec:quality_measures} for more quality measures).
AlpaGasus \citep{chen2024alpagasustrainingbetteralpaca} filters the synthetically generated Alpaca instruction tuning dataset from 52k samples to 9k samples via Likert-scale quality annotations from ChatGPT, with a threshold $\tau$ of 4.5, manually selected by looking at the histogram of the scores.\citet{bukharin2024datadiversitymattersrobust} filters common synthetically generated chat datasets for quality using a trained contrastive reward model. SPIN \citep{chen2024selfplayfinetuningconvertsweak} iteratively trains an instruction tuned model by generating synthetic data and prompting the model to distinguish between synthetic and human generated responses.

%\fabrizio{Self-Instruct uses Rouge-L with a threshold of 0.7 for inter-sample diversity.}

% TODO: fill out the above section more

\paragraph{Quality in Reinforcement Learning for LLMs} Additionally, nearly all RL-based approaches
for LLM fine-tuning promote quality via their choice of reward function used. Approaches including ReST \citep{gulcehre2023reinforcedselftrainingrestlanguage}, ReST$^\text{EM}$ \citep{singh2024humandatascalingselftraining}, 
Expert Iteration \citep{havrilla2024teachinglargelanguagemodels, havrilla2024glorewhenwhereimprove}, and Reward Ranked fine-tuning \citep{dong2023raftrewardrankedfinetuning, yuan2023scalingrelationshiplearningmathematical}
assume access to a fixed set of prompts and generate data by conditioning an initial LLM to generate responses on this prompt set.
On training data with a ground truth reference often exact match on the final answer. When this is not feasible, again reward models can be trained to judge correctness as is commonly done during RLHF \citep{ziegler2020finetuninglanguagemodelshuman,bai2022traininghelpfulharmlessassistant}. 
For example, ReST \citep{gulcehre2023reinforcedselftrainingrestlanguage}, which trains an LLM to do translation, alternates between a ``Grow'' step, which samples many candidate translations on a set English paragraphs and filters them using a pre-trained reward model, and an ``Improve'' step which fine-tunes the generator LLM on the filtered data. 
They use reward models to judge the quality of the translation and filter out samples below a certain threshold.

\paragraph{3. Quality Mechanisms in Data Distillation} More sophisticated RL algorithms influence model output quality both by rewarding high-quality samples and penalizing low-quality samples \citep{rafailov2024directpreferenceoptimizationlanguage, pang2024iterativereasoningpreferenceoptimization,schulman2017proximalpolicyoptimizationalgorithms,ahmadian2024basicsrevisitingreinforcestyle,setlur2024rlincorrectsyntheticdata} thus further improving model output quality. 
DPO \citep{rafailov2024directpreferenceoptimizationlanguage} formulates a contrastive objective between the logprobs of high-quality samples and low-quality samples. 
Iterative DPO \citep{pang2024iterativereasoningpreferenceoptimization, setlur2024rlincorrectsyntheticdata} iteratively samples synthetic data and trains a policy using the DPO objective.
On-policy policy-gradient algorithms such as PPO \citep{schulman2017proximalpolicyoptimizationalgorithms} and REINFORCE \citep{Williams1992, ahmadian2024basicsrevisitingreinforcestyle} can be used to directly optimize for the given reward. 
In addition to RLHF training, both PPO and REINFORCE have been successfully applied to improving LLM reasoning capability \citep{wang2024mathshepherdverifyreinforcellms,havrilla2024teachinglargelanguagemodels,sun2024easytohardgeneralizationscalablealignment,le2022coderlmasteringcodegeneration}. 
\citet{havrilla2024teachinglargelanguagemodels} finds training with outcome based reward models trained using data from the SFT policy improves RL sample efficiency. \citep{wang2024mathshepherdverifyreinforcellms,sun2024easytohardgeneralizationscalablealignment} both find RL training using dense process based rewards can lead to performance improvements over sparse terminal ground truth or model based outcome rewards. 
Also related is unlikelihood training \citep{Welleck2020Neural} which combines the standard log-likelihood objective with an ``unlikelihood'' objective penalizing undesirable tokens at each step.

%\begin{openquestions}
%    \setlength{\leftmargini}{0pt}
%    \begin{itemize}
%        \item How do response quality promoting training paradigms that explicitly reduce the likelihood of undesirable answers (like DPO) affect model output diversity?
%    \end{itemize}
%\end{openquestions}

% TODO: highlight question about how DPO/PPO improving quality via downweighting negatives impacts model diversity

% TODO,Question: how does PPO, DPO affect model output diversity?

% TODO: are there any quality mechanisms in the distillation phase?

\subsection{Diversity Promoting Mechanisms}

\begin{figure}
    \centering
    \includegraphics[scale=0.05]{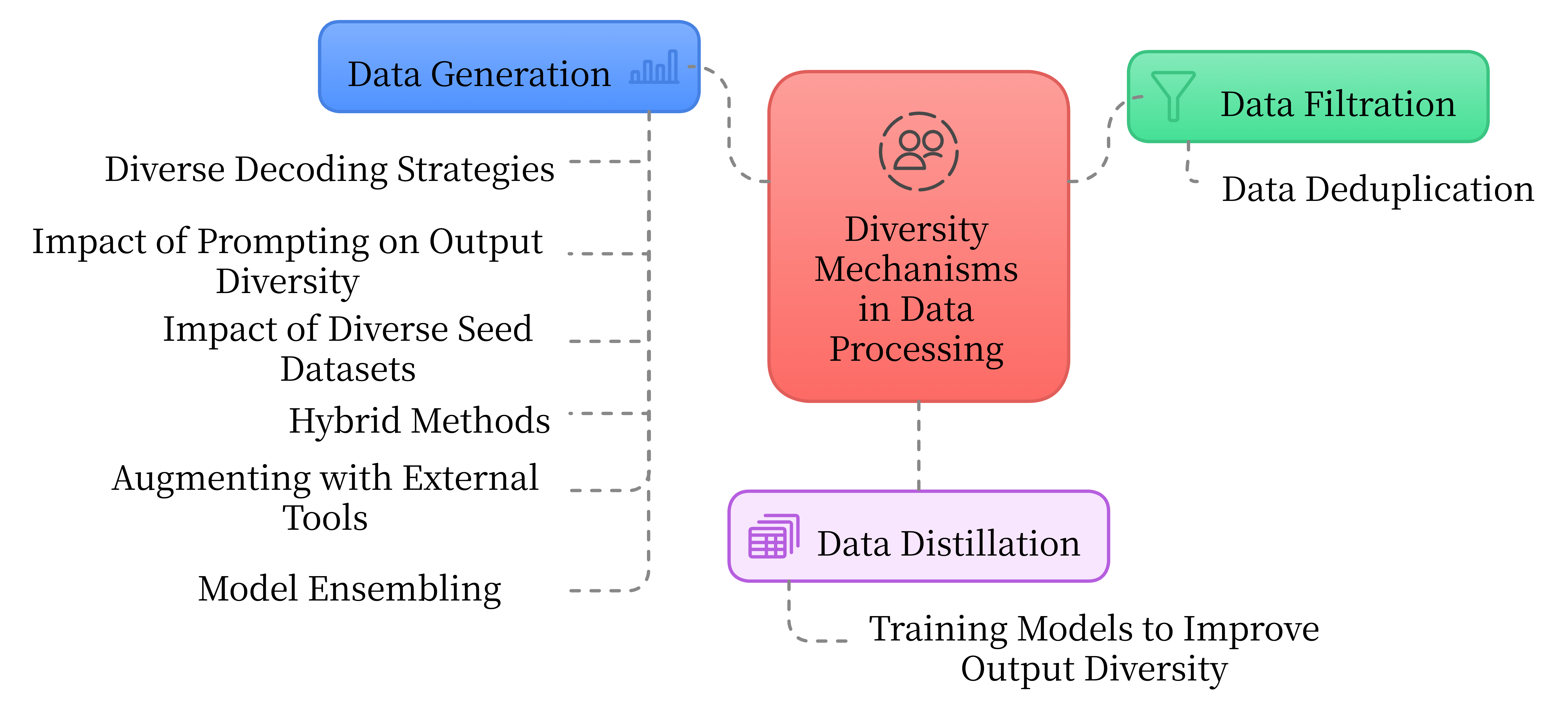}
    \caption{Mechanisms for promoting data diversity in synthetic data.}
    \label{fig:diversity_mechanisms}
\end{figure}

Of equal importance to promoting quality in synthetic data is the promotion of diversity. 
Here we survey the impact of components of synthetic data generation pipelines on data diversity.

\paragraph{1. Diversity in Data Generation} In contrast to most quality promoting mechanisms, the majority of diversity promoting mechanisms take effect during the initial data generation phase. We group these into several categories of techniques for improving diversity.

\textbf{Diverse Decoding Strategies}\quad  Changing the sampling temperature is one of the simplest ways to change model output diversity \citep{viswanathan2023prompt2modelgeneratingdeployablemodels}. \citet{havrilla2024teachinglargelanguagemodels} experimented with different sampling temperatures and rollout prompts when training using PPO. 
They found the best temperature depended on the initialization of the model being trained, with higher temperatures useful for SFT initialized models and lower-temperatures useful for prompted pre-trained models. 
\citet{ye2022zerogenefficientzeroshotlearning} generate task specific data by prompting large pre-trained models. They find varying top-k, top-p and nucleus sampling \citep{holtzman2020curiouscaseneuraltext} can also improve output diversity where diversity is measured by Self-Bleu score \citep{Welleck2020Neural,holtzman2020curiouscaseneuraltext}. 
Nucleus sampling \citep{holtzman2020curiouscaseneuraltext} is designed to reduce token repetition and promote output diversity while maintaining fidelity and coherence of the generated text. Nucleus sampling proceeds at each step by truncating the token distribution at the first $k$ tokens with cumulative probability above some chosen threshold $0 < p \leq 1$ and re-normalizing. 

\paragraph{Impact of prompting on output diversity:} Varying LLM prompts is another simple yet highly effective method of introducing more diversity into LLM outputs.\citet{jiang2024selfplanningcodegenerationlarge} prompt GPT to generate diverse plans 
in preparation to solve coding problems. 
\citet{naik2024diversitythoughtimprovesreasoning} try several prompting strategies to improve math problem solution output diversity including prompting with multiple personas (also explored in \citep{chan2024scalingsyntheticdatacreation}), working backwards, and method of elimination.
\citet{toshniwal2024openmathinstruct118millionmath} prompt for solutions to problems based on the desired subject/skills.
\citet{yu2023largelanguagemodelattributed} tries varying prompt length and prompt style.
\citet{ye2022zerogenefficientzeroshotlearning} adopts task specific prompts depending on the domain.
\citet{bradley2023qualitydiversityaifeedback} and \citet{fernando2023promptbreeder} show that maintaining and selecting for a continually changing in-context examples pool in an evolutionary fashion is important for overcoming limited output diversity from (fine-tuned) LLMs, leading to improved diversity in responses from LLMs, while maintaining/improving output quality in code generation and task prompt generation domains respectively. 
\citet{naik2024diversitythoughtimprovesreasoning} devise a multi-stage process of creating diverse prompts by first identifying appropriate approaches for solving a problem, then identifying relevant personas to imitate and finally using the cross-product of approaches and personas to generate a set of diverse candidate prompts.
Similarly,~\citet{yu2023largelanguagemodelattributed} generate diverse prompts by explicitly specifying the desired attributes (e.g.\ length and style).
\citet{yang2024rlcdreinforcementlearningcontrastive} create diversity across their chosen and rejected continuations by employing contrasting prompts. They use a positive prompt designed to encourage adhering to the given principles for the chosen continuation, and a negative prompt designed to encourage violations of the same principles.\citet{liu2023explorationprinciplesdiverseai} utilizes an LLM to act as both a data generator and critic which assesses sample correctness and novelty. Sample generation is done by mutating an existing problem with respect to a set of mutation ``principles''. The novelty of the new sample is then assessed relative to its parent.

%\vera{ Self-instruct \citet{wang2023selfinstructaligninglanguagemodels} is a method for generating data followed by a quality-diversity filtering. The purpose of the method is to generate data so that the specific instruction following capabilities of the LM will improve. Could fall in a category of quality-diversity driven algorithms for synthetic data generation, since the filtering phase removes invalid and similar samples.   

%Orca relies on a diverse seed dataset \citep{mukherjee2023orcaprogressivelearningcomplex}.
%Unnatural instructions uses paraphrasing on the instruction part with sampling and greedy decoding on output.

%We characterize synthetic data generation algorithms as diversity seeking if they explicitly encourage 
%more dataset diversity in either the generation or filtration phase. We start delineating between two categories of synthetic data generation:
%dataset generation vs. dataset adaptation.

\textbf{Impact of diverse seed datasets}\quad All synthetic data generation algorithms start from a set of initial seed prompts from which data is generated. In principle, algorithms successfully bootstrapping from a limited set of prompts are more desirable, as they are in some sense generating truly synthetic data. 
In practice however, the size of the set of seed prompts, and the manner in which they are transformed, dramatically affects diversity. 
We call algorithms which do not rely on a large set of seed prompts \textit{generation} algorithms. 
We call algorithms which do leverage large sets of seed prompts \textit{adaptation} algorithms. 
These two categories are mainly differentiated by the scale of the seed dataset.

Self-instruct~\citep{wang2023selfinstructaligninglanguagemodels} is an example of a dataset generation algorithm which generates an instruction following dataset 
from less than 500 seed prompts. These examples are used to prompt the pre-trained LLM to produce new instructions. Any low-quality or overly similar responses are then removed, and the remaining instructions are reintegrated into the task pool for further iterations.
\citet{toshniwal2024openmathinstruct118millionmath} presents a hybrid method which generates solutions to math problems in natural language and then converts them to a representation
in code. 
Metamath \citep{yu2024metamathbootstrapmathematicalquestions} prompts GPT-4 to rephrase existing seed questions in addition to generating new solutions. 
In contrast to dataset generation algorithms, dataset adaptation algorithms typically generate data using much larger seed datasets. 
Because the seed-datasets are so large, often the focus is on annotating or transforming instead of generating entirely new samples. 
Toolformer~\citep{schick2023toolformerlanguagemodelsteach} uses gpt-j~\citep{wang2021gpt} to annotate common-crawl with tool calls lowering the perplexity of subsequent tokens. 
The annotated samples are then used to construct a synthetic fine-tuning dataset. 
prompt2model \citep{viswanathan2023prompt2modelgeneratingdeployablemodels} and DataTune \citep{gandhi2024bettersyntheticdataretrieving} are given a target task and retrieve relevant publicly available fine-tuning datasets, using an LLM to adapt the retrieved data to the format of the given task. 
Several papers \citep{viswanathan2023prompt2modelgeneratingdeployablemodels, ding2023gpt3gooddataannotator, huang2022largelanguagemodelsselfimprove} use LLMs to generate labels for unlabeled datasets using self-consistency techniques. 
\citet{lanchantin2023learningreasonmemorizeselfnotes} annotates a pre-training dataset with self-notes improving the model's downstream predictions. 
\citet{yue2024mammoth2scalinginstructionsweb} scrapes ``educational'' websites for STEM instruction style data, filtering out high-quality samples using a FastText classifier and GPT-4. 
Rephrasing is then done using Qwen-72B with CoT reasoning added in when necessary. 
This broadly improves benchmark performance across the board. 
\citet{gandhi2024bettersyntheticdataretrieving} proposes a recipe for task specific dataset transformation.
The method generates new data from existing one with the aim of boosting performance for a specific task.
The newly generated data can be used for pre-training. 
\citet{agarwal-etal-2021-knowledge} synthetically generate a knowledge-enhanced corpus for language model training by verbalizing a Wikipedia knowledge graph.

Many algorithms exist which exhibit characteristics of both dataset generation and dataset adaptation algorithms. 
Cosmopedia~\citep{benallal2024cosmopedia} is a hybrid method which generates a new pre-training dataset by sampling continuations from strategically selected prompts in an existing corpus.Data Advisor \citep{wang2024dataadvisor} proposes a systematic framework where one LLM acts as an advisor to monitor dataset coverage, identify underrepresented aspects, and guide another LLM's generation of samples to fill these gaps. Backtranslation methods in machine translation~\citep{sennrich-etal-2016-improving} and Instruction backtranslation \citep{li2024selfalignmentinstructionbacktranslation} both operate on large amounts of unstructured prompt data and generate large amounts of synthetic data by backtranslating to the target domain.

\textbf{Augmenting with external tools:}\quad Integrating LLMs with tools can also be used to improve 
model output diversity. \citet{ni2024nextteachinglargelanguage} annotates program repair tasks with execution traces and
generates zero-shot critiques and refinements (improves output quality and diversity). 
\citet{gou2024criticlargelanguagemodels, lozhkov2024starcoder2stackv2} leverages a python interpreter to engage in more diverse self-corrections.

%temp 

\textbf{Search}\quad Combining LLMs with exterior search algorithms can also be used to improve model output diversity.
The most naive example of this is i.i.d sampling \citep{cobbe2021trainingverifierssolvemath}. 
Tree of thoughts \citep{yao2023treethoughtsdeliberateproblem} proposed combining generative LLMs with breadth first search and an LLM for state evaluation to improve state diversity in Game of 24, creative writing and Crosswords.
MCTS \citep{tian2024selfimprovementllmsimaginationsearching}  has also been used to in an attempt to find more diverse solutions. 
Stream of Search \citep{gandhi2024streamsearchsoslearning} demonstrates the search process can also be performed fully in-context. 
Self-reflection techniques \citep{shinn2023reflexionlanguageagentsverbal,madaan2023selfrefineiterativerefinementselffeedback} can also be seen as a type of in-context search process. \citet{havrilla2024glorewhenwhereimprove,qu2024recursive,snell2024scalingllmtesttimecompute,kumar2024traininglanguagemodelsselfcorrect} generates refinement data synthetically for reasoning tasks, demonstrating sampling more refinements can achieve better diversity than i.i.d generation. 
\citet{miao2023selfcheckusingllmszeroshot} uses LLMs to self-check solutions as process based reasoner and improve multi-sample solution accuracy with weighted voting. 
Also related is the use of LLM generated plans \citep{jiang2024selfplanningcodegenerationlarge, wang2024planningnaturallanguageimproves} in natural language which can improve the diversity of generated code. 

Several recent papers \citep{li2024common7blanguagemodels, brown2024largelanguagemonkeysscaling,snell2024scalingllmtesttimecompute,wu2024inferencescalinglawsempirical} compare the performance of various inference time strategies (parallel i.i.d sampling, Monte Carlo Tree Search, REBALANCE \citep{wu2024inferencescalinglawsempirical}, Self-reflection) when sampling large vs. small models with a fixed compute budget.
\citet{li2024common7blanguagemodels} shows that Llama-2-7B \citep{touvron2023llama2openfoundation} can obtain 72\% accuracy on MATH when i.i.d sampled 256 times despite achieving less than 20\% accuracy when sampled once.
\citet{brown2024largelanguagemonkeysscaling} finds that repeated i.i.d sampling scales log-linearly across many benchmarks (GSM8K, MATH, MiniF2f \citep{zheng2022minif2fcrosssystembenchmarkformal}).
Additionally they find sampling smaller, cheaper models can outperform sampling larger, expensive models at the same cost budget.
However, the benefit of smaller models greatly depends on problem complexity, with harder problems being less amenable to efficient sampling with smaller models. \citet{snell2024scalingllmtesttimecompute,wu2024inferencescalinglawsempirical} make similar observations, additionally comparing i.i.d sampling to more sophisticated search strategies leading to improvements in output diversity as measured by pass@n performance. 

\textbf{Model Ensembling}\quad  \citet{yuan2023scalingrelationshiplearningmathematical} shows that sampling multiple models results in more diverse samples than a sampling from a single model.
Using multiple training checkpoints of a single LLM can also improve diversity \citep{liu2023tinygsmachieving80gsm8k}. 
\citet{kim2023aligninglargelanguagemodels} generate a dataset of pairwise preferences by prompting multiple models with varying parameter counts, quantities of in-context demonstrations, and qualities of in-context demonstrations. 
They make the assumption that larger models with higher quality and quantity of in-context examples \textit{should} generally be preferred to smaller models with fewer and lower quality demonstrations.

\paragraph{2. Diversity Mechanisms in Data Filtration}As discussed in Section \ref{sec:qd_effects}, there are many notions of data diversity that can be used to filter training data. Most prominent are various forms of deduplication \citep{wenzek-etal-2020-ccnet, abbas2023semdedupdataefficientlearningwebscale} which de-bias pre-training data away from frequently occurring strings of text. Synthetic solutions to math word problems can be deduplicated by their order of operations \citep{yu2023scaling,havrilla2024teachinglargelanguagemodels} resulting in improved downstream performance. \citet{tian2024selfimprovementllmsimaginationsearching} measures semantic similarity of reasoning steps using heuristics and LLM-as-a-judge to combine semantic similar actions. \citet{lu2023instaginstructiontagginganalyzing} measures the diversity of synthetic chat data by tagging it with topics/skills using prompted LLM classifiers. 
Dataset diversity can then be measured as the number/overlap of unique tags in a dataset.

% TODO: \citep{lozhkov2024starcoder2stackv2} should talk more about synthtic data generation in code

\paragraph{3. Diversity Mechanisms in Data Distillation} While the vast majority of methods for improving model output diversity are done at inference time, some work exists attempting to train models for improved output diversity \citep{zhang2024forcingdiffusedistributionslanguage,li2024entropicdistributionmatchingsupervised,li2016diversitypromotingobjectivefunctionneural,havrilla2024teachinglargelanguagemodels}. \citet{li2016diversitypromotingobjectivefunctionneural} introduces a mutual information based objective that improves dialogue diversity as measured with Bleu score.
\citet{zhang2024forcingdiffusedistributionslanguage} trains a language model to generate factual data by directly optimizing for the frequency of each fact in a given context. 
They use this to generate a dataset of synthetic biographies which is measured to be more diverse when counting the number of unique biographical combinations.
\citet{li2024entropicdistributionmatchingsupervised} combines the standard log-likelihood loss with a regularization term maximizing response entropy.
Resulting models attain higher pass@n scores when fine-tuned on reasoning benchmarks.
\citet{havrilla2024teachinglargelanguagemodels,singh2024humandatascalingselftraining} finds that simply training on synthetic reasoning solutions with diverse solution paths outperforms the pass@n score of a SFT baseline.

\subsection{Complexity Promoting Mechanisms}

\begin{figure}
    \centering
    \includegraphics[scale=0.1]{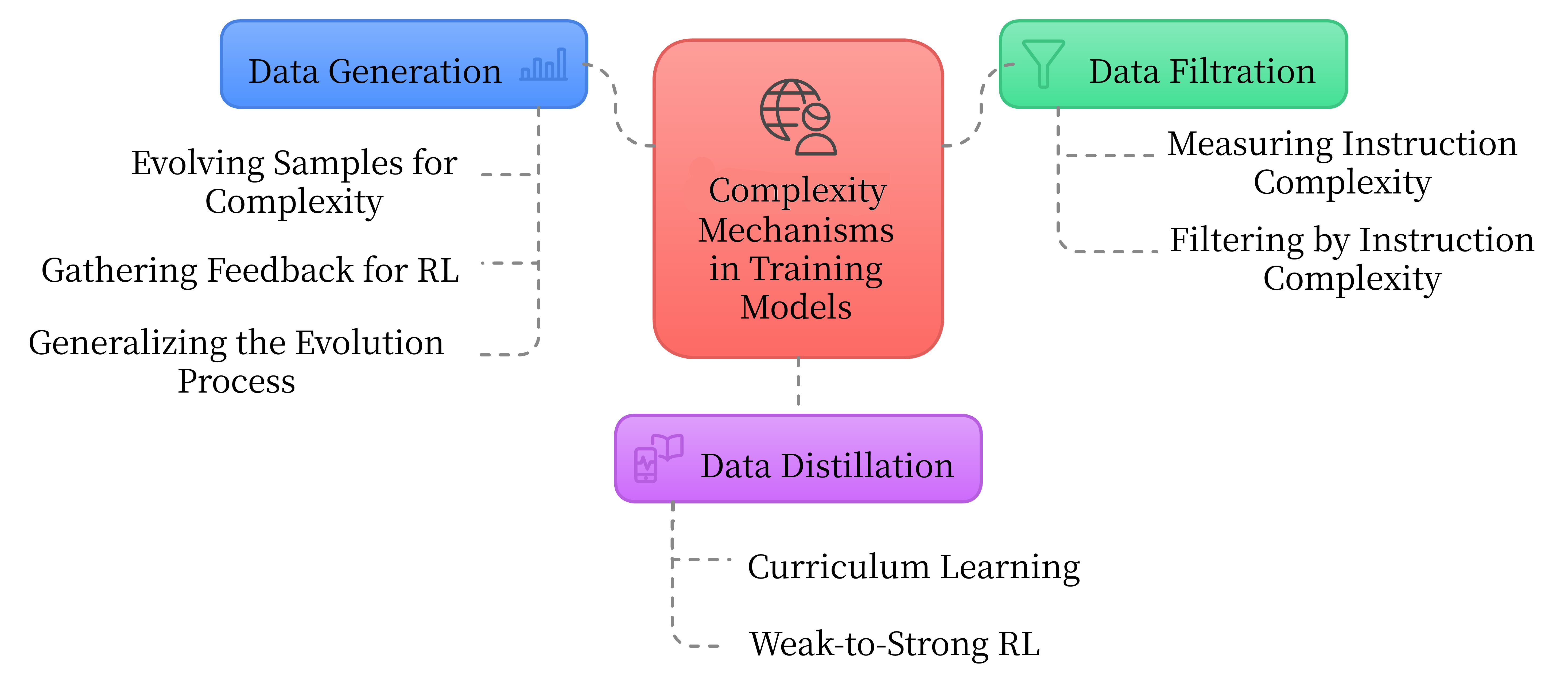}
    \caption{Mechanisms for promoting data complexity in synthetic data generation.}
    \label{fig:quality_mechanisms}
\end{figure}

Finally, we come to the impact of synthetic data generation components on data complexity.

\paragraph{1. Complexity Mechanisms in Data Generation} Perhaps one of the best examples of generating increasingly complex data using LLMs is given by the WizardLM series \citep{xu2023wizardlmempoweringlargelanguage, luo2023wizardcoderempoweringcodelarge,luo2023wizardmathempoweringmathematicalreasoning, zeng2024automaticinstructionevolvinglarge}.
\citet{xu2023wizardlmempoweringlargelanguage} prompts GPT-4 \citep{openai2024gpt4technicalreport} to ``evolve'' samples from seed instruction datasets towards both increasingly complex and increasingly simple solutions. 
Follow up works \citep{luo2023wizardcoderempoweringcodelarge,luo2023wizardmathempoweringmathematicalreasoning} apply similar ideas to generate large amounts of synthetic instruction following data in code and math domains respectively. 
In addition, \citet{luo2023wizardmathempoweringmathematicalreasoning} gathers feedback from GPT-4 to train correctness and instructiveness reward models used for RL on the generated data.
\citet{zeng2024automaticinstructionevolvinglarge} further generalizes the evolution process, evolving not just the samples but also the prompts used to guide the evolution of sample complexity.

\paragraph{2. Complexity Mechanisms in Data Filtration}\quad As covered in Section \ref{sec:qd_effects}, there are many competing measures of complexity that can be applied to filter data.
Tree instruct \citep{zhao2024preliminarystudyintrinsicrelationship} measures the complexity of instruction following samples by decomposing instructions into a semantic tree structure.
Complexity can then be measured by counting the number of nodes in the tree and filtering out data below a certain number of nodes.
InsTag \citep{lu2023instaginstructiontagginganalyzing} can also be used to measure instruction complexity by counting the number of tags describing an instruction sample. \citet{liu2024makesgooddataalignment} proposes a new measure of complexity using GPT-4 to judge multiple samples in-context at the same time. They demonstrate filtering instruction following data with this measure outperforms filteration with alternative complexity measures.

\paragraph{3. Complexity Mechanisms in Data Distillation} Curriculum learning training approaches order training data/environments in increasing levels of complexity in the hope this will allow models to generalize to more complex tasks. For example, \citet{graves2017automatedcurriculumlearningneural}, \citet{mindermann2022prioritizedtrainingpointslearnable}, \citet{albalak2023efficient}, \citet{fan2023irreduciblecurriculumlanguagemodel}, and \citet{jiang2024adaptivedataoptimizationdynamic} all apply a complexity-based curricula to language model pre-training. The underlying measure of complexity for all these methods is a variant of entropy, as measured by the language model itself.
Curriculum learning has also been used during post-training. Unlike the pre-training curricula,~\citet{lee2024instructiontuninghumancurriculum} use a curriculum that considers the difficulty of the subject matter (e.g. high school vs. college-level questions) for instruction tuning. Additionally,~\citet{chen2023skillitdatadrivenskillsframework} extract an ordered set of ``skills'' from a dataset (either instruction tuning or continued pre-training), and develop a learning curriculum based on the ordering of skill difficulty.

Weak to strong \citep{burns2023weaktostronggeneralizationelicitingstrong} is a recent LLM alignment approach aiming to supervise strong models through the use of already aligned weaker models.
\citet{sun2024easytohardgeneralizationscalablealignment} apply weak to strong ideas to training process based reward models on MATH problems. 
They find training PRMs on easier questions sometimes allows models to generalize better than training models on all available data. This provides more evidence that the appropriate level of complexity in training depends heavily on the capability of the model being trained. 

\paragraph{Recap} The past three sections have explored the impact of various components in synthetic data generation pipelines on data quality, diversity, and complexity. Quality, diversity, or complexity promoting components can be found in all three stages of the pipeline ranging from data generation to filtration to distillation.

\vspace{0.3cm}
\begin{takeaways}
    \setlength{\leftmargini}{0pt}
    \begin{itemize}
        \item Synthetic data quality, diversity, and complexity can be controlled during any of the three stages of synthetic data pipelines.
    \end{itemize}
\end{takeaways}
\vspace{0.3cm}

Given the existence of quality-diversity trade-offs discussed in Section \ref{subsec:qd_tradeoffs} we can ask similar questions about the effect of quality promoting components on synthetic data diversity levels and vice versa. For example, how will the QD composition of an algorithm sampling only for quality then filtering via deduplication for diversity compare to the QD composition of an algorithm sampling data for diversity and then filtering for quality? How will the a distillation algorithm designed to improve model output quality affect model output diversity? Answers to these questions will help us design more \textbf{\textit{efficient}} synthetic data generation algorithms capable of generating a new dataset $\mathcal{D}$ with a target number of samples and QDC composition. More efficient algorithms may require less compute or external resources (e.g. the size of a seed dataset).

\vspace{0.3cm}
\begin{openquestions}
    \setlength{\leftmargini}{0pt}
    \begin{itemize}
        \item Quantitatively, what effect do various component choices have on QDC trade-offs in the resulting synthetic data?
        \item How \textit{efficient} is one algorithm versus another in terms of the amount of compute expended to generate a target number of samples with desired QDC composition?
    \end{itemize}
\end{openquestions}
\vspace{0.3cm}

\subsection{Impacts on Recursive Self-Improvement}
\label{sec:qdc_generation}

\paragraph{QDC trade-offs in the model output distribution} Up until now, we have primarily considered synthetic data generation as a static, single iteration process: we sample data from a single frozen model $T_\theta$, filter it in some way, and then distill into another model $S_\theta$. 
Many papers (including those employing online RL for LLM training) iterate this pipeline multiple times by using the same model $M$ as both the generator $T_\theta$ and the student $S_\theta$. 
This forms the basis of iterative self-improvement. It is a significant departure from the single iteration regime as now QDC composition of training data from the previous round will directly affect the QDC composition of synthetically generated data in the next round. From the model's perspective, the model must attain good performance (i.e., output quality) while also exhibiting good exploration (i.e., output diversity) and complexity to drive gains in future rounds. In other words, the distilled model at each iteration must be capable of generating synthetic data sets with sufficient QDC to drive future self-improvement forward. Inspired by our discussion on the trade-off between quality and diversity in static datasets, the following two questions emerge:

\begin{itemize}
    \item \textbf{Q1:} \textit{How does the QDC composition of training data affect the quality, diversity, and complexity of the model output distribution?}
    \item \textbf{Q2:} \textit{Are there trade-offs between model output quality, output diversity, and output complexity?}
\end{itemize}

In a partial answer to \textbf{Q2}, indeed it seems to be the case that some trade-off between model output quality and model output diversity exists. 
For example, standard RLHF practices \citep{bai2022traininghelpfulharmlessassistant} heavily optimize models for answer quality with little to no components promoting answer diversity. 
It is often anecdotally observed RLHF models suffer from a lack of diversity when compared to their base model and even SFT counterparts. \citet{kirk2024understandingeffectsrlhfllm} carefully investigates these claims, finding RLHF models generalize better than SFT models but have worse sample diversity.\citet{korbak2022rlklpenaltiesbetter} shows that RL for LLM fine-tuning with no KL penalty minimizes a reverse KL objective with the target distribution defined by the reward function. As a result, the trained model tends to collapse to a few high-reward samples. 
However, adding in the KL penalty regularizes this behavior, preventing collapse (and therefore potentially preserving output diversity if the reference policy is diverse). However, such a penalty can also prevent improvement to output diversity/exploration capability, as the student policy may not diverge too far from the reference \citep{yang2024asymptoticslanguagemodelalignment}.

\begin{figure}
    \centering
    \includegraphics[width=0.65\linewidth]{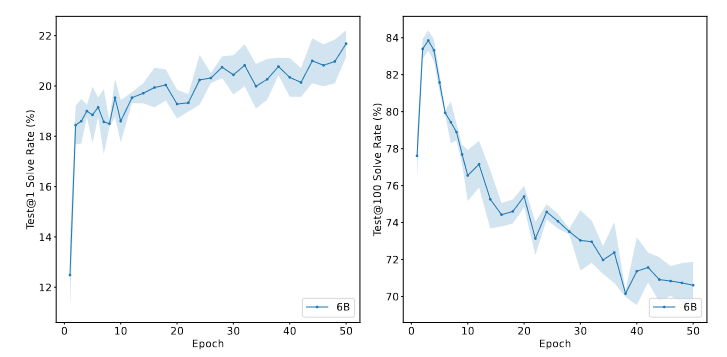}
    \caption{Trade-off between pass@1 (output quality) versus pass@100 (output diversity) throughout training.}
    \label{fig:pass_1_100_tradeoff}
\end{figure}

In the reasoning setting, \citet{cobbe2021trainingverifierssolvemath} plot the pass@1 versus pass@n performance of SFT models trained on GSM8K math word problems over several epochs. They show pass@1 and pass@100 scores both sharply increase early on in training. However, pass@n performance quickly peaks and starts decreasing while pass@1 performance continues to improve. This demonstrates another significant trade-off between model output quality and diversity. More recent works \citep{havrilla2024teachinglargelanguagemodels,singh2024humandatascalingselftraining} find similar behavior with RL fine-tuned models: pass@1 performance is greatly improved over the SFT baseline yet pass@n performance is not. However, \citep{havrilla2024teachinglargelanguagemodels} also finds that simply training on more diverse data improves the pass@n and diversity of solutions of the resulting models. This suggests that the output quality and diversity of models can be controlled in part by the composition of quality and diversity in training data.

\paragraph{Better benchmarking of model QDC output distribution} A better understanding of the relationship between model output quality and model output diversity is fundamental to the development efficient and effective iterative self-improvement algorithms. 
Models optimizing too heavily for diversity will fail to improve performance at the target task.
Models optimizing too heavily for quality will struggle to explore diversely and quickly converge in the self-improvement process. 
However, we observe that the vast majority of model benchmarking work focuses almost exclusively on evaluating answer quality \citep{hendrycks2021measuringmathematicalproblemsolving,hendrycks2021measuringmassivemultitasklanguage,cobbe2021trainingverifierssolvemath,rein2023gpqagraduatelevelgoogleproofqa,pang2024iterativereasoningpreferenceoptimization} by evaluating against a single expected answer. Comparatively few benchmarks exist which are equipped to evaluate solution diversity \citep{boratko2020protoqaquestionansweringdataset}. This disparity is likely due to the relative difficulty of checking for solution quality (checking for a single expected answer is easy) versus solution diversity. The majority of self-improvement/RL approaches also only reward for answer quality \citep{gulcehre2023reinforcedselftrainingrestlanguage,bai2022traininghelpfulharmlessassistant,havrilla2024teachinglargelanguagemodels,sun2024easytohardgeneralizationscalablealignment,singh2024humandatascalingselftraining}. However, considering the importance of model output diversity for synthetic data generation and self-improvement capabilities, evaluation of model output diversity is also essential. Since the majority of benchmarks and self-improvement algorithms are designed only to optimize only for answer quality it is likely that answer diversity suffers. This emphasizes the need for better benchmarking of both model output diversity and the balance between model output quality and diversity.

% Taken together, the above work suggests that model \textit{self-improvement capability} and \textit{synthetic data generation capability} should be essential metrics in model evaluation benchmarks. Such a benchmark would fix an initial seed dataset and compute budget and then evaluate the generalization of a trained model $M_\theta$ after the budget is exhausted. 

\vspace{0.3cm}
\begin{takeaways}
    \setlength{\leftmargini}{0pt}
    \begin{itemize}
        \item Model output quality and model output diversity trade-off.
        \item Many current models are benchmarked and optimized only for quality with output diversity likely suffering as a result.
    \end{itemize}
\end{takeaways}

Additionally, there are very few works investigating the role of complexity in iterative self-improvement algorithms for LLMs. As in Section \ref{subsec:qd_tradeoffs}, better understanding the relationship between model output complexity and quality and diversity will be essential for future algorithm development.

\vspace{0.3cm}
\begin{openquestions}
    \setlength{\leftmargini}{0pt}
    \begin{itemize}
        \item How can we better benchmark model output diversity?
        \item How can we design self-improvement algorithms balancing output quality and diversity for optimal improvement?
        \item What is the relationship between model output complexity and quality and diversity?
    \end{itemize}
\end{openquestions}
\vspace{0.3cm}

%\alexh{Sec 3.5 should tie the self-improvement setting back to reinforcement learning, which is what these methods are effectively doing}

%\elliot{Ideally, each Takeaway would also be evident from a figure/table.}

% TODO: I will have to talk about RL in this section
% Will also need to look at other self-improvement works such as self-play fine-tuning (SPIN)
% Maybe some others

\subsection{QDC-Aware Synthetic Data Generation Algorithms}

% TODO: tradeofs between quality, diversity, and complexity?
% how does complexity fit into the picture?

In the previous section we identified a trade-off between model output quality and model output diversity analogous to the trade-offs in training data from Section \ref{subsec:qd_tradeoffs}. As a result certain components of the synthetic data generation which promote quality may worsen dataset diversity and vice-versa. 
Most algorithms fail to take this trade-off into account. Further, very little research has been done investigating the synergistic and antagonistic effects of various pipeline components on dataset quality, diversity and complexity. However, a small set of synthetic data generation algorithms do attempt to directly optimize for both characteristics during the generation, filtration, or distillation phases. We call such algorithms \textit{quality-diversity synthetic data generation algorithms}. Several of these algorithms are designed using tools from the quality-diversity and novelty search search literature \citep{lehman2022evolutionlargemodels,ding2024qualitydiversityhumanfeedback,bradley2023qualitydiversityaifeedback,samvelyan2024rainbowteamingopenendedgeneration,chan2024scalingsyntheticdatacreation}.

\vspace{0.3cm}
{\textbf{1. QDC Mechanisms in Data Generation}}\quad One of the most widely known QD algorithms is MAP-Elites \citep{mouret2015illuminatingsearchspacesmapping}.
Given a sample space $\Omega$, MAP-Elites describes each $\omega \in \Omega$ via a set of designed behavioral descriptors $T_1,...,T_k:\Omega \to \mathbb{N}$. 
These are used to discretize the sample space $\Omega$ into a hyper-grid. 
The algorithm's goal is then to efficiently explore the grid (thereby maximizing diversity) while also selecting for high-quality solutions according to a pre-defined quality objective $Q$. 
New solutions are sampled from old-solutions via a pre-defined mutation operator.
By mutating increasingly high-quality solutions across many different parts of the grid MAP-Elites simultaneously attempts to simultaneously maximize sample quality and sample diversity. 
Evolution through Language Models (\textbf{ELM})~\citep{lehman2022evolutionlargemodels} was one of the first works to combine MAP-Elites with LLMs acting as a sophisticated mutation operator. 
They apply the LLM mutation driven algorithm to a synthetic robotic racing task where various \textit{sodaracer} configurations are defined programmatically.
The LLM is used to efficiently mutate programmatic configurations, improving over less sophisticated mutation baselines. 
Since the task is synthetic various notions of behavioral diversity are defined manually.

Quality-Diversity through Human Feedback (\textbf{QDHF}) \citep{ding2024qualitydiversityhumanfeedback} proposes to learn a sample level diversity function $D$ by collecting human preference triples $(y_1, y_2,y_3, y_1 \sim y_i)$ where $y_1 \sim y_2$ indicates $y_1$ is more similar to $y_2$ than to $y_3$. 
The diversity model is then trained with a contrastive objective on these preference triplets. 
MAP-Elites is then used to discover diverse and high-quality samples across several tasks. 
QDAIF \citep{bradley2023qualitydiversityaifeedback} extends the applicability of MAP-Elites to text generation domains by utilizing LLMs to both mutate and evaluate text through natural language AI feedback.
Rainbow Teaming~\citep{samvelyan2024rainbowteamingopenendedgeneration} applies MAP-Elites to generate adversarial red-teaming prompts for LLMs, using LLMs as a judge to evaluate responses.
ACES \citep{pourcel2024acesgeneratingdiverseprogramming} proposes a goal conditioned version of MAP-Elites for python program generation, utilizing LLMs to describe the relevant programming techniques used in a solution.
QDGS: Quality diversity generative sampling is used to repair biases in classifiers and improve diversity \citep{chang2024qualitydiversitygenerativesamplinglearning}. \citet{zhou2024balancingdiversityriskllm} explores varying decoding strategies at inference time to balance quality and diversity.

\vspace{0.3cm}
{\textbf{2. QDC Mechanisms in Data Filtration}}\quad The above QDC generation algorithms focus on controlling both quality and diversity during the data generation phase. 
A number of recent papers also investigate balancing synthetic dataset quality and diversity in the data filtration phase. 
InsTag measures both dataset diversity and complexity via LLM generated attributes \citep{lu2023instaginstructiontagginganalyzing}. 
ChatGPT is used to tag instructions by intention for each label, after which tags are combined if they are sufficiently similar. 
Dataset diversity is then measured by the diversity of tags occurring in the dataset.
Complexity of a sample is measured by the number of tags it has.
%% the previous sentences seems are a repetition of the previous sections.
They then filter publicly available chat datasets using by selecting to produce complex, highly diverse fine-tuning data.
\citet{liu2024makesgooddataalignment} proposes new quality and complexity measures using in-context multi-sample comparison.
They compare a number of quality, diversity, and complexity metrics on chat data, finding attribute based measurements coming from LLMs perform best.

\vspace{0.3cm}
{\textbf{3. QDC Mechanisms in Data Distillation}}\quad \cite{cideron2024diversityrewardedcfgdistillation} propose diversity-rewarded CFG distillation, which combines a distillation objective for maintaining generation quality, with a reinforcement learning objective that explicitly rewards diversity across generations. Their approach first distills classifier-free guidance (CFG) into model weights to preserve quality without the typical inference overhead, and then uses a diversity reward based on embedding similarity to encourage varied outputs. They further demonstrate that interpolating between models trained with different diversity coefficients creates a controllable quality-diversity front at deployment time. These results suggest that QDC-aware distillation strategies can effectively balance quality and diversity while reducing computational costs compared to traditional inference-time techniques. In a similar direction, \citep{omura2024entropycontrollabledirectpreference} train an LLM to dynamically select its own sampling temperature using DPO.

\vspace{0.3cm}
\begin{takeaways}
    \setlength{\leftmargini}{0pt}
    \begin{itemize}
        \item QDC synthetic data algorithms try to directly balance the quality and diversity of synthetic data.
    \end{itemize}
\end{takeaways}
\vspace{0.3cm}

While there exist several quality-diversity inspired algorithms for synthetic data generation, rarely is complexity simultaneously taken into account as a third important factor. We highlight this gap as an important open question.

\vspace{0.3cm}
\begin{openquestions}
    \setlength{\leftmargini}{0pt}
    \begin{itemize}
        \item How can we design synthetic data algorithms jointly balancing quality, diversity, \textbf{and complexity}?
    \end{itemize}
\end{openquestions}

%% file: sections/alternative_tasks.tex
\section{QDC Synthetic Data Algorithms Outside Common LLM Tasks}
\label{sec:alternative_data_generation}

%\andrew{What is the key takeaway we want to share here? Is this section looking to diverge slightly from the title/main LLM line, or broader foundation models line (e.g. with QDHF/QDGS with image gen components)? A few papers referenced here do use LLMs, but some are not based on LLMs at all (e.g. \citep{yu2023scaling, chen2023genaug, mandlekar2023mimicgen, lu2023syntheticexperiencereplay} \citep{fontaine2021quality,fontaine2021importance,openendedlearningteam2021openendedlearningleadsgenerally,fontaine2021illuminatingmariosceneslatent,zhang2024arbitrarily}). Sharing more works in the OE/QD space is always something I'm glad to see, at the same time I'm sure the reader navigating a lengthier survey could appreciate hand-holding and scoping/train-of-thought.}

%\alexh{This section is meant to be a sampler for QDC style generation outside of common LLM tasks (e.g. math, reasoning, code, etc). So not necessarily utilizing LLMs. But I think it's nice to have something to make it clear these ideas have been around for a while}

Most of the papers covered in the prior section focus on popular LLM benchmark tasks, e.g., question answering, instruction following, chat, math, and code. However, there are many of tasks outside this standard set to which LLM driven evolutionary/quality-diversity inspired algorithms have been successfully applied. We highlight several representative works in this section to illustrate the broad applicability and effectiveness of these methods \citep{chen2023evopromptinglanguagemodelscodelevel,Romera-Paredes2024,lu2023syntheticexperiencereplay,fontaine2021illuminatingmariosceneslatent,wang2023voyageropenendedembodiedagent,openendedlearningteam2021openendedlearningleadsgenerally}.

One particularly exciting application of generative models game environment design \citep{fontaine2021illuminatingmariosceneslatent,zhang2023omni,sudhakaran2024mariogpt}. \citet{fontaine2021illuminatingmariosceneslatent} and later \citet{sudhakaran2024mariogpt} generate synthetic Mario levels to train a Mario RL policy. Omni-Epic \citep{zhang2023omni} generates a diverse, open-ended set of learning environments for RL agents using code-writing LLMs. These environments are used to teach agents increasingly complex skills by designing a curriculum of tasks tailored specifically the agent's abilities. Similar to \citet{samvelyan2024rainbowteamingopenendedgeneration}, quality-diversity search can be used to automatically red-team RL pre-trained agents~\citep{bhatt2022deep,samvelyan2024multi}. For example, \citet{samvelyan2024multi} show MAP-Elites can be used to discover exploitations in a strong soccer playing RL policy. Quality-diversity and synthetic data generation algorithms have also found many applications in robotics \citep{fontaine2021quality,fontaine2021importance,yu2023scaling, chen2023genaug, mandlekar2023mimicgen, lu2023syntheticexperiencereplay}. Several papers utilize quality-diversity approaches to generate human-robot interaction scenarios~\citep{fontaine2021quality,fontaine2021importance}. Many other works \citep{yu2023scaling, chen2023genaug, mandlekar2023mimicgen, lu2023syntheticexperiencereplay} generate synthetic data to train robotics policies in a variety of domains. 

LLM driven evolutionary algorithms can also be used to solve complex, open-ended scientific optimization problems \citep{Romera-Paredes2024,chen2023evopromptinglanguagemodelscodelevel,nasir2023llmatic}. FunSearch \cite{Romera-Paredes2024} presents an evolutionary algorithm for producing programmatic solutions to combinatorics problems by sampling a small coding LLM. They show that, with enough sampling, novel solutions representing mathemematically interesting counter-examples can be found for multiple open problems. \citet{chen2023evopromptinglanguagemodelscodelevel} proposes to run an evolutionary neural architecture search using a coding LLM to produce network architectures in Pytorch. They are able to produce architectures competitive with SOTA methods while containing a fraction of the parameters in SOTA models. \citet{nasir2023llmatic} shows the neural architecture search process can be improved using quality-diversity search. \citet{zhang2024arbitrarily} evolve Neural Cellular Automata (NCA) environment generators. PromptBreeder \citep{fernando2023promptbreederselfreferentialselfimprovementprompt} and APE \citep{zhou2023largelanguagemodelshumanlevel} evolve improved LLM prompts across a wide-range of tasks.

Finally, we highlight a set of papers motivating the development of and integration LLMs in increasingly open-ended problem solving environments. The ability of LLMs to explore and learn in such environments is (almost definitionally) necessary for the development of generally capable intelligence. Voyager \citep{wang2023voyageropenendedembodiedagent} proposes a complex system deploying GPT-4 in minecraft via API-based action interaction. Their system utilizes several key components including a continuously updated skill library, a planning module proposing candidate tasks, and a self-repair module attempting to fix malformed action code. They show the resulting agent significantly outperforms RL based agents \citep{hafner2024masteringdiversedomainsworld} in skill acquisition and open-ended world exploration. \citet{openendedlearningteam2021openendedlearningleadsgenerally} discusses the essential role of open-ended learning in the development of more sophisticated, generally intelligent AI systems.

%% file: sections/conclusion.tex
\section{Conclusions and Open Questions}
\label{sec:conclusions}

\paragraph{Conclusions} Over the course of this survey we examined the composition of synthetic data in terms of its quality, diversity, and complexity and the resulting effects on model generalization when used for training. We began in Section \ref{sec:qdc_measures} by defining at a high-level what is meant by quality, diversity, and complexity in data. We then reviewed and categorized numerous practical measures of quality, diversity, and complexity in the literature. Overall, we found that domain specific, attribute measures utilizing LLMs-as-a-judge provide the best measures in complex tasks and domains in terms of correlation with downstream metrics. However, often these approaches require specific domain knowledge to implement effectively. In Section \ref{sec:qd_effects} we looked at the effects of different QDC compositions in training data (as measured by implementations in Section \ref{sec:qdc_measures}). We concluded that high-quality data primarily benefits in-distribution generalization, diverse data primarily benefits OOD generalization, and appropriate levels of data complexity can benefit both. In addition, there is a natural trade-off between high-quality and highly-diverse data. As a result, there can be trade-offs for in-distribution and OOD generalization. 

In Section \ref{sec:qd_algo} we taxonomized synthetic data generation algorithms in terms of the components each use to promote synthetic data quality, diversity, and complexity. Many algorithms promote quality by sampling from a large, SOTA LLM and maximize diversity by expanding the dataset of seed prompts. Very few algorithms explicitly consider trade-offs between quality, diversity, and complexity or attempt to maximize all three together. We then investigated the effect of training data QDC composition on the QDC of resulting synthetic data. We found a trade-off between model output quality and output diversity, resulting in higher-quality but less diverse synthetic data. Further, we observe that the majority of models today are evaluated almost exclusively for output quality, thereby limiting output diversity and the diversity of generated synthetic data. We argue that future algorithms for synthetic data generation and self-improvement should carefully trade-off data QDC composition and highlight a number of works going in this direction.

%\begin{itemize}
%    \item Quality tends to in-distribution generalization.
%    \item Diversity tends to improve OOD generalization.
%    \item Appropriate levels of complexity can benefit both in-distribution and OOD generalization.
%    \item Domain specific measures of quality, diversity, and complexity often correlate better with downstream model performance.
%    \item Most common synthetic data generation algorithms inject quality through SOTA LLM sampling and diversity through large seed datasets.
%    \item Quality and diversity often naturally trade-off both in naturally occurring data and in model output (i.e. synthetic data).
%    \item Quality-Diversity methods offer more sophisticated methods to generating high-quality, highly-diverse data.
 %   \item Balacing quality, diversity, and complexity is crurical for better self-improvement algorithms.
%\end{itemize}

\paragraph{Open Questions} Our discussions on QDC in synthetic data highlight many open questions and potential directions for future research. Most important is the development of new synthetic data generation algorithms for self-improvement that optimally trade-off QDC composition both in training data and model output distribution. Development of such algorithms will also require better benchmarking of model output \textit{diversity} and \textit{complexity} in addition to quality. Better benchmarking will allow researchers to identify the frontier of model output QDC trade-offs and find techniques moving the entire frontier forward. In particular, in comparison to quality and diversity, complexity in both data and model output has been understudied. Better understanding of what factors affect data/model complexity and how complexity is affected by quality and diversity will be crucial to future algorithm design. Finally, development of better measures of quality, diversity, and complexity will be necessary as tasks become increasingly complex and open-ended. Ideally these measures should be general purpose, inexpensive to compute, and correlate well with downstream metrics of interest.

%\begin{itemize}
%    \item How to approriately evaluate synthetic data generation with respect to QDC metrics?
%    \item How to measure appropriate level of complexity for a model?
%    \item Impact of negative sample training on output diversity?
%    \item A clearer understanding of which measures of QDC correlate with desired downstream performance
%    \item more emphasis on evaluating model diversity in addition to model quality. More holisitc evaluation
%    \item How does complexity trade-off with quality and diversity?
%    \item How can we design algorithms balancing output quality, diversity, and complexity?
%    \item How to develop domain agnostic measurs of quality, diversity, and complexity capable of learning domain specific features of interest?
%\end{itemize}

%Another point worth discussing in the conclusion is whether these take-aways can also apply to other modalities, and if not, where the gap lies. Also we did not discuss the relative costs of synthetic data generation algorithms.

\paragraph{Acknowledgements} Many, many thanks to Minqi Jiang, Mikayel Samvelyan, Alex Bukharin, and Reshinth Adithyan for their helpful feedback on earlier versions of the survey. Additionally, thank you to Hailey Schoelkopf, Louis Castricato, and Yasin Abbasi Yadkori for enlightening conversations about the role of quality, diversity, and complexity in LLMs!

%% file: sections/appendix.tex
\appendix

\section{Table of QDC Metrics}
\label{sec:qdc_metrics_table}

\begin{table}[htbp]
  \centering
  \caption{Clustering of Works Based on Quality Metrics}
  \label{tab:metrics_quality_clustering}
  \begin{tabular}{>{\centering\arraybackslash}p{3cm} p{5cm} p{7cm}}
    \toprule
    \textbf{Category} & \textbf{Subcategory} & \textbf{Works} \\
    \midrule
    \textbf{Ground Truth} & Computational Correctness & \citep{yu2024metamathbootstrapmathematicalquestions} , \citep{toshniwal2024openmathinstruct118millionmath} , \citep{singh2024humandatascalingselftraining} \\
    \cline{2-3}
    & Code Correctness & \citep{pourcel2024acesgeneratingdiverseprogramming} \\
    \cline{2-3}
    & N-gram overlap metrics such as BLEU & \citep{papineni2002bleu}, \citep{samvelyan2024rainbowteamingopenendedgeneration} \\
    \cline{2-3}
    & MAUVE & \citep{ye2022progenprogressivezeroshotdataset} \\
    \midrule
    \textbf{Reward Modeling} & Bradley–Terry & \citep{ziegler2020finetuninglanguagemodelshuman}, \citep{ouyang2022traininglanguagemodelsfollow} \\
    \cline{2-3}
    & Outcome-Based & \citep{uesato2022solvingmathwordproblems} \\
    \cline{2-3}
    & Process-Based  & \citep{lightman2023letsverifystepstep} \\
    \midrule
    \textbf{Attribute-Based} & HHH (Helpfulness, Honesty, and Harmlessness) Comparison Evaluations & \citep{bai2022traininghelpfulharmlessassistant} \\
    \cline{2-3}
    & Comparison Evaluations for Summarization & \citep{stiennon2022learningsummarizehumanfeedback} \\
    \cline{2-3}
    & Constitutional AI Critiques & \citep{bai2022constitutionalaiharmlessnessai} \\
    \cline{2-3}
    & AI-Generated Comparisons for Categories of Human Intuitions about Data Quality & \citep{wettig2024quratingselectinghighqualitydata} \\
    \cline{2-3}
    & Quality Increase through Evolution & \citep{wettig2024quratingselectinghighqualitydata} \\
    \cline{2-3}
    & UniEval & \citep{zhong2022towards} \\
    \midrule
    \textbf{Other} & Student Solve Rate & \citep{havrilla2024teachinglargelanguagemodels} \\
    & Downstream Performance Increase & \citep{zhou2023limaalignment}, \citep{viswanathan2023prompt2modelgeneratingdeployablemodels}, \citep{gandhi2024streamsearchsoslearning} \\
    \cline{2-3}
    & Binary Classification (High-Quality vs Unfiltered Data) & \citep{brown2020languagemodelsfewshotlearners}, \citep{gao2020pile800gbdatasetdiverse}, \citep{xie2023dataselectionlanguagemodels} \\
    \cline{2-3}
    & Perplexity & \citep{wenzek-etal-2020-ccnet}, \citep{sharma2024textqualitybasedpruningefficient} \\
    \bottomrule
  \end{tabular}
\end{table}

\begin{table}[htbp]
  \centering
  \caption{Clustering of Works Based on Diversity Metrics}
  \label{tab:metrics_diversity_clustering}
  \begin{tabular}{>{\centering\arraybackslash}p{3cm} p{5cm} p{7cm}}
    \toprule
    \textbf{Category} & \textbf{Subcategory} & \textbf{Works} \\
    \midrule
    \textbf{Lexical} & Inter-Sample N-gram Frequency & \citep{yu2023largelanguagemodelattributed} \\
    \midrule
    \textbf{Attribute-Based} & Number of Unique Task Types & \citep{wang2022supernaturalinstructionsgeneralizationdeclarativeinstructions} \\
    \cline{2-3}
    & Number of Unique Languages & \citep{wang2022supernaturalinstructionsgeneralizationdeclarativeinstructions} \\
    \cline{2-3}
    & Number of Unique Domains or Topics & \citep{wang2022supernaturalinstructionsgeneralizationdeclarativeinstructions} \\
    \cline{2-3}
    & Number of Unique Root Verbs and Nouns & \citep{li2024selfalignmentinstructionbacktranslation} \\
    \cline{2-3}
    & Skills Used to Arrive at a Solution & \citep{pourcel2024acesgeneratingdiverseprogramming} \\
    \cline{2-3}
    & Order of Mathematical Operations & \citep{tian2024selfimprovementllmsimaginationsearching}, \citep{havrilla2024teachinglargelanguagemodels} \\
    \cline{2-3}
    & Attribute Labels & \citep{zhou2023limaalignment}, \citep{wang2022supernaturalinstructionsgeneralizationdeclarativeinstructions} \\
    \cline{2-3}
    & MAP-Elites Coverage & \citep{samvelyan2024rainbowteamingopenendedgeneration}, \citep{bradley2023qualitydiversityaifeedback} \\
    & Automatically Discovered Attributes & \citep{yu2023largelanguagemodelattributed} \\
    \midrule
    \textbf{Embedding} & Average Pairwise Cosine Similarity & \citep{yu2023largelanguagemodelattributed}, \citep{kirk2024understandingeffectsrlhfllm}, \citep{yu2024metamathbootstrapmathematicalquestions} \\
    \cline{2-3}
    & i-th Nearest Neighbor & \citep{cao2024instructionmininginstructiondata} \\
    \cline{2-3}
    & SemDeDup & \citep{abbas2023semdedupdataefficientlearningwebscale} \\
    \cline{2-3}
    & Facility Location Function & \citep{reimers2019sentencebertsentenceembeddingsusing}, \citep{bukharin2024datadiversitymattersrobust} \\
    \midrule
    \textbf{Other} & Diversity Coefficient for Natural Language & \citep{lee2023scalediversitycoefficientdata} \\
    \cline{2-3}
    & NLI Diversity & \citep{kirk2024understandingeffectsrlhfllm} \\
    \cline{2-3}
    & MTLD & \citep{mtld} \\
    \cline{2-3}
    & N-gram Frequency & \citep{li-etal-2016-diversity} \\
    \bottomrule
  \end{tabular}
\end{table}

\begin{table}[htbp]
  \centering
  \caption{Clustering of Works Based on Complexity Metrics}
  \label{tab:metrics_complexity_clustering}
  \begin{tabular}{>{\centering\arraybackslash}p{3cm} p{5cm} p{7cm}}
    \toprule
    \textbf{Category} & \textbf{Subcategory} & \textbf{Works} \\
    \midrule
    \textbf{Attribute-Based} & Semantic Node Count & \citep{zhao2024preliminarystudyintrinsicrelationship} \\
    \cline{2-3}
    & Number of Intention and Semantics Tags & \citep{lu2023instaginstructiontagginganalyzing} \\
    \cline{2-3}
    & Direct scoring & \citep{chen2024alpagasustrainingbetteralpaca} \\
    \cline{2-3}
    & Difficulty Level & \citep{hendrycks2021measuringmathematicalproblemsolving} \\
    \cline{2-3}
    & Evolution Complexity & \citep{liu2024makesgooddataalignment} \\
    \midrule
    \textbf{Other} & Instruction Following Difficulty & \citep{li2024quantityqualityboostingllm} \\
    \cline{2-3}
    & Parse Tree Complexity & \citep{sharma2024textqualitybasedpruningefficient} \\
    \cline{2-3}
    & Perplexity & \citep{albalak2023efficient} \\
    \cline{2-3}
    & Instruction Length & \citep{liu2024makesgooddataalignment} \\
    \bottomrule
  \end{tabular}
\end{table}

\FloatBarrier
\newpage

\section{Table of QDC Mechanisms in Synthetic Data}
\label{sec:synthetic_data_tables}

\begin{table}[htbp]
  \centering
  \caption{Clustering of Works Based on Quality Mechanisms}
  \label{tab:clustering}
  \begin{tabular}{>{\centering\arraybackslash}p{3cm} p{5cm} p{7cm}}
    \toprule
    \textbf{Category} & \textbf{Subcategory} & \textbf{Works} \\
    \midrule
    \textbf{Data Generation} & Using Strong Teacher Models & \citep{mukherjee2023orcaprogressivelearningcomplex}, \citep{gunasekar2023textbooks,li2023textbooks,javaheripi2023phi,abdin2024phi}, \citep{vicuna2023}, \citep{openai2024gpt4technicalreport}, \citep{wang2022supernaturalinstructionsgeneralizationdeclarativeinstructions}, \citep{honovich2022unnaturalinstructionstuninglanguage} \\
    \cline{2-3}
    & Generating Complex Chain-of-Thought & \citep{gunasekar2023textbooks,li2023textbooks,javaheripi2023phi,abdin2024phi}, \citep{mukherjee2023orcaprogressivelearningcomplex}\citep{yu2024metamathbootstrapmathematicalquestions}, \citep{liu2023tinygsmachieving80gsm8k}, \citep{yue2023mammothbuildingmathgeneralist}, \citep{toshniwal2024openmathinstruct118millionmath} \\
    \cline{2-3}
    % & Synthetic Data for Math and Reasoning & \citep{yu2024metamathbootstrapmathematicalquestions}, \citep{liu2023tinygsmachieving80gsm8k}, \citep{yue2023mammothbuildingmathgeneralist}, \citep{toshniwal2024openmathinstruct118millionmath} \\
    \cline{2-3}
    & Generating Pairwise Preferences & \citep{kim2023aligninglargelanguagemodels}, \citep{yang2024rlcdreinforcementlearningcontrastive}, \citep{li2023contrastivedecodingopenendedtext}, \citep{pace2024westofnsyntheticpreferencegeneration} \\
    \cline{2-3}
    & Meta-Prompting and Prompt Programming & \citep{reynolds2021promptprogramminglargelanguage} \\
    \midrule
    \textbf{Data Filtration} & Filtering Using Annotations & \citep{zhou2023limaalignment}, \citep{chen2024alpagasustrainingbetteralpaca}, \citep{cao2024instructionmininginstructiondata} \\
    \cline{2-3}
    & Using Trained Reward Models & \citep{bukharin2024datadiversitymattersrobust}, \citep{gulcehre2023reinforcedselftrainingrestlanguage}, \citep{singh2024humandatascalingselftraining} \\
    \cline{2-3}
    & Filtering in RL Approaches & \citep{havrilla2024teachinglargelanguagemodels}, \citep{dong2023raftrewardrankedfinetuning}, \citep{yuan2023scalingrelationshiplearningmathematical} \\
    \midrule
    \textbf{Data Distillation} & RL Algorithms Influencing Quality & \citep{rafailov2024directpreferenceoptimizationlanguage}, \citep{pang2024iterativereasoningpreferenceoptimization}, \citep{setlur2024rlincorrectsyntheticdata}, \citep{schulman2017proximalpolicyoptimizationalgorithms}, \citep{Williams1992}, \citep{ahmadian2024basicsrevisitingreinforcestyle} \\
    \cline{2-3}
    & Applying RL to Improve Reasoning & \citep{havrilla2024teachinglargelanguagemodels}, \citep{wang2024mathshepherdverifyreinforcellms}, \citep{sun2024easytohardgeneralizationscalablealignment} \\
    \cline{2-3}
    & Unlikelihood Training & \citep{Welleck2020Neural} \\
    \bottomrule
  \end{tabular}
\end{table}

\begin{table}[htbp]
  \centering
  \caption{Clustering of Works Based on Diversity Mechanisms}
  \label{tab:diversity_clustering}
  \begin{tabular}{>{\centering\arraybackslash}p{3cm} p{5cm} p{7cm}}
    \toprule
    \textbf{Category} & \textbf{Subcategory} & \textbf{Works} \\
    \midrule
    \textbf{Data Generation} & Diverse Decoding Strategies & \citep{viswanathan2023prompt2modelgeneratingdeployablemodels}, \citep{havrilla2024teachinglargelanguagemodels}, \citep{ye2022zerogenefficientzeroshotlearning}, \citep{holtzman2020curiouscaseneuraltext}, \citep{Welleck2020Neural} \\
    \cline{2-3}
    & Impact of Prompting on Output Diversity & \citep{jiang2024selfplanningcodegenerationlarge}, \citep{naik2024diversitythoughtimprovesreasoning}, \citep{chan2024scalingsyntheticdatacreation}, \citep{toshniwal2024openmathinstruct118millionmath}, \citep{yu2023largelanguagemodelattributed}, \citep{bradley2023qualitydiversityaifeedback}, \citep{fernando2023promptbreeder}, \citep{yang2024rlcdreinforcementlearningcontrastive} \\
    \cline{2-3}
    & Impact of Diverse Seed Datasets & \textit{Generation algorithms}: \citep{wang2023selfinstructaligninglanguagemodels}, \citep{toshniwal2024openmathinstruct118millionmath}, \citep{yu2024metamathbootstrapmathematicalquestions}; \textit{Adaptation algorithms}: \citep{schick2023toolformerlanguagemodelsteach}, \citep{wang2021gpt}, \citep{viswanathan2023prompt2modelgeneratingdeployablemodels}, \citep{gandhi2024bettersyntheticdataretrieving}, \citep{ding2023gpt3gooddataannotator}, \citep{huang2022largelanguagemodelsselfimprove}, \citep{lanchantin2023learningreasonmemorizeselfnotes}, \citep{yue2024mammoth2scalinginstructionsweb}, \citep{agarwal-etal-2021-knowledge} \\
    \cline{2-3}
    & Hybrid Methods & \citep{benallal2024cosmopedia}, \citep{sennrich-etal-2016-improving}, \citep{li2024selfalignmentinstructionbacktranslation} \\
    \cline{2-3}
    & Augmenting with External Tools & \citep{ni2024nextteachinglargelanguage}, \citep{gou2024criticlargelanguagemodels}, \citep{lozhkov2024starcoder2stackv2} \\
    \cline{2-3}
    & Search & \citep{cobbe2021trainingverifierssolvemath}, \citep{yao2023treethoughtsdeliberateproblem}, \citep{tian2024selfimprovementllmsimaginationsearching}, \citep{gandhi2024streamsearchsoslearning}, \citep{shinn2023reflexionlanguageagentsverbal}, \citep{madaan2023selfrefineiterativerefinementselffeedback}, \citep{havrilla2024glorewhenwhereimprove}, \citep{qu2024recursive}, \citep{miao2023selfcheckusingllmszeroshot}, \citep{wang2024planningnaturallanguageimproves} \\
    \cline{2-3}
    & Model Ensembling & \citep{yuan2023scalingrelationshiplearningmathematical}, \citep{liu2023tinygsmachieving80gsm8k}, \citep{kim2023aligninglargelanguagemodels} \\
    \midrule
    \textbf{Data Filtration} & Data Deduplication & \citep{wenzek-etal-2020-ccnet}, \citep{abbas2023semdedupdataefficientlearningwebscale}, \citep{yu2023scaling}, \citep{havrilla2024teachinglargelanguagemodels}, \citep{tian2024selfimprovementllmsimaginationsearching}, \citep{lu2023instaginstructiontagginganalyzing} \\
    \midrule
    \textbf{Data Distillation} & Training Models for Improved Output Diversity & \citep{zhang2024forcingdiffusedistributionslanguage}, \citep{li2024entropicdistributionmatchingsupervised}, \citep{li2016diversitypromotingobjectivefunctionneural}, \citep{havrilla2024teachinglargelanguagemodels} \\
    \bottomrule
  \end{tabular}
\end{table}

\begin{table}[htbp]
  \centering
  \caption{Clustering of Works Based on Complexity Mechanisms}
  \label{tab:complexity_clustering}
  \begin{tabular}{>{\centering\arraybackslash}p{3cm} p{5cm} p{7cm}}
    \toprule
    \textbf{Category} & \textbf{Subcategory} & \textbf{Works} \\
    \midrule
    \textbf{Data Generation} & Evolving Samples for Complexity & \citep{xu2023wizardlmempoweringlargelanguage}, \citep{luo2023wizardcoderempoweringcodelarge}, \citep{luo2023wizardmathempoweringmathematicalreasoning}, \citep{zeng2024automaticinstructionevolvinglarge} \\
    \cline{2-3}
    & Gathering Feedback for RL & \citep{luo2023wizardmathempoweringmathematicalreasoning} \\
    \cline{2-3}
    & Generalizing the Evolution Process & \citep{zeng2024automaticinstructionevolvinglarge} \\
    \midrule
    \textbf{Data Filtration} & Measuring and Filtering by Instruction Complexity & \citep{zhao2024preliminarystudyintrinsicrelationship}, \citep{lu2023instaginstructiontagginganalyzing} \\
    \midrule
    \textbf{Data Distillation} & Curriculum Learning and Weak-to-Strong RL & \citep{sun2024easytohardgeneralizationscalablealignment} \\
    \bottomrule
  \end{tabular}
\end{table}

\FloatBarrier
\newpage

%% file: survey.bbl
\begin{thebibliography}{247}
\providecommand{\natexlab}[1]{#1}
\providecommand{\url}[1]{\texttt{#1}}
\expandafter\ifx\csname urlstyle\endcsname\relax
  \providecommand{\doi}[1]{doi: #1}\else
  \providecommand{\doi}{doi: \begingroup \urlstyle{rm}\Url}\fi

\bibitem[Abbas et~al.(2023)Abbas, Tirumala, Simig, Ganguli, and Morcos]{abbas2023semdedupdataefficientlearningwebscale}
Amro Abbas, Kushal Tirumala, Dániel Simig, Surya Ganguli, and Ari~S. Morcos.
\newblock Semdedup: Data-efficient learning at web-scale through semantic deduplication, 2023.
\newblock URL \url{https://arxiv.org/abs/2303.09540}.

\bibitem[Abdin et~al.(2024)Abdin, Jacobs, Awan, Aneja, Awadallah, Awadalla, Bach, Bahree, Bakhtiari, Behl, et~al.]{abdin2024phi}
Marah Abdin, Sam~Ade Jacobs, Ammar~Ahmad Awan, Jyoti Aneja, Ahmed Awadallah, Hany Awadalla, Nguyen Bach, Amit Bahree, Arash Bakhtiari, Harkirat Behl, et~al.
\newblock Phi-3 technical report: A highly capable language model locally on your phone.
\newblock \emph{arXiv preprint arXiv:2404.14219}, 2024.

\bibitem[Agarwal et~al.(2021)Agarwal, Ge, Shakeri, and Al-Rfou]{agarwal-etal-2021-knowledge}
Oshin Agarwal, Heming Ge, Siamak Shakeri, and Rami Al-Rfou.
\newblock Knowledge graph based synthetic corpus generation for knowledge-enhanced language model pre-training.
\newblock In Kristina Toutanova, Anna Rumshisky, Luke Zettlemoyer, Dilek Hakkani-Tur, Iz~Beltagy, Steven Bethard, Ryan Cotterell, Tanmoy Chakraborty, and Yichao Zhou (eds.), \emph{Proceedings of the 2021 Conference of the North American Chapter of the Association for Computational Linguistics: Human Language Technologies}, pp.\  3554--3565, Online, June 2021. Association for Computational Linguistics.
\newblock \doi{10.18653/v1/2021.naacl-main.278}.
\newblock URL \url{https://aclanthology.org/2021.naacl-main.278}.

\bibitem[Ahmadian et~al.(2024)Ahmadian, Cremer, Gallé, Fadaee, Kreutzer, Pietquin, Üstün, and Hooker]{ahmadian2024basicsrevisitingreinforcestyle}
Arash Ahmadian, Chris Cremer, Matthias Gallé, Marzieh Fadaee, Julia Kreutzer, Olivier Pietquin, Ahmet Üstün, and Sara Hooker.
\newblock Back to basics: Revisiting reinforce style optimization for learning from human feedback in llms, 2024.
\newblock URL \url{https://arxiv.org/abs/2402.14740}.

\bibitem[Albalak et~al.(2023{\natexlab{a}})Albalak, Pan, Raffel, and Wang]{albalak2023efficient}
Alon Albalak, Liangming Pan, Colin Raffel, and William~Yang Wang.
\newblock Efficient online data mixing for language model pre-training.
\newblock In \emph{R0-FoMo:Robustness of Few-shot and Zero-shot Learning in Large Foundation Models}, 2023{\natexlab{a}}.
\newblock URL \url{https://openreview.net/forum?id=9Tze4oy4lw}.

\bibitem[Albalak et~al.(2023{\natexlab{b}})Albalak, Raffel, and Wang]{albalak2024improving}
Alon Albalak, Colin Raffel, and William~Yang Wang.
\newblock Improving few-shot generalization by exploring and exploiting auxiliary data.
\newblock In \emph{Thirty-seventh Conference on Neural Information Processing Systems}, 2023{\natexlab{b}}.
\newblock URL \url{https://openreview.net/forum?id=JDnLXc4NOn}.

\bibitem[Albalak et~al.(2024)Albalak, Elazar, Xie, Longpre, Lambert, Wang, Muennighoff, Hou, Pan, Jeong, Raffel, Chang, Hashimoto, and Wang]{albalak2024surveydataselectionlanguage}
Alon Albalak, Yanai Elazar, Sang~Michael Xie, Shayne Longpre, Nathan Lambert, Xinyi Wang, Niklas Muennighoff, Bairu Hou, Liangming Pan, Haewon Jeong, Colin Raffel, Shiyu Chang, Tatsunori Hashimoto, and William~Yang Wang.
\newblock A survey on data selection for language models, 2024.
\newblock URL \url{https://arxiv.org/abs/2402.16827}.

\bibitem[Ankner et~al.(2024)Ankner, Paul, Cui, Chang, and Ammanabrolu]{ankner2024critiqueoutloudrewardmodels}
Zachary Ankner, Mansheej Paul, Brandon Cui, Jonathan~D. Chang, and Prithviraj Ammanabrolu.
\newblock Critique-out-loud reward models, 2024.
\newblock URL \url{https://arxiv.org/abs/2408.11791}.

\bibitem[Bai et~al.(2022{\natexlab{a}})Bai, Jones, Ndousse, Askell, Chen, DasSarma, Drain, Fort, Ganguli, Henighan, Joseph, Kadavath, Kernion, Conerly, El-Showk, Elhage, Hatfield-Dodds, Hernandez, Hume, Johnston, Kravec, Lovitt, Nanda, Olsson, Amodei, Brown, Clark, McCandlish, Olah, Mann, and Kaplan]{bai2022traininghelpfulharmlessassistant}
Yuntao Bai, Andy Jones, Kamal Ndousse, Amanda Askell, Anna Chen, Nova DasSarma, Dawn Drain, Stanislav Fort, Deep Ganguli, Tom Henighan, Nicholas Joseph, Saurav Kadavath, Jackson Kernion, Tom Conerly, Sheer El-Showk, Nelson Elhage, Zac Hatfield-Dodds, Danny Hernandez, Tristan Hume, Scott Johnston, Shauna Kravec, Liane Lovitt, Neel Nanda, Catherine Olsson, Dario Amodei, Tom Brown, Jack Clark, Sam McCandlish, Chris Olah, Ben Mann, and Jared Kaplan.
\newblock Training a helpful and harmless assistant with reinforcement learning from human feedback, 2022{\natexlab{a}}.
\newblock URL \url{https://arxiv.org/abs/2204.05862}.

\bibitem[Bai et~al.(2022{\natexlab{b}})Bai, Kadavath, Kundu, Askell, Kernion, Jones, Chen, Goldie, Mirhoseini, McKinnon, Chen, Olsson, Olah, Hernandez, Drain, Ganguli, Li, Tran-Johnson, Perez, Kerr, Mueller, Ladish, Landau, Ndousse, Lukosuite, Lovitt, Sellitto, Elhage, Schiefer, Mercado, DasSarma, Lasenby, Larson, Ringer, Johnston, Kravec, Showk, Fort, Lanham, Telleen-Lawton, Conerly, Henighan, Hume, Bowman, Hatfield-Dodds, Mann, Amodei, Joseph, McCandlish, Brown, and Kaplan]{bai2022constitutionalaiharmlessnessai}
Yuntao Bai, Saurav Kadavath, Sandipan Kundu, Amanda Askell, Jackson Kernion, Andy Jones, Anna Chen, Anna Goldie, Azalia Mirhoseini, Cameron McKinnon, Carol Chen, Catherine Olsson, Christopher Olah, Danny Hernandez, Dawn Drain, Deep Ganguli, Dustin Li, Eli Tran-Johnson, Ethan Perez, Jamie Kerr, Jared Mueller, Jeffrey Ladish, Joshua Landau, Kamal Ndousse, Kamile Lukosuite, Liane Lovitt, Michael Sellitto, Nelson Elhage, Nicholas Schiefer, Noemi Mercado, Nova DasSarma, Robert Lasenby, Robin Larson, Sam Ringer, Scott Johnston, Shauna Kravec, Sheer~El Showk, Stanislav Fort, Tamera Lanham, Timothy Telleen-Lawton, Tom Conerly, Tom Henighan, Tristan Hume, Samuel~R. Bowman, Zac Hatfield-Dodds, Ben Mann, Dario Amodei, Nicholas Joseph, Sam McCandlish, Tom Brown, and Jared Kaplan.
\newblock Constitutional ai: Harmlessness from ai feedback, 2022{\natexlab{b}}.
\newblock URL \url{https://arxiv.org/abs/2212.08073}.

\bibitem[Bauer et~al.(2024)Bauer, Trapp, Stenger, Leppich, Kounev, Leznik, Chard, and Foster]{bauer2024comprehensiveexplorationsyntheticdata}
André Bauer, Simon Trapp, Michael Stenger, Robert Leppich, Samuel Kounev, Mark Leznik, Kyle Chard, and Ian Foster.
\newblock Comprehensive exploration of synthetic data generation: A survey, 2024.
\newblock URL \url{https://arxiv.org/abs/2401.02524}.

\bibitem[Ben~Allal et~al.(2024)Ben~Allal, Lozhkov, Penedo, Wolf, and von Werra]{benallal2024cosmopedia}
Loubna Ben~Allal, Anton Lozhkov, Guilherme Penedo, Thomas Wolf, and Leandro von Werra.
\newblock Cosmopedia, February 2024.
\newblock URL \url{https://huggingface.co/datasets/HuggingFaceTB/cosmopedia}.

\bibitem[Bennett(1988)]{Bennett1988LogicalDA}
Charles~H. Bennett.
\newblock Logical depth and physical complexity.
\newblock 1988.
\newblock URL \url{https://api.semanticscholar.org/CorpusID:6365510}.

\bibitem[Bhatt et~al.(2022)Bhatt, Tjanaka, Fontaine, and Nikolaidis]{bhatt2022deep}
Varun Bhatt, Bryon Tjanaka, Matthew Fontaine, and Stefanos Nikolaidis.
\newblock Deep surrogate assisted generation of environments.
\newblock \emph{Advances in Neural Information Processing Systems}, 35:\penalty0 37762--37777, 2022.

\bibitem[Boratko et~al.(2020)Boratko, Li, Das, O'Gorman, Le, and McCallum]{boratko2020protoqaquestionansweringdataset}
Michael Boratko, Xiang~Lorraine Li, Rajarshi Das, Tim O'Gorman, Dan Le, and Andrew McCallum.
\newblock Protoqa: A question answering dataset for prototypical common-sense reasoning, 2020.
\newblock URL \url{https://arxiv.org/abs/2005.00771}.

\bibitem[Bradley et~al.(2023)Bradley, Dai, Teufel, Zhang, Oostermeijer, Bellagente, Clune, Stanley, Schott, and Lehman]{bradley2023qualitydiversityaifeedback}
Herbie Bradley, Andrew Dai, Hannah Teufel, Jenny Zhang, Koen Oostermeijer, Marco Bellagente, Jeff Clune, Kenneth Stanley, Grégory Schott, and Joel Lehman.
\newblock Quality-diversity through ai feedback, 2023.
\newblock URL \url{https://arxiv.org/abs/2310.13032}.

\bibitem[Brown et~al.(2024)Brown, Juravsky, Ehrlich, Clark, Le, Ré, and Mirhoseini]{brown2024largelanguagemonkeysscaling}
Bradley Brown, Jordan Juravsky, Ryan Ehrlich, Ronald Clark, Quoc~V. Le, Christopher Ré, and Azalia Mirhoseini.
\newblock Large language monkeys: Scaling inference compute with repeated sampling, 2024.
\newblock URL \url{https://arxiv.org/abs/2407.21787}.

\bibitem[Brown et~al.(2020{\natexlab{a}})Brown, Mann, Ryder, Subbiah, Kaplan, Dhariwal, Neelakantan, Shyam, Sastry, Askell, Agarwal, Herbert-Voss, Krueger, Henighan, Child, Ramesh, Ziegler, Wu, Winter, Hesse, Chen, Sigler, Litwin, Gray, Chess, Clark, Berner, McCandlish, Radford, Sutskever, and Amodei]{NEURIPS2020_1457c0d6}
Tom Brown, Benjamin Mann, Nick Ryder, Melanie Subbiah, Jared~D Kaplan, Prafulla Dhariwal, Arvind Neelakantan, Pranav Shyam, Girish Sastry, Amanda Askell, Sandhini Agarwal, Ariel Herbert-Voss, Gretchen Krueger, Tom Henighan, Rewon Child, Aditya Ramesh, Daniel Ziegler, Jeffrey Wu, Clemens Winter, Chris Hesse, Mark Chen, Eric Sigler, Mateusz Litwin, Scott Gray, Benjamin Chess, Jack Clark, Christopher Berner, Sam McCandlish, Alec Radford, Ilya Sutskever, and Dario Amodei.
\newblock Language models are few-shot learners.
\newblock In H.~Larochelle, M.~Ranzato, R.~Hadsell, M.F. Balcan, and H.~Lin (eds.), \emph{Advances in Neural Information Processing Systems}, volume~33, pp.\  1877--1901. Curran Associates, Inc., 2020{\natexlab{a}}.
\newblock URL \url{https://proceedings.neurips.cc/paper_files/paper/2020/file/1457c0d6bfcb4967418bfb8ac142f64a-Paper.pdf}.

\bibitem[Brown et~al.(2020{\natexlab{b}})Brown, Mann, Ryder, Subbiah, Kaplan, Dhariwal, Neelakantan, Shyam, Sastry, Askell, Agarwal, Herbert-Voss, Krueger, Henighan, Child, Ramesh, Ziegler, Wu, Winter, Hesse, Chen, Sigler, Litwin, Gray, Chess, Clark, Berner, McCandlish, Radford, Sutskever, and Amodei]{brown2020languagemodelsfewshotlearners}
Tom~B. Brown, Benjamin Mann, Nick Ryder, Melanie Subbiah, Jared Kaplan, Prafulla Dhariwal, Arvind Neelakantan, Pranav Shyam, Girish Sastry, Amanda Askell, Sandhini Agarwal, Ariel Herbert-Voss, Gretchen Krueger, Tom Henighan, Rewon Child, Aditya Ramesh, Daniel~M. Ziegler, Jeffrey Wu, Clemens Winter, Christopher Hesse, Mark Chen, Eric Sigler, Mateusz Litwin, Scott Gray, Benjamin Chess, Jack Clark, Christopher Berner, Sam McCandlish, Alec Radford, Ilya Sutskever, and Dario Amodei.
\newblock Language models are few-shot learners, 2020{\natexlab{b}}.
\newblock URL \url{https://arxiv.org/abs/2005.14165}.

\bibitem[Bukharin \& Zhao(2024)Bukharin and Zhao]{bukharin2024datadiversitymattersrobust}
Alexander Bukharin and Tuo Zhao.
\newblock Data diversity matters for robust instruction tuning, 2024.
\newblock URL \url{https://arxiv.org/abs/2311.14736}.

\bibitem[Burns et~al.(2023)Burns, Izmailov, Kirchner, Baker, Gao, Aschenbrenner, Chen, Ecoffet, Joglekar, Leike, Sutskever, and Wu]{burns2023weaktostronggeneralizationelicitingstrong}
Collin Burns, Pavel Izmailov, Jan~Hendrik Kirchner, Bowen Baker, Leo Gao, Leopold Aschenbrenner, Yining Chen, Adrien Ecoffet, Manas Joglekar, Jan Leike, Ilya Sutskever, and Jeff Wu.
\newblock Weak-to-strong generalization: Eliciting strong capabilities with weak supervision, 2023.
\newblock URL \url{https://arxiv.org/abs/2312.09390}.

\bibitem[Cao et~al.(2024)Cao, Kang, Wang, and Sun]{cao2024instructionmininginstructiondata}
Yihan Cao, Yanbin Kang, Chi Wang, and Lichao Sun.
\newblock Instruction mining: Instruction data selection for tuning large language models, 2024.
\newblock URL \url{https://arxiv.org/abs/2307.06290}.

\bibitem[Chakrabarty(2019)]{chakrabarty2019machinelearningapproachcomment}
Navoneel Chakrabarty.
\newblock A machine learning approach to comment toxicity classification, 2019.
\newblock URL \url{https://arxiv.org/abs/1903.06765}.

\bibitem[Chan et~al.(2024)Chan, Wang, Yu, Mi, and Yu]{chan2024scalingsyntheticdatacreation}
Xin Chan, Xiaoyang Wang, Dian Yu, Haitao Mi, and Dong Yu.
\newblock Scaling synthetic data creation with 1,000,000,000 personas, 2024.
\newblock URL \url{https://arxiv.org/abs/2406.20094}.

\bibitem[Chang et~al.(2024)Chang, Fontaine, Booth, Matarić, and Nikolaidis]{chang2024qualitydiversitygenerativesamplinglearning}
Allen Chang, Matthew~C. Fontaine, Serena Booth, Maja~J. Matarić, and Stefanos Nikolaidis.
\newblock Quality-diversity generative sampling for learning with synthetic data, 2024.
\newblock URL \url{https://arxiv.org/abs/2312.14369}.

\bibitem[Chao et~al.(2024)Chao, Zhao, Jiao, Li, Liu, and Yang]{chao2024match}
Wang Chao, Jiaxuan Zhao, Licheng Jiao, Lingling Li, Fang Liu, and Shuyuan Yang.
\newblock A match made in consistency heaven: when large language models meet evolutionary algorithms.
\newblock \emph{arXiv preprint arXiv:2401.10510}, 2024.

\bibitem[Chatzilygeroudis et~al.(2021)Chatzilygeroudis, Cully, Vassiliades, and Mouret]{chatzilygeroudis2021quality}
Konstantinos Chatzilygeroudis, Antoine Cully, Vassilis Vassiliades, and Jean-Baptiste Mouret.
\newblock Quality-diversity optimization: a novel branch of stochastic optimization.
\newblock In \emph{Black Box Optimization, Machine Learning, and No-Free Lunch Theorems}, pp.\  109--135. Springer, 2021.

\bibitem[Chen et~al.(2023{\natexlab{a}})Chen, Dohan, and So]{chen2023evopromptinglanguagemodelscodelevel}
Angelica Chen, David~M. Dohan, and David~R. So.
\newblock Evoprompting: Language models for code-level neural architecture search, 2023{\natexlab{a}}.
\newblock URL \url{https://arxiv.org/abs/2302.14838}.

\bibitem[Chen et~al.(2024{\natexlab{a}})Chen, Waheed, Li, Wang, Wang, Raj, and Abdin]{chen2024diversitysyntheticdataimpact}
Hao Chen, Abdul Waheed, Xiang Li, Yidong Wang, Jindong Wang, Bhiksha Raj, and Marah~I. Abdin.
\newblock On the diversity of synthetic data and its impact on training large language models, 2024{\natexlab{a}}.
\newblock URL \url{https://arxiv.org/abs/2410.15226}.

\bibitem[Chen et~al.(2024{\natexlab{b}})Chen, Li, Yan, Wang, Gunaratna, Yadav, Tang, Srinivasan, Zhou, Huang, and Jin]{chen2024alpagasustrainingbetteralpaca}
Lichang Chen, Shiyang Li, Jun Yan, Hai Wang, Kalpa Gunaratna, Vikas Yadav, Zheng Tang, Vijay Srinivasan, Tianyi Zhou, Heng Huang, and Hongxia Jin.
\newblock Alpagasus: Training a better alpaca with fewer data, 2024{\natexlab{b}}.
\newblock URL \url{https://arxiv.org/abs/2307.08701}.

\bibitem[Chen et~al.(2023{\natexlab{b}})Chen, Roberts, Bhatia, Wang, Zhang, Sala, and Ré]{chen2023skillitdatadrivenskillsframework}
Mayee~F. Chen, Nicholas Roberts, Kush Bhatia, Jue Wang, Ce~Zhang, Frederic Sala, and Christopher Ré.
\newblock Skill-it! a data-driven skills framework for understanding and training language models, 2023{\natexlab{b}}.
\newblock URL \url{https://arxiv.org/abs/2307.14430}.

\bibitem[Chen et~al.(2023{\natexlab{c}})Chen, Ma, Wang, and Cohen]{chen2023programthoughtspromptingdisentangling}
Wenhu Chen, Xueguang Ma, Xinyi Wang, and William~W. Cohen.
\newblock Program of thoughts prompting: Disentangling computation from reasoning for numerical reasoning tasks, 2023{\natexlab{c}}.
\newblock URL \url{https://arxiv.org/abs/2211.12588}.

\bibitem[Chen et~al.(2024{\natexlab{c}})Chen, Deng, Yuan, Ji, and Gu]{chen2024selfplayfinetuningconvertsweak}
Zixiang Chen, Yihe Deng, Huizhuo Yuan, Kaixuan Ji, and Quanquan Gu.
\newblock Self-play fine-tuning converts weak language models to strong language models, 2024{\natexlab{c}}.
\newblock URL \url{https://arxiv.org/abs/2401.01335}.

\bibitem[Chen et~al.(2023{\natexlab{d}})Chen, Kiami, Gupta, and Kumar]{chen2023genaug}
Zoey Chen, Sho Kiami, Abhishek Gupta, and Vikash Kumar.
\newblock Genaug: Retargeting behaviors to unseen situations via generative augmentation.
\newblock \emph{arXiv preprint arXiv:2302.06671}, 2023{\natexlab{d}}.

\bibitem[Chiang et~al.(2023)Chiang, Li, Lin, Sheng, Wu, Zhang, Zheng, Zhuang, Zhuang, Gonzalez, Stoica, and Xing]{vicuna2023}
Wei-Lin Chiang, Zhuohan Li, Zi~Lin, Ying Sheng, Zhanghao Wu, Hao Zhang, Lianmin Zheng, Siyuan Zhuang, Yonghao Zhuang, Joseph~E. Gonzalez, Ion Stoica, and Eric~P. Xing.
\newblock Vicuna: An open-source chatbot impressing gpt-4 with 90\%* chatgpt quality, March 2023.
\newblock URL \url{https://lmsys.org/blog/2023-03-30-vicuna/}.

\bibitem[Cideron et~al.(2024)Cideron, Agostinelli, Ferret, Girgin, Elie, Bachem, Perrin, and Ramé]{cideron2024diversityrewardedcfgdistillation}
Geoffrey Cideron, Andrea Agostinelli, Johan Ferret, Sertan Girgin, Romuald Elie, Olivier Bachem, Sarah Perrin, and Alexandre Ramé.
\newblock Diversity-rewarded cfg distillation, 2024.
\newblock URL \url{https://arxiv.org/abs/2410.06084}.

\bibitem[Cobbe et~al.(2021)Cobbe, Kosaraju, Bavarian, Chen, Jun, Kaiser, Plappert, Tworek, Hilton, Nakano, Hesse, and Schulman]{cobbe2021trainingverifierssolvemath}
Karl Cobbe, Vineet Kosaraju, Mohammad Bavarian, Mark Chen, Heewoo Jun, Lukasz Kaiser, Matthias Plappert, Jerry Tworek, Jacob Hilton, Reiichiro Nakano, Christopher Hesse, and John Schulman.
\newblock Training verifiers to solve math word problems, 2021.
\newblock URL \url{https://arxiv.org/abs/2110.14168}.

\bibitem[Cully \& Demiris(2017)Cully and Demiris]{cully2017quality}
Antoine Cully and Yiannis Demiris.
\newblock Quality and diversity optimization: A unifying modular framework.
\newblock \emph{IEEE Transactions on Evolutionary Computation}, 22\penalty0 (2):\penalty0 245--259, 2017.

\bibitem[DeMars(2010)]{DeMars2010ItemRT}
Christine~E. DeMars.
\newblock Item response theory.
\newblock \emph{Assessing Measurement Invariance for Applied Research}, 2010.
\newblock URL \url{https://api.semanticscholar.org/CorpusID:15155431}.

\bibitem[Ding et~al.(2023)Ding, Qin, Liu, Chia, Joty, Li, and Bing]{ding2023gpt3gooddataannotator}
Bosheng Ding, Chengwei Qin, Linlin Liu, Yew~Ken Chia, Shafiq Joty, Boyang Li, and Lidong Bing.
\newblock Is gpt-3 a good data annotator?, 2023.
\newblock URL \url{https://arxiv.org/abs/2212.10450}.

\bibitem[Ding et~al.(2024)Ding, Zhang, Clune, Spector, and Lehman]{ding2024qualitydiversityhumanfeedback}
Li~Ding, Jenny Zhang, Jeff Clune, Lee Spector, and Joel Lehman.
\newblock Quality diversity through human feedback: Towards open-ended diversity-driven optimization, 2024.
\newblock URL \url{https://arxiv.org/abs/2310.12103}.

\bibitem[Dong et~al.(2024)Dong, Yuan, Lu, Li, Xue, Liu, Wang, Yuan, Zhou, and Zhou]{dong2024abilitieslargelanguagemodels}
Guanting Dong, Hongyi Yuan, Keming Lu, Chengpeng Li, Mingfeng Xue, Dayiheng Liu, Wei Wang, Zheng Yuan, Chang Zhou, and Jingren Zhou.
\newblock How abilities in large language models are affected by supervised fine-tuning data composition, 2024.
\newblock URL \url{https://arxiv.org/abs/2310.05492}.

\bibitem[Dong et~al.(2023)Dong, Xiong, Goyal, Zhang, Chow, Pan, Diao, Zhang, Shum, and Zhang]{dong2023raftrewardrankedfinetuning}
Hanze Dong, Wei Xiong, Deepanshu Goyal, Yihan Zhang, Winnie Chow, Rui Pan, Shizhe Diao, Jipeng Zhang, Kashun Shum, and Tong Zhang.
\newblock Raft: Reward ranked finetuning for generative foundation model alignment, 2023.
\newblock URL \url{https://arxiv.org/abs/2304.06767}.

\bibitem[Du et~al.(2022)Du, Huang, Dai, Tong, Lepikhin, Xu, Krikun, Zhou, Yu, Firat, Zoph, Fedus, Bosma, Zhou, Wang, Wang, Webster, Pellat, Robinson, Meier-Hellstern, Duke, Dixon, Zhang, Le, Wu, Chen, and Cui]{du2022glamefficientscalinglanguage}
Nan Du, Yanping Huang, Andrew~M. Dai, Simon Tong, Dmitry Lepikhin, Yuanzhong Xu, Maxim Krikun, Yanqi Zhou, Adams~Wei Yu, Orhan Firat, Barret Zoph, Liam Fedus, Maarten Bosma, Zongwei Zhou, Tao Wang, Yu~Emma Wang, Kellie Webster, Marie Pellat, Kevin Robinson, Kathleen Meier-Hellstern, Toju Duke, Lucas Dixon, Kun Zhang, Quoc~V Le, Yonghui Wu, Zhifeng Chen, and Claire Cui.
\newblock Glam: Efficient scaling of language models with mixture-of-experts, 2022.
\newblock URL \url{https://arxiv.org/abs/2112.06905}.

\bibitem[Dubey et~al.(2024{\natexlab{a}})Dubey, Jauhri, Pandey, Kadian, Al-Dahle, Letman, Mathur, Schelten, Yang, Fan, Goyal, Hartshorn, Yang, Mitra, Sravankumar, Korenev, Hinsvark, Rao, Zhang, Rodriguez, Gregerson, Spataru, Roziere, Biron, Tang, Chern, Caucheteux, Nayak, Bi, Marra, McConnell, Keller, Touret, Wu, Wong, Ferrer, Nikolaidis, Allonsius, Song, Pintz, Livshits, Esiobu, Choudhary, Mahajan, Garcia-Olano, Perino, Hupkes, Lakomkin, AlBadawy, Lobanova, Dinan, Smith, Radenovic, Zhang, Synnaeve, Lee, Anderson, Nail, Mialon, Pang, Cucurell, Nguyen, Korevaar, Xu, Touvron, Zarov, Ibarra, Kloumann, Misra, Evtimov, Copet, Lee, Geffert, Vranes, Park, Mahadeokar, Shah, van~der Linde, Billock, Hong, Lee, Fu, Chi, Huang, Liu, Wang, Yu, Bitton, Spisak, Park, Rocca, Johnstun, Saxe, Jia, Alwala, Upasani, Plawiak, Li, Heafield, Stone, El-Arini, Iyer, Malik, Chiu, Bhalla, Rantala-Yeary, van~der Maaten, Chen, Tan, Jenkins, Martin, Madaan, Malo, Blecher, Landzaat, de~Oliveira, Muzzi, Pasupuleti, Singh, Paluri, Kardas,
  Oldham, Rita, Pavlova, Kambadur, Lewis, Si, Singh, Hassan, Goyal, Torabi, Bashlykov, Bogoychev, Chatterji, Duchenne, Çelebi, Alrassy, Zhang, Li, Vasic, Weng, Bhargava, Dubal, Krishnan, Koura, Xu, He, Dong, Srinivasan, Ganapathy, Calderer, Cabral, Stojnic, Raileanu, Girdhar, Patel, Sauvestre, Polidoro, Sumbaly, Taylor, Silva, Hou, Wang, Hosseini, Chennabasappa, Singh, Bell, Kim, Edunov, Nie, Narang, Raparthy, Shen, Wan, Bhosale, Zhang, Vandenhende, Batra, Whitman, Sootla, Collot, Gururangan, Borodinsky, Herman, Fowler, Sheasha, Georgiou, Scialom, Speckbacher, Mihaylov, Xiao, Karn, Goswami, Gupta, Ramanathan, Kerkez, Gonguet, Do, Vogeti, Petrovic, Chu, Xiong, Fu, Meers, Martinet, Wang, Tan, Xie, Jia, Wang, Goldschlag, Gaur, Babaei, Wen, Song, Zhang, Li, Mao, Coudert, Yan, Chen, Papakipos, Singh, Grattafiori, Jain, Kelsey, Shajnfeld, Gangidi, Victoria, Goldstand, Menon, Sharma, Boesenberg, Vaughan, Baevski, Feinstein, Kallet, Sangani, Yunus, Lupu, Alvarado, Caples, Gu, Ho, Poulton, Ryan, Ramchandani, Franco,
  Saraf, Chowdhury, Gabriel, Bharambe, Eisenman, Yazdan, James, Maurer, Leonhardi, Huang, Loyd, Paola, Paranjape, Liu, Wu, Ni, Hancock, Wasti, Spence, Stojkovic, Gamido, Montalvo, Parker, Burton, Mejia, Wang, Kim, Zhou, Hu, Chu, Cai, Tindal, Feichtenhofer, Civin, Beaty, Kreymer, Li, Wyatt, Adkins, Xu, Testuggine, David, Parikh, Liskovich, Foss, Wang, Le, Holland, Dowling, Jamil, Montgomery, Presani, Hahn, Wood, Brinkman, Arcaute, Dunbar, Smothers, Sun, Kreuk, Tian, Ozgenel, Caggioni, Guzmán, Kanayet, Seide, Florez, Schwarz, Badeer, Swee, Halpern, Thattai, Herman, Sizov, Guangyi, Zhang, Lakshminarayanan, Shojanazeri, Zou, Wang, Zha, Habeeb, Rudolph, Suk, Aspegren, Goldman, Damlaj, Molybog, Tufanov, Veliche, Gat, Weissman, Geboski, Kohli, Asher, Gaya, Marcus, Tang, Chan, Zhen, Reizenstein, Teboul, Zhong, Jin, Yang, Cummings, Carvill, Shepard, McPhie, Torres, Ginsburg, Wang, Wu, U, Saxena, Prasad, Khandelwal, Zand, Matosich, Veeraraghavan, Michelena, Li, Huang, Chawla, Lakhotia, Huang, Chen, Garg, A, Silva,
  Bell, Zhang, Guo, Yu, Moshkovich, Wehrstedt, Khabsa, Avalani, Bhatt, Tsimpoukelli, Mankus, Hasson, Lennie, Reso, Groshev, Naumov, Lathi, Keneally, Seltzer, Valko, Restrepo, Patel, Vyatskov, Samvelyan, Clark, Macey, Wang, Hermoso, Metanat, Rastegari, Bansal, Santhanam, Parks, White, Bawa, Singhal, Egebo, Usunier, Laptev, Dong, Zhang, Cheng, Chernoguz, Hart, Salpekar, Kalinli, Kent, Parekh, Saab, Balaji, Rittner, Bontrager, Roux, Dollar, Zvyagina, Ratanchandani, Yuvraj, Liang, Alao, Rodriguez, Ayub, Murthy, Nayani, Mitra, Li, Hogan, Battey, Wang, Maheswari, Howes, Rinott, Bondu, Datta, Chugh, Hunt, Dhillon, Sidorov, Pan, Verma, Yamamoto, Ramaswamy, Lindsay, Lindsay, Feng, Lin, Zha, Shankar, Zhang, Zhang, Wang, Agarwal, Sajuyigbe, Chintala, Max, Chen, Kehoe, Satterfield, Govindaprasad, Gupta, Cho, Virk, Subramanian, Choudhury, Goldman, Remez, Glaser, Best, Kohler, Robinson, Li, Zhang, Matthews, Chou, Shaked, Vontimitta, Ajayi, Montanez, Mohan, Kumar, Mangla, Albiero, Ionescu, Poenaru, Mihailescu, Ivanov, Li,
  Wang, Jiang, Bouaziz, Constable, Tang, Wang, Wu, Wang, Xia, Wu, Gao, Chen, Hu, Jia, Qi, Li, Zhang, Zhang, Adi, Nam, Yu, Wang, Hao, Qian, He, Rait, DeVito, Rosnbrick, Wen, Yang, and Zhao]{dubey2024llama3herdmodels}
Abhimanyu Dubey, Abhinav Jauhri, Abhinav Pandey, Abhishek Kadian, Ahmad Al-Dahle, Aiesha Letman, Akhil Mathur, Alan Schelten, Amy Yang, Angela Fan, Anirudh Goyal, Anthony Hartshorn, Aobo Yang, Archi Mitra, Archie Sravankumar, Artem Korenev, Arthur Hinsvark, Arun Rao, Aston Zhang, Aurelien Rodriguez, Austen Gregerson, Ava Spataru, Baptiste Roziere, Bethany Biron, Binh Tang, Bobbie Chern, Charlotte Caucheteux, Chaya Nayak, Chloe Bi, Chris Marra, Chris McConnell, Christian Keller, Christophe Touret, Chunyang Wu, Corinne Wong, Cristian~Canton Ferrer, Cyrus Nikolaidis, Damien Allonsius, Daniel Song, Danielle Pintz, Danny Livshits, David Esiobu, Dhruv Choudhary, Dhruv Mahajan, Diego Garcia-Olano, Diego Perino, Dieuwke Hupkes, Egor Lakomkin, Ehab AlBadawy, Elina Lobanova, Emily Dinan, Eric~Michael Smith, Filip Radenovic, Frank Zhang, Gabriel Synnaeve, Gabrielle Lee, Georgia~Lewis Anderson, Graeme Nail, Gregoire Mialon, Guan Pang, Guillem Cucurell, Hailey Nguyen, Hannah Korevaar, Hu~Xu, Hugo Touvron, Iliyan Zarov,
  Imanol~Arrieta Ibarra, Isabel Kloumann, Ishan Misra, Ivan Evtimov, Jade Copet, Jaewon Lee, Jan Geffert, Jana Vranes, Jason Park, Jay Mahadeokar, Jeet Shah, Jelmer van~der Linde, Jennifer Billock, Jenny Hong, Jenya Lee, Jeremy Fu, Jianfeng Chi, Jianyu Huang, Jiawen Liu, Jie Wang, Jiecao Yu, Joanna Bitton, Joe Spisak, Jongsoo Park, Joseph Rocca, Joshua Johnstun, Joshua Saxe, Junteng Jia, Kalyan~Vasuden Alwala, Kartikeya Upasani, Kate Plawiak, Ke~Li, Kenneth Heafield, Kevin Stone, Khalid El-Arini, Krithika Iyer, Kshitiz Malik, Kuenley Chiu, Kunal Bhalla, Lauren Rantala-Yeary, Laurens van~der Maaten, Lawrence Chen, Liang Tan, Liz Jenkins, Louis Martin, Lovish Madaan, Lubo Malo, Lukas Blecher, Lukas Landzaat, Luke de~Oliveira, Madeline Muzzi, Mahesh Pasupuleti, Mannat Singh, Manohar Paluri, Marcin Kardas, Mathew Oldham, Mathieu Rita, Maya Pavlova, Melanie Kambadur, Mike Lewis, Min Si, Mitesh~Kumar Singh, Mona Hassan, Naman Goyal, Narjes Torabi, Nikolay Bashlykov, Nikolay Bogoychev, Niladri Chatterji, Olivier
  Duchenne, Onur Çelebi, Patrick Alrassy, Pengchuan Zhang, Pengwei Li, Petar Vasic, Peter Weng, Prajjwal Bhargava, Pratik Dubal, Praveen Krishnan, Punit~Singh Koura, Puxin Xu, Qing He, Qingxiao Dong, Ragavan Srinivasan, Raj Ganapathy, Ramon Calderer, Ricardo~Silveira Cabral, Robert Stojnic, Roberta Raileanu, Rohit Girdhar, Rohit Patel, Romain Sauvestre, Ronnie Polidoro, Roshan Sumbaly, Ross Taylor, Ruan Silva, Rui Hou, Rui Wang, Saghar Hosseini, Sahana Chennabasappa, Sanjay Singh, Sean Bell, Seohyun~Sonia Kim, Sergey Edunov, Shaoliang Nie, Sharan Narang, Sharath Raparthy, Sheng Shen, Shengye Wan, Shruti Bhosale, Shun Zhang, Simon Vandenhende, Soumya Batra, Spencer Whitman, Sten Sootla, Stephane Collot, Suchin Gururangan, Sydney Borodinsky, Tamar Herman, Tara Fowler, Tarek Sheasha, Thomas Georgiou, Thomas Scialom, Tobias Speckbacher, Todor Mihaylov, Tong Xiao, Ujjwal Karn, Vedanuj Goswami, Vibhor Gupta, Vignesh Ramanathan, Viktor Kerkez, Vincent Gonguet, Virginie Do, Vish Vogeti, Vladan Petrovic, Weiwei Chu,
  Wenhan Xiong, Wenyin Fu, Whitney Meers, Xavier Martinet, Xiaodong Wang, Xiaoqing~Ellen Tan, Xinfeng Xie, Xuchao Jia, Xuewei Wang, Yaelle Goldschlag, Yashesh Gaur, Yasmine Babaei, Yi~Wen, Yiwen Song, Yuchen Zhang, Yue Li, Yuning Mao, Zacharie~Delpierre Coudert, Zheng Yan, Zhengxing Chen, Zoe Papakipos, Aaditya Singh, Aaron Grattafiori, Abha Jain, Adam Kelsey, Adam Shajnfeld, Adithya Gangidi, Adolfo Victoria, Ahuva Goldstand, Ajay Menon, Ajay Sharma, Alex Boesenberg, Alex Vaughan, Alexei Baevski, Allie Feinstein, Amanda Kallet, Amit Sangani, Anam Yunus, Andrei Lupu, Andres Alvarado, Andrew Caples, Andrew Gu, Andrew Ho, Andrew Poulton, Andrew Ryan, Ankit Ramchandani, Annie Franco, Aparajita Saraf, Arkabandhu Chowdhury, Ashley Gabriel, Ashwin Bharambe, Assaf Eisenman, Azadeh Yazdan, Beau James, Ben Maurer, Benjamin Leonhardi, Bernie Huang, Beth Loyd, Beto~De Paola, Bhargavi Paranjape, Bing Liu, Bo~Wu, Boyu Ni, Braden Hancock, Bram Wasti, Brandon Spence, Brani Stojkovic, Brian Gamido, Britt Montalvo, Carl
  Parker, Carly Burton, Catalina Mejia, Changhan Wang, Changkyu Kim, Chao Zhou, Chester Hu, Ching-Hsiang Chu, Chris Cai, Chris Tindal, Christoph Feichtenhofer, Damon Civin, Dana Beaty, Daniel Kreymer, Daniel Li, Danny Wyatt, David Adkins, David Xu, Davide Testuggine, Delia David, Devi Parikh, Diana Liskovich, Didem Foss, Dingkang Wang, Duc Le, Dustin Holland, Edward Dowling, Eissa Jamil, Elaine Montgomery, Eleonora Presani, Emily Hahn, Emily Wood, Erik Brinkman, Esteban Arcaute, Evan Dunbar, Evan Smothers, Fei Sun, Felix Kreuk, Feng Tian, Firat Ozgenel, Francesco Caggioni, Francisco Guzmán, Frank Kanayet, Frank Seide, Gabriela~Medina Florez, Gabriella Schwarz, Gada Badeer, Georgia Swee, Gil Halpern, Govind Thattai, Grant Herman, Grigory Sizov, Guangyi, Zhang, Guna Lakshminarayanan, Hamid Shojanazeri, Han Zou, Hannah Wang, Hanwen Zha, Haroun Habeeb, Harrison Rudolph, Helen Suk, Henry Aspegren, Hunter Goldman, Ibrahim Damlaj, Igor Molybog, Igor Tufanov, Irina-Elena Veliche, Itai Gat, Jake Weissman, James
  Geboski, James Kohli, Japhet Asher, Jean-Baptiste Gaya, Jeff Marcus, Jeff Tang, Jennifer Chan, Jenny Zhen, Jeremy Reizenstein, Jeremy Teboul, Jessica Zhong, Jian Jin, Jingyi Yang, Joe Cummings, Jon Carvill, Jon Shepard, Jonathan McPhie, Jonathan Torres, Josh Ginsburg, Junjie Wang, Kai Wu, Kam~Hou U, Karan Saxena, Karthik Prasad, Kartikay Khandelwal, Katayoun Zand, Kathy Matosich, Kaushik Veeraraghavan, Kelly Michelena, Keqian Li, Kun Huang, Kunal Chawla, Kushal Lakhotia, Kyle Huang, Lailin Chen, Lakshya Garg, Lavender A, Leandro Silva, Lee Bell, Lei Zhang, Liangpeng Guo, Licheng Yu, Liron Moshkovich, Luca Wehrstedt, Madian Khabsa, Manav Avalani, Manish Bhatt, Maria Tsimpoukelli, Martynas Mankus, Matan Hasson, Matthew Lennie, Matthias Reso, Maxim Groshev, Maxim Naumov, Maya Lathi, Meghan Keneally, Michael~L. Seltzer, Michal Valko, Michelle Restrepo, Mihir Patel, Mik Vyatskov, Mikayel Samvelyan, Mike Clark, Mike Macey, Mike Wang, Miquel~Jubert Hermoso, Mo~Metanat, Mohammad Rastegari, Munish Bansal, Nandhini
  Santhanam, Natascha Parks, Natasha White, Navyata Bawa, Nayan Singhal, Nick Egebo, Nicolas Usunier, Nikolay~Pavlovich Laptev, Ning Dong, Ning Zhang, Norman Cheng, Oleg Chernoguz, Olivia Hart, Omkar Salpekar, Ozlem Kalinli, Parkin Kent, Parth Parekh, Paul Saab, Pavan Balaji, Pedro Rittner, Philip Bontrager, Pierre Roux, Piotr Dollar, Polina Zvyagina, Prashant Ratanchandani, Pritish Yuvraj, Qian Liang, Rachad Alao, Rachel Rodriguez, Rafi Ayub, Raghotham Murthy, Raghu Nayani, Rahul Mitra, Raymond Li, Rebekkah Hogan, Robin Battey, Rocky Wang, Rohan Maheswari, Russ Howes, Ruty Rinott, Sai~Jayesh Bondu, Samyak Datta, Sara Chugh, Sara Hunt, Sargun Dhillon, Sasha Sidorov, Satadru Pan, Saurabh Verma, Seiji Yamamoto, Sharadh Ramaswamy, Shaun Lindsay, Shaun Lindsay, Sheng Feng, Shenghao Lin, Shengxin~Cindy Zha, Shiva Shankar, Shuqiang Zhang, Shuqiang Zhang, Sinong Wang, Sneha Agarwal, Soji Sajuyigbe, Soumith Chintala, Stephanie Max, Stephen Chen, Steve Kehoe, Steve Satterfield, Sudarshan Govindaprasad, Sumit Gupta,
  Sungmin Cho, Sunny Virk, Suraj Subramanian, Sy~Choudhury, Sydney Goldman, Tal Remez, Tamar Glaser, Tamara Best, Thilo Kohler, Thomas Robinson, Tianhe Li, Tianjun Zhang, Tim Matthews, Timothy Chou, Tzook Shaked, Varun Vontimitta, Victoria Ajayi, Victoria Montanez, Vijai Mohan, Vinay~Satish Kumar, Vishal Mangla, Vítor Albiero, Vlad Ionescu, Vlad Poenaru, Vlad~Tiberiu Mihailescu, Vladimir Ivanov, Wei Li, Wenchen Wang, Wenwen Jiang, Wes Bouaziz, Will Constable, Xiaocheng Tang, Xiaofang Wang, Xiaojian Wu, Xiaolan Wang, Xide Xia, Xilun Wu, Xinbo Gao, Yanjun Chen, Ye~Hu, Ye~Jia, Ye~Qi, Yenda Li, Yilin Zhang, Ying Zhang, Yossi Adi, Youngjin Nam, Yu, Wang, Yuchen Hao, Yundi Qian, Yuzi He, Zach Rait, Zachary DeVito, Zef Rosnbrick, Zhaoduo Wen, Zhenyu Yang, and Zhiwei Zhao.
\newblock The llama 3 herd of models, 2024{\natexlab{a}}.
\newblock URL \url{https://arxiv.org/abs/2407.21783}.

\bibitem[Dubey et~al.(2024{\natexlab{b}})Dubey, Jauhri, Pandey, Kadian, Al-Dahle, Letman, Mathur, Schelten, Yang, Fan, et~al.]{dubey2024llama}
Abhimanyu Dubey, Abhinav Jauhri, Abhinav Pandey, Abhishek Kadian, Ahmad Al-Dahle, Aiesha Letman, Akhil Mathur, Alan Schelten, Amy Yang, Angela Fan, et~al.
\newblock The llama 3 herd of models.
\newblock \emph{arXiv preprint arXiv:2407.21783}, 2024{\natexlab{b}}.

\bibitem[Eldan \& Li(2023)Eldan and Li]{eldan2023tinystoriessmalllanguagemodels}
Ronen Eldan and Yuanzhi Li.
\newblock Tinystories: How small can language models be and still speak coherent english?, 2023.
\newblock URL \url{https://arxiv.org/abs/2305.07759}.

\bibitem[Ethayarajh et~al.(2022)Ethayarajh, Choi, and Swayamdipta]{pmlr-v162-ethayarajh22a}
Kawin Ethayarajh, Yejin Choi, and Swabha Swayamdipta.
\newblock Understanding dataset difficulty with $\mathcal{V}$-usable information.
\newblock In Kamalika Chaudhuri, Stefanie Jegelka, Le~Song, Csaba Szepesvari, Gang Niu, and Sivan Sabato (eds.), \emph{Proceedings of the 39th International Conference on Machine Learning}, volume 162 of \emph{Proceedings of Machine Learning Research}, pp.\  5988--6008. PMLR, 17--23 Jul 2022.

\bibitem[Fan et~al.(2023)Fan, Chen, Krishnan, Katabi, Isola, and Tian]{fan2023scalinglawssyntheticimages}
Lijie Fan, Kaifeng Chen, Dilip Krishnan, Dina Katabi, Phillip Isola, and Yonglong Tian.
\newblock Scaling laws of synthetic images for model training ... for now, 2023.
\newblock URL \url{https://arxiv.org/abs/2312.04567}.

\bibitem[Fan \& Jaggi(2023)Fan and Jaggi]{fan2023irreduciblecurriculumlanguagemodel}
Simin Fan and Martin Jaggi.
\newblock Irreducible curriculum for language model pretraining, 2023.
\newblock URL \url{https://arxiv.org/abs/2310.15389}.

\bibitem[Fernando et~al.(2023{\natexlab{a}})Fernando, Banarse, Michalewski, Osindero, and Rocktäschel]{fernando2023promptbreeder}
Chrisantha Fernando, Dylan Banarse, Henryk Michalewski, Simon Osindero, and Tim Rocktäschel.
\newblock Promptbreeder: Self-referential self-improvement via prompt evolution, 2023{\natexlab{a}}.

\bibitem[Fernando et~al.(2023{\natexlab{b}})Fernando, Banarse, Michalewski, Osindero, and Rocktäschel]{fernando2023promptbreederselfreferentialselfimprovementprompt}
Chrisantha Fernando, Dylan Banarse, Henryk Michalewski, Simon Osindero, and Tim Rocktäschel.
\newblock Promptbreeder: Self-referential self-improvement via prompt evolution, 2023{\natexlab{b}}.
\newblock URL \url{https://arxiv.org/abs/2309.16797}.

\bibitem[Fontaine \& Nikolaidis(2021)Fontaine and Nikolaidis]{fontaine2021quality}
Matthew Fontaine and Stefanos Nikolaidis.
\newblock A quality diversity approach to automatically generating human-robot interaction scenarios in shared autonomy.
\newblock \emph{Robotics: Science and Systems (RSS)}, 2021.

\bibitem[Fontaine et~al.(2021{\natexlab{a}})Fontaine, Hsu, Zhang, Tjanaka, and Nikolaidis]{fontaine2021importance}
Matthew Fontaine, Sophie Hsu, Yulun Zhang, Bryon Tjanaka, and Stefanos Nikolaidis.
\newblock On the importance of environments in human-robot coordination.
\newblock \emph{Robotics: Science and Systems (RSS)}, 2021{\natexlab{a}}.

\bibitem[Fontaine et~al.(2021{\natexlab{b}})Fontaine, Liu, Khalifa, Modi, Togelius, Hoover, and Nikolaidis]{fontaine2021illuminatingmariosceneslatent}
Matthew~C. Fontaine, Ruilin Liu, Ahmed Khalifa, Jignesh Modi, Julian Togelius, Amy~K. Hoover, and Stefanos Nikolaidis.
\newblock Illuminating mario scenes in the latent space of a generative adversarial network, 2021{\natexlab{b}}.
\newblock URL \url{https://arxiv.org/abs/2007.05674}.

\bibitem[Gandhi et~al.(2024{\natexlab{a}})Gandhi, Lee, Grand, Liu, Cheng, Sharma, and Goodman]{gandhi2024streamsearchsoslearning}
Kanishk Gandhi, Denise Lee, Gabriel Grand, Muxin Liu, Winson Cheng, Archit Sharma, and Noah~D. Goodman.
\newblock Stream of search (sos): Learning to search in language, 2024{\natexlab{a}}.
\newblock URL \url{https://arxiv.org/abs/2404.03683}.

\bibitem[Gandhi et~al.(2024{\natexlab{b}})Gandhi, Gala, Viswanathan, Wu, and Neubig]{gandhi2024bettersyntheticdataretrieving}
Saumya Gandhi, Ritu Gala, Vijay Viswanathan, Tongshuang Wu, and Graham Neubig.
\newblock Better synthetic data by retrieving and transforming existing datasets, 2024{\natexlab{b}}.
\newblock URL \url{https://arxiv.org/abs/2404.14361}.

\bibitem[Gao et~al.(2020)Gao, Biderman, Black, Golding, Hoppe, Foster, Phang, He, Thite, Nabeshima, Presser, and Leahy]{gao2020pile800gbdatasetdiverse}
Leo Gao, Stella Biderman, Sid Black, Laurence Golding, Travis Hoppe, Charles Foster, Jason Phang, Horace He, Anish Thite, Noa Nabeshima, Shawn Presser, and Connor Leahy.
\newblock The pile: An 800gb dataset of diverse text for language modeling, 2020.
\newblock URL \url{https://arxiv.org/abs/2101.00027}.

\bibitem[Gou et~al.(2024)Gou, Shao, Gong, Shen, Yang, Duan, and Chen]{gou2024criticlargelanguagemodels}
Zhibin Gou, Zhihong Shao, Yeyun Gong, Yelong Shen, Yujiu Yang, Nan Duan, and Weizhu Chen.
\newblock Critic: Large language models can self-correct with tool-interactive critiquing, 2024.
\newblock URL \url{https://arxiv.org/abs/2305.11738}.

\bibitem[Graves et~al.(2017)Graves, Bellemare, Menick, Munos, and Kavukcuoglu]{graves2017automatedcurriculumlearningneural}
Alex Graves, Marc~G. Bellemare, Jacob Menick, Remi Munos, and Koray Kavukcuoglu.
\newblock Automated curriculum learning for neural networks, 2017.
\newblock URL \url{https://arxiv.org/abs/1704.03003}.

\bibitem[Gulcehre et~al.(2023)Gulcehre, Paine, Srinivasan, Konyushkova, Weerts, Sharma, Siddhant, Ahern, Wang, Gu, Macherey, Doucet, Firat, and de~Freitas]{gulcehre2023reinforcedselftrainingrestlanguage}
Caglar Gulcehre, Tom~Le Paine, Srivatsan Srinivasan, Ksenia Konyushkova, Lotte Weerts, Abhishek Sharma, Aditya Siddhant, Alex Ahern, Miaosen Wang, Chenjie Gu, Wolfgang Macherey, Arnaud Doucet, Orhan Firat, and Nando de~Freitas.
\newblock Reinforced self-training (rest) for language modeling, 2023.
\newblock URL \url{https://arxiv.org/abs/2308.08998}.

\bibitem[Gunasekar et~al.(2023)Gunasekar, Zhang, Aneja, Mendes, Del~Giorno, Gopi, Javaheripi, Kauffmann, de~Rosa, Saarikivi, et~al.]{gunasekar2023textbooks}
Suriya Gunasekar, Yi~Zhang, Jyoti Aneja, Caio C{\'e}sar~Teodoro Mendes, Allie Del~Giorno, Sivakanth Gopi, Mojan Javaheripi, Piero Kauffmann, Gustavo de~Rosa, Olli Saarikivi, et~al.
\newblock Textbooks are all you need.
\newblock \emph{arXiv preprint arXiv:2306.11644}, 2023.

\bibitem[Guo \& Chen(2024)Guo and Chen]{guo2024generativeaisyntheticdata}
Xu~Guo and Yiqiang Chen.
\newblock Generative ai for synthetic data generation: Methods, challenges and the future, 2024.
\newblock URL \url{https://arxiv.org/abs/2403.04190}.

\bibitem[Hafner et~al.(2024)Hafner, Pasukonis, Ba, and Lillicrap]{hafner2024masteringdiversedomainsworld}
Danijar Hafner, Jurgis Pasukonis, Jimmy Ba, and Timothy Lillicrap.
\newblock Mastering diverse domains through world models, 2024.
\newblock URL \url{https://arxiv.org/abs/2301.04104}.

\bibitem[Hashimoto et~al.(2019)Hashimoto, Zhang, and Liang]{hashimoto2019unifyinghumanstatisticalevaluation}
Tatsunori~B. Hashimoto, Hugh Zhang, and Percy Liang.
\newblock Unifying human and statistical evaluation for natural language generation, 2019.
\newblock URL \url{https://arxiv.org/abs/1904.02792}.

\bibitem[Havrilla \& Liao(2024)Havrilla and Liao]{havrilla2024understandingscalinglawsstatistical}
Alex Havrilla and Wenjing Liao.
\newblock Understanding scaling laws with statistical and approximation theory for transformer neural networks on intrinsically low-dimensional data, 2024.
\newblock URL \url{https://arxiv.org/abs/2411.06646}.

\bibitem[Havrilla et~al.(2024{\natexlab{a}})Havrilla, Du, Raparthy, Nalmpantis, Dwivedi-Yu, Zhuravinskyi, Hambro, Sukhbaatar, and Raileanu]{havrilla2024teachinglargelanguagemodels}
Alex Havrilla, Yuqing Du, Sharath~Chandra Raparthy, Christoforos Nalmpantis, Jane Dwivedi-Yu, Maksym Zhuravinskyi, Eric Hambro, Sainbayar Sukhbaatar, and Roberta Raileanu.
\newblock Teaching large language models to reason with reinforcement learning, 2024{\natexlab{a}}.
\newblock URL \url{https://arxiv.org/abs/2403.04642}.

\bibitem[Havrilla et~al.(2024{\natexlab{b}})Havrilla, Raparthy, Nalmpantis, Dwivedi-Yu, Zhuravinskyi, Hambro, and Raileanu]{havrilla2024glorewhenwhereimprove}
Alex Havrilla, Sharath Raparthy, Christoforus Nalmpantis, Jane Dwivedi-Yu, Maksym Zhuravinskyi, Eric Hambro, and Roberta Raileanu.
\newblock Glore: When, where, and how to improve llm reasoning via global and local refinements, 2024{\natexlab{b}}.
\newblock URL \url{https://arxiv.org/abs/2402.10963}.

\bibitem[Havrilla et~al.(2023)Havrilla, Zhuravinskyi, Phung, Tiwari, Tow, Biderman, Anthony, and Castricato]{havrilla-etal-2023-trlx}
Alexander Havrilla, Maksym Zhuravinskyi, Duy Phung, Aman Tiwari, Jonathan Tow, Stella Biderman, Quentin Anthony, and Louis Castricato.
\newblock trl{X}: A framework for large scale reinforcement learning from human feedback.
\newblock In \emph{Proceedings of the 2023 Conference on Empirical Methods in Natural Language Processing}, pp.\  8578--8595, Singapore, December 2023. Association for Computational Linguistics.
\newblock \doi{10.18653/v1/2023.emnlp-main.530}.
\newblock URL \url{https://aclanthology.org/2023.emnlp-main.530}.

\bibitem[He-Yueya et~al.(2024)He-Yueya, Ma, Gandhi, Domingue, Brunskill, and Goodman]{he2024psychometric}
Joy He-Yueya, Wanjing~Anya Ma, Kanishk Gandhi, Benjamin~W Domingue, Emma Brunskill, and Noah~D Goodman.
\newblock Psychometric alignment: Capturing human knowledge distributions via language models.
\newblock \emph{arXiv preprint arXiv:2407.15645}, 2024.

\bibitem[Hendrycks et~al.(2021{\natexlab{a}})Hendrycks, Burns, Basart, Zou, Mazeika, Song, and Steinhardt]{hendrycks2021measuringmassivemultitasklanguage}
Dan Hendrycks, Collin Burns, Steven Basart, Andy Zou, Mantas Mazeika, Dawn Song, and Jacob Steinhardt.
\newblock Measuring massive multitask language understanding, 2021{\natexlab{a}}.
\newblock URL \url{https://arxiv.org/abs/2009.03300}.

\bibitem[Hendrycks et~al.(2021{\natexlab{b}})Hendrycks, Burns, Kadavath, Arora, Basart, Tang, Song, and Steinhardt]{hendrycks2021measuringmathematicalproblemsolving}
Dan Hendrycks, Collin Burns, Saurav Kadavath, Akul Arora, Steven Basart, Eric Tang, Dawn Song, and Jacob Steinhardt.
\newblock Measuring mathematical problem solving with the math dataset, 2021{\natexlab{b}}.
\newblock URL \url{https://arxiv.org/abs/2103.03874}.

\bibitem[Hoffmann et~al.(2022)Hoffmann, Borgeaud, Mensch, Buchatskaya, Cai, Rutherford, de~Las~Casas, Hendricks, Welbl, Clark, Hennigan, Noland, Millican, van~den Driessche, Damoc, Guy, Osindero, Simonyan, Elsen, Rae, Vinyals, and Sifre]{hoffmann2022trainingcomputeoptimallargelanguage}
Jordan Hoffmann, Sebastian Borgeaud, Arthur Mensch, Elena Buchatskaya, Trevor Cai, Eliza Rutherford, Diego de~Las~Casas, Lisa~Anne Hendricks, Johannes Welbl, Aidan Clark, Tom Hennigan, Eric Noland, Katie Millican, George van~den Driessche, Bogdan Damoc, Aurelia Guy, Simon Osindero, Karen Simonyan, Erich Elsen, Jack~W. Rae, Oriol Vinyals, and Laurent Sifre.
\newblock Training compute-optimal large language models, 2022.
\newblock URL \url{https://arxiv.org/abs/2203.15556}.

\bibitem[Holtzman et~al.(2020)Holtzman, Buys, Du, Forbes, and Choi]{holtzman2020curiouscaseneuraltext}
Ari Holtzman, Jan Buys, Li~Du, Maxwell Forbes, and Yejin Choi.
\newblock The curious case of neural text degeneration, 2020.
\newblock URL \url{https://arxiv.org/abs/1904.09751}.

\bibitem[Honovich et~al.(2022)Honovich, Scialom, Levy, and Schick]{honovich2022unnaturalinstructionstuninglanguage}
Or~Honovich, Thomas Scialom, Omer Levy, and Timo Schick.
\newblock Unnatural instructions: Tuning language models with (almost) no human labor, 2022.
\newblock URL \url{https://arxiv.org/abs/2212.09689}.

\bibitem[Huang et~al.(2022)Huang, Gu, Hou, Wu, Wang, Yu, and Han]{huang2022largelanguagemodelsselfimprove}
Jiaxin Huang, Shixiang~Shane Gu, Le~Hou, Yuexin Wu, Xuezhi Wang, Hongkun Yu, and Jiawei Han.
\newblock Large language models can self-improve, 2022.
\newblock URL \url{https://arxiv.org/abs/2210.11610}.

\bibitem[Hughes et~al.(2024)Hughes, Dennis, Parker-Holder, Behbahani, Mavalankar, Shi, Schaul, and Rocktaschel]{hughes2024openendedness}
Edward Hughes, Michael Dennis, Jack Parker-Holder, Feryal Behbahani, Aditi Mavalankar, Yuge Shi, Tom Schaul, and Tim Rocktaschel.
\newblock Open-endedness is essential for artificial superhuman intelligence, 2024.

\bibitem[Javaheripi et~al.(2023)Javaheripi, Bubeck, Abdin, Aneja, Bubeck, Mendes, Chen, Del~Giorno, Eldan, Gopi, et~al.]{javaheripi2023phi}
Mojan Javaheripi, S{\'e}bastien Bubeck, Marah Abdin, Jyoti Aneja, Sebastien Bubeck, Caio C{\'e}sar~Teodoro Mendes, Weizhu Chen, Allie Del~Giorno, Ronen Eldan, Sivakanth Gopi, et~al.
\newblock Phi-2: The surprising power of small language models.
\newblock \emph{Microsoft Research Blog}, 2023.

\bibitem[Jiang et~al.(2023)Jiang, Rockt{\"a}schel, and Grefenstette]{jiang2023general}
Minqi Jiang, Tim Rockt{\"a}schel, and Edward Grefenstette.
\newblock General intelligence requires rethinking exploration.
\newblock \emph{Royal Society Open Science}, 10\penalty0 (6):\penalty0 230539, 2023.

\bibitem[Jiang et~al.(2024{\natexlab{a}})Jiang, Dong, Wang, Fang, Shang, Li, Jin, and Jiao]{jiang2024selfplanningcodegenerationlarge}
Xue Jiang, Yihong Dong, Lecheng Wang, Zheng Fang, Qiwei Shang, Ge~Li, Zhi Jin, and Wenpin Jiao.
\newblock Self-planning code generation with large language models, 2024{\natexlab{a}}.
\newblock URL \url{https://arxiv.org/abs/2303.06689}.

\bibitem[Jiang et~al.(2024{\natexlab{b}})Jiang, Zhou, Feng, Malladi, and Kolter]{jiang2024adaptivedataoptimizationdynamic}
Yiding Jiang, Allan Zhou, Zhili Feng, Sadhika Malladi, and J.~Zico Kolter.
\newblock Adaptive data optimization: Dynamic sample selection with scaling laws, 2024{\natexlab{b}}.
\newblock URL \url{https://arxiv.org/abs/2410.11820}.

\bibitem[Kallini et~al.(2024)Kallini, Papadimitriou, Futrell, Mahowald, and Potts]{kallini2024missionimpossible}
Julie Kallini, Isabel Papadimitriou, Richard Futrell, Kyle Mahowald, and Christopher Potts.
\newblock Mission: Impossible language models, 2024.
\newblock URL \url{https://arxiv.org/abs/2401.06416}.

\bibitem[Kaplan et~al.(2020)Kaplan, McCandlish, Henighan, Brown, Chess, Child, Gray, Radford, Wu, and Amodei]{kaplan2020scalinglawsneurallanguage}
Jared Kaplan, Sam McCandlish, Tom Henighan, Tom~B. Brown, Benjamin Chess, Rewon Child, Scott Gray, Alec Radford, Jeffrey Wu, and Dario Amodei.
\newblock Scaling laws for neural language models, 2020.
\newblock URL \url{https://arxiv.org/abs/2001.08361}.

\bibitem[Kim et~al.(2023)Kim, Bae, Shin, Kang, Kwak, Yoo, and Seo]{kim2023aligninglargelanguagemodels}
Sungdong Kim, Sanghwan Bae, Jamin Shin, Soyoung Kang, Donghyun Kwak, Kang~Min Yoo, and Minjoon Seo.
\newblock Aligning large language models through synthetic feedback, 2023.
\newblock URL \url{https://arxiv.org/abs/2305.13735}.

\bibitem[Kirk et~al.(2024)Kirk, Mediratta, Nalmpantis, Luketina, Hambro, Grefenstette, and Raileanu]{kirk2024understandingeffectsrlhfllm}
Robert Kirk, Ishita Mediratta, Christoforos Nalmpantis, Jelena Luketina, Eric Hambro, Edward Grefenstette, and Roberta Raileanu.
\newblock Understanding the effects of rlhf on llm generalisation and diversity, 2024.
\newblock URL \url{https://arxiv.org/abs/2310.06452}.

\bibitem[Korbak et~al.(2022)Korbak, Perez, and Buckley]{korbak2022rlklpenaltiesbetter}
Tomasz Korbak, Ethan Perez, and Christopher~L Buckley.
\newblock Rl with kl penalties is better viewed as bayesian inference, 2022.
\newblock URL \url{https://arxiv.org/abs/2205.11275}.

\bibitem[Kumar et~al.(2024)Kumar, Zhuang, Agarwal, Su, Co-Reyes, Singh, Baumli, Iqbal, Bishop, Roelofs, Zhang, McKinney, Shrivastava, Paduraru, Tucker, Precup, Behbahani, and Faust]{kumar2024traininglanguagemodelsselfcorrect}
Aviral Kumar, Vincent Zhuang, Rishabh Agarwal, Yi~Su, John~D Co-Reyes, Avi Singh, Kate Baumli, Shariq Iqbal, Colton Bishop, Rebecca Roelofs, Lei~M Zhang, Kay McKinney, Disha Shrivastava, Cosmin Paduraru, George Tucker, Doina Precup, Feryal Behbahani, and Aleksandra Faust.
\newblock Training language models to self-correct via reinforcement learning, 2024.
\newblock URL \url{https://arxiv.org/abs/2409.12917}.

\bibitem[Lanchantin et~al.(2023)Lanchantin, Toshniwal, Weston, Szlam, and Sukhbaatar]{lanchantin2023learningreasonmemorizeselfnotes}
Jack Lanchantin, Shubham Toshniwal, Jason Weston, Arthur Szlam, and Sainbayar Sukhbaatar.
\newblock Learning to reason and memorize with self-notes, 2023.
\newblock URL \url{https://arxiv.org/abs/2305.00833}.

\bibitem[Le et~al.(2022)Le, Wang, Gotmare, Savarese, and Hoi]{le2022coderlmasteringcodegeneration}
Hung Le, Yue Wang, Akhilesh~Deepak Gotmare, Silvio Savarese, and Steven C.~H. Hoi.
\newblock Coderl: Mastering code generation through pretrained models and deep reinforcement learning, 2022.
\newblock URL \url{https://arxiv.org/abs/2207.01780}.

\bibitem[Lee et~al.(2023)Lee, Miranda, Sundar, and Koyejo]{lee2023scalediversitycoefficientdata}
Alycia Lee, Brando Miranda, Sudharsan Sundar, and Sanmi Koyejo.
\newblock Beyond scale: the diversity coefficient as a data quality metric demonstrates llms are pre-trained on formally diverse data, 2023.
\newblock URL \url{https://arxiv.org/abs/2306.13840}.

\bibitem[Lee et~al.(2024)Lee, Cho, and Yoo]{lee2024instructiontuninghumancurriculum}
Bruce~W. Lee, Hyunsoo Cho, and Kang~Min Yoo.
\newblock Instruction tuning with human curriculum, 2024.
\newblock URL \url{https://arxiv.org/abs/2310.09518}.

\bibitem[Lee et~al.(2022)Lee, Ippolito, Nystrom, Zhang, Eck, Callison-Burch, and Carlini]{lee-etal-2022-deduplicating}
Katherine Lee, Daphne Ippolito, Andrew Nystrom, Chiyuan Zhang, Douglas Eck, Chris Callison-Burch, and Nicholas Carlini.
\newblock Deduplicating training data makes language models better.
\newblock In Smaranda Muresan, Preslav Nakov, and Aline Villavicencio (eds.), \emph{Proceedings of the 60th Annual Meeting of the Association for Computational Linguistics (Volume 1: Long Papers)}, pp.\  8424--8445, Dublin, Ireland, May 2022. Association for Computational Linguistics.
\newblock \doi{10.18653/v1/2022.acl-long.577}.
\newblock URL \url{https://aclanthology.org/2022.acl-long.577}.

\bibitem[Lehman \& Stanley(2011{\natexlab{a}})Lehman and Stanley]{lehman2011abandoning}
Joel Lehman and Kenneth~O Stanley.
\newblock Abandoning objectives: Evolution through the search for novelty alone.
\newblock \emph{Evolutionary computation}, 19\penalty0 (2):\penalty0 189--223, 2011{\natexlab{a}}.

\bibitem[Lehman \& Stanley(2011{\natexlab{b}})Lehman and Stanley]{lehman2011evolving}
Joel Lehman and Kenneth~O Stanley.
\newblock Evolving a diversity of virtual creatures through novelty search and local competition.
\newblock In \emph{Proceedings of the 13th annual conference on Genetic and evolutionary computation}, pp.\  211--218, 2011{\natexlab{b}}.

\bibitem[Lehman et~al.(2022)Lehman, Gordon, Jain, Ndousse, Yeh, and Stanley]{lehman2022evolutionlargemodels}
Joel Lehman, Jonathan Gordon, Shawn Jain, Kamal Ndousse, Cathy Yeh, and Kenneth~O. Stanley.
\newblock Evolution through large models, 2022.
\newblock URL \url{https://arxiv.org/abs/2206.08896}.

\bibitem[Li et~al.(2024{\natexlab{a}})Li, Wang, Hu, Wei, Zheng, Hu, Zhang, and Peng]{li2024common7blanguagemodels}
Chen Li, Weiqi Wang, Jingcheng Hu, Yixuan Wei, Nanning Zheng, Han Hu, Zheng Zhang, and Houwen Peng.
\newblock Common 7b language models already possess strong math capabilities, 2024{\natexlab{a}}.
\newblock URL \url{https://arxiv.org/abs/2403.04706}.

\bibitem[Li et~al.(2024{\natexlab{b}})Li, Fang, Smyrnis, Ivgi, Jordan, Gadre, Bansal, Guha, Keh, Arora, Garg, Xin, Muennighoff, Heckel, Mercat, Chen, Gururangan, Wortsman, Albalak, Bitton, Nezhurina, Abbas, Hsieh, Ghosh, Gardner, Kilian, Zhang, Shao, Pratt, Sanyal, Ilharco, Daras, Marathe, Gokaslan, Zhang, Chandu, Nguyen, Vasiljevic, Kakade, Song, Sanghavi, Faghri, Oh, Zettlemoyer, Lo, El-Nouby, Pouransari, Toshev, Wang, Groeneveld, Soldaini, Koh, Jitsev, Kollar, Dimakis, Carmon, Dave, Schmidt, and Shankar]{li2024datacomplmsearchgenerationtraining}
Jeffrey Li, Alex Fang, Georgios Smyrnis, Maor Ivgi, Matt Jordan, Samir Gadre, Hritik Bansal, Etash Guha, Sedrick Keh, Kushal Arora, Saurabh Garg, Rui Xin, Niklas Muennighoff, Reinhard Heckel, Jean Mercat, Mayee Chen, Suchin Gururangan, Mitchell Wortsman, Alon Albalak, Yonatan Bitton, Marianna Nezhurina, Amro Abbas, Cheng-Yu Hsieh, Dhruba Ghosh, Josh Gardner, Maciej Kilian, Hanlin Zhang, Rulin Shao, Sarah Pratt, Sunny Sanyal, Gabriel Ilharco, Giannis Daras, Kalyani Marathe, Aaron Gokaslan, Jieyu Zhang, Khyathi Chandu, Thao Nguyen, Igor Vasiljevic, Sham Kakade, Shuran Song, Sujay Sanghavi, Fartash Faghri, Sewoong Oh, Luke Zettlemoyer, Kyle Lo, Alaaeldin El-Nouby, Hadi Pouransari, Alexander Toshev, Stephanie Wang, Dirk Groeneveld, Luca Soldaini, Pang~Wei Koh, Jenia Jitsev, Thomas Kollar, Alexandros~G. Dimakis, Yair Carmon, Achal Dave, Ludwig Schmidt, and Vaishaal Shankar.
\newblock Datacomp-lm: In search of the next generation of training sets for language models, 2024{\natexlab{b}}.
\newblock URL \url{https://arxiv.org/abs/2406.11794}.

\bibitem[Li et~al.(2016{\natexlab{a}})Li, Galley, Brockett, Gao, and Dolan]{li-etal-2016-diversity}
Jiwei Li, Michel Galley, Chris Brockett, Jianfeng Gao, and Bill Dolan.
\newblock A diversity-promoting objective function for neural conversation models.
\newblock In Kevin Knight, Ani Nenkova, and Owen Rambow (eds.), \emph{Proceedings of the 2016 Conference of the North {A}merican Chapter of the Association for Computational Linguistics: Human Language Technologies}, pp.\  110--119, San Diego, California, June 2016{\natexlab{a}}. Association for Computational Linguistics.
\newblock \doi{10.18653/v1/N16-1014}.
\newblock URL \url{https://aclanthology.org/N16-1014}.

\bibitem[Li et~al.(2016{\natexlab{b}})Li, Galley, Brockett, Gao, and Dolan]{li2016diversitypromotingobjectivefunctionneural}
Jiwei Li, Michel Galley, Chris Brockett, Jianfeng Gao, and Bill Dolan.
\newblock A diversity-promoting objective function for neural conversation models, 2016{\natexlab{b}}.
\newblock URL \url{https://arxiv.org/abs/1510.03055}.

\bibitem[Li \& Vit{\'a}nyi(1997)Li and Vit{\'a}nyi]{li1997introduction}
Ming Li and Paul~M.B. Vit{\'a}nyi.
\newblock \emph{An Introduction to Kolmogorov Complexity and Its Applications}.
\newblock Springer-Verlag, New York, 1997.

\bibitem[Li et~al.(2024{\natexlab{c}})Li, Zhang, Li, Chen, Chen, Cheng, Wang, Zhou, and Xiao]{li2024quantityqualityboostingllm}
Ming Li, Yong Zhang, Zhitao Li, Jiuhai Chen, Lichang Chen, Ning Cheng, Jianzong Wang, Tianyi Zhou, and Jing Xiao.
\newblock From quantity to quality: Boosting llm performance with self-guided data selection for instruction tuning, 2024{\natexlab{c}}.
\newblock URL \url{https://arxiv.org/abs/2308.12032}.

\bibitem[Li et~al.(2024{\natexlab{d}})Li, Yu, Zhou, Schick, Levy, Zettlemoyer, Weston, and Lewis]{li2024selfalignmentinstructionbacktranslation}
Xian Li, Ping Yu, Chunting Zhou, Timo Schick, Omer Levy, Luke Zettlemoyer, Jason Weston, and Mike Lewis.
\newblock Self-alignment with instruction backtranslation, 2024{\natexlab{d}}.
\newblock URL \url{https://arxiv.org/abs/2308.06259}.

\bibitem[Li et~al.(2023{\natexlab{a}})Li, Holtzman, Fried, Liang, Eisner, Hashimoto, Zettlemoyer, and Lewis]{li2023contrastivedecodingopenendedtext}
Xiang~Lisa Li, Ari Holtzman, Daniel Fried, Percy Liang, Jason Eisner, Tatsunori Hashimoto, Luke Zettlemoyer, and Mike Lewis.
\newblock Contrastive decoding: Open-ended text generation as optimization, 2023{\natexlab{a}}.
\newblock URL \url{https://arxiv.org/abs/2210.15097}.

\bibitem[Li et~al.(2024{\natexlab{e}})Li, Gao, Zhang, Yue, and Hu]{li2024rule}
Xiaomin Li, Mingye Gao, Zhiwei Zhang, Chang Yue, and Hong Hu.
\newblock Rule-based data selection for large language models.
\newblock \emph{arXiv preprint arXiv:2410.04715}, 2024{\natexlab{e}}.

\bibitem[Li et~al.(2023{\natexlab{b}})Li, Bubeck, Eldan, Del~Giorno, Gunasekar, and Lee]{li2023textbooks}
Yuanzhi Li, S{\'e}bastien Bubeck, Ronen Eldan, Allie Del~Giorno, Suriya Gunasekar, and Yin~Tat Lee.
\newblock Textbooks are all you need ii: phi-1.5 technical report.
\newblock \emph{arXiv preprint arXiv:2309.05463}, 2023{\natexlab{b}}.

\bibitem[Li et~al.(2024{\natexlab{f}})Li, Chen, Xu, Qin, Xiao, Sun, and Luo]{li2024entropicdistributionmatchingsupervised}
Ziniu Li, Congliang Chen, Tian Xu, Zeyu Qin, Jiancong Xiao, Ruoyu Sun, and Zhi-Quan Luo.
\newblock Entropic distribution matching in supervised fine-tuning of llms: Less overfitting and better diversity, 2024{\natexlab{f}}.
\newblock URL \url{https://arxiv.org/abs/2408.16673}.

\bibitem[Lightman et~al.(2023)Lightman, Kosaraju, Burda, Edwards, Baker, Lee, Leike, Schulman, Sutskever, and Cobbe]{lightman2023letsverifystepstep}
Hunter Lightman, Vineet Kosaraju, Yura Burda, Harri Edwards, Bowen Baker, Teddy Lee, Jan Leike, John Schulman, Ilya Sutskever, and Karl Cobbe.
\newblock Let's verify step by step, 2023.
\newblock URL \url{https://arxiv.org/abs/2305.20050}.

\bibitem[Likert(1932)]{alma991033023919703276}
Rensis Likert.
\newblock \emph{A technique for the measurement of attitudes / by Rensis Likert.}
\newblock Archives of psychology ; no. 140. [s.n.], New York, 1932.

\bibitem[Lin(2004)]{lin-2004-rouge}
Chin-Yew Lin.
\newblock {ROUGE}: A package for automatic evaluation of summaries.
\newblock In \emph{Text Summarization Branches Out}, pp.\  74--81, Barcelona, Spain, July 2004. Association for Computational Linguistics.
\newblock URL \url{https://aclanthology.org/W04-1013}.

\bibitem[Liu et~al.(2023{\natexlab{a}})Liu, Bubeck, Eldan, Kulkarni, Li, Nguyen, Ward, and Zhang]{liu2023tinygsmachieving80gsm8k}
Bingbin Liu, Sebastien Bubeck, Ronen Eldan, Janardhan Kulkarni, Yuanzhi Li, Anh Nguyen, Rachel Ward, and Yi~Zhang.
\newblock Tinygsm: achieving >80% on gsm8k with small language models, 2023{\natexlab{a}}.
\newblock URL \url{https://arxiv.org/abs/2312.09241}.

\bibitem[Liu et~al.(2023{\natexlab{b}})Liu, Zaharia, and Abbeel]{liu2023explorationprinciplesdiverseai}
Hao Liu, Matei Zaharia, and Pieter Abbeel.
\newblock Exploration with principles for diverse ai supervision, 2023{\natexlab{b}}.
\newblock URL \url{https://arxiv.org/abs/2310.08899}.

\bibitem[Liu et~al.(2024{\natexlab{a}})Liu, Wei, Liu, Si, Zhang, Rao, Zheng, Peng, Yang, Zhou, and Dai]{liu2024bestpracticeslessonslearned}
Ruibo Liu, Jerry Wei, Fangyu Liu, Chenglei Si, Yanzhe Zhang, Jinmeng Rao, Steven Zheng, Daiyi Peng, Diyi Yang, Denny Zhou, and Andrew~M. Dai.
\newblock Best practices and lessons learned on synthetic data for language models, 2024{\natexlab{a}}.
\newblock URL \url{https://arxiv.org/abs/2404.07503}.

\bibitem[Liu et~al.(2024{\natexlab{b}})Liu, Zeng, He, Jiang, and He]{liu2024makesgooddataalignment}
Wei Liu, Weihao Zeng, Keqing He, Yong Jiang, and Junxian He.
\newblock What makes good data for alignment? a comprehensive study of automatic data selection in instruction tuning, 2024{\natexlab{b}}.
\newblock URL \url{https://arxiv.org/abs/2312.15685}.

\bibitem[Long et~al.(2024)Long, Wang, Xiao, Zhao, Ding, Chen, and Wang]{long2024llmsdrivensyntheticdatageneration}
Lin Long, Rui Wang, Ruixuan Xiao, Junbo Zhao, Xiao Ding, Gang Chen, and Haobo Wang.
\newblock On llms-driven synthetic data generation, curation, and evaluation: A survey, 2024.
\newblock URL \url{https://arxiv.org/abs/2406.15126}.

\bibitem[Longpre et~al.(2023)Longpre, Yauney, Reif, Lee, Roberts, Zoph, Zhou, Wei, Robinson, Mimno, and Ippolito]{longpre2023pretrainersguidetrainingdata}
Shayne Longpre, Gregory Yauney, Emily Reif, Katherine Lee, Adam Roberts, Barret Zoph, Denny Zhou, Jason Wei, Kevin Robinson, David Mimno, and Daphne Ippolito.
\newblock A pretrainer's guide to training data: Measuring the effects of data age, domain coverage, quality, toxicity, 2023.
\newblock URL \url{https://arxiv.org/abs/2305.13169}.

\bibitem[Lozhkov et~al.(2024)Lozhkov, Li, Allal, Cassano, Lamy-Poirier, Tazi, Tang, Pykhtar, Liu, Wei, Liu, Tian, Kocetkov, Zucker, Belkada, Wang, Liu, Abulkhanov, Paul, Li, Li, Risdal, Li, Zhu, Zhuo, Zheltonozhskii, Dade, Yu, Krauß, Jain, Su, He, Dey, Abati, Chai, Muennighoff, Tang, Oblokulov, Akiki, Marone, Mou, Mishra, Gu, Hui, Dao, Zebaze, Dehaene, Patry, Xu, McAuley, Hu, Scholak, Paquet, Robinson, Anderson, Chapados, Patwary, Tajbakhsh, Jernite, Ferrandis, Zhang, Hughes, Wolf, Guha, von Werra, and de~Vries]{lozhkov2024starcoder2stackv2}
Anton Lozhkov, Raymond Li, Loubna~Ben Allal, Federico Cassano, Joel Lamy-Poirier, Nouamane Tazi, Ao~Tang, Dmytro Pykhtar, Jiawei Liu, Yuxiang Wei, Tianyang Liu, Max Tian, Denis Kocetkov, Arthur Zucker, Younes Belkada, Zijian Wang, Qian Liu, Dmitry Abulkhanov, Indraneil Paul, Zhuang Li, Wen-Ding Li, Megan Risdal, Jia Li, Jian Zhu, Terry~Yue Zhuo, Evgenii Zheltonozhskii, Nii Osae~Osae Dade, Wenhao Yu, Lucas Krauß, Naman Jain, Yixuan Su, Xuanli He, Manan Dey, Edoardo Abati, Yekun Chai, Niklas Muennighoff, Xiangru Tang, Muhtasham Oblokulov, Christopher Akiki, Marc Marone, Chenghao Mou, Mayank Mishra, Alex Gu, Binyuan Hui, Tri Dao, Armel Zebaze, Olivier Dehaene, Nicolas Patry, Canwen Xu, Julian McAuley, Han Hu, Torsten Scholak, Sebastien Paquet, Jennifer Robinson, Carolyn~Jane Anderson, Nicolas Chapados, Mostofa Patwary, Nima Tajbakhsh, Yacine Jernite, Carlos~Muñoz Ferrandis, Lingming Zhang, Sean Hughes, Thomas Wolf, Arjun Guha, Leandro von Werra, and Harm de~Vries.
\newblock Starcoder 2 and the stack v2: The next generation, 2024.
\newblock URL \url{https://arxiv.org/abs/2402.19173}.

\bibitem[Lu et~al.(2023{\natexlab{a}})Lu, Ball, Teh, and Parker-Holder]{lu2023syntheticexperiencereplay}
Cong Lu, Philip~J. Ball, Yee~Whye Teh, and Jack Parker-Holder.
\newblock Synthetic experience replay, 2023{\natexlab{a}}.
\newblock URL \url{https://arxiv.org/abs/2303.06614}.

\bibitem[Lu et~al.(2023{\natexlab{b}})Lu, Yuan, Yuan, Lin, Lin, Tan, Zhou, and Zhou]{lu2023instaginstructiontagginganalyzing}
Keming Lu, Hongyi Yuan, Zheng Yuan, Runji Lin, Junyang Lin, Chuanqi Tan, Chang Zhou, and Jingren Zhou.
\newblock Instag: Instruction tagging for analyzing supervised fine-tuning of large language models, 2023{\natexlab{b}}.
\newblock URL \url{https://arxiv.org/abs/2308.07074}.

\bibitem[Luo et~al.(2023{\natexlab{a}})Luo, Sun, Xu, Zhao, Lou, Tao, Geng, Lin, Chen, and Zhang]{luo2023wizardmathempoweringmathematicalreasoning}
Haipeng Luo, Qingfeng Sun, Can Xu, Pu~Zhao, Jianguang Lou, Chongyang Tao, Xiubo Geng, Qingwei Lin, Shifeng Chen, and Dongmei Zhang.
\newblock Wizardmath: Empowering mathematical reasoning for large language models via reinforced evol-instruct, 2023{\natexlab{a}}.
\newblock URL \url{https://arxiv.org/abs/2308.09583}.

\bibitem[Luo et~al.(2023{\natexlab{b}})Luo, Xu, Zhao, Sun, Geng, Hu, Tao, Ma, Lin, and Jiang]{luo2023wizardcoderempoweringcodelarge}
Ziyang Luo, Can Xu, Pu~Zhao, Qingfeng Sun, Xiubo Geng, Wenxiang Hu, Chongyang Tao, Jing Ma, Qingwei Lin, and Daxin Jiang.
\newblock Wizardcoder: Empowering code large language models with evol-instruct, 2023{\natexlab{b}}.
\newblock URL \url{https://arxiv.org/abs/2306.08568}.

\bibitem[Madaan et~al.(2023)Madaan, Tandon, Gupta, Hallinan, Gao, Wiegreffe, Alon, Dziri, Prabhumoye, Yang, Gupta, Majumder, Hermann, Welleck, Yazdanbakhsh, and Clark]{madaan2023selfrefineiterativerefinementselffeedback}
Aman Madaan, Niket Tandon, Prakhar Gupta, Skyler Hallinan, Luyu Gao, Sarah Wiegreffe, Uri Alon, Nouha Dziri, Shrimai Prabhumoye, Yiming Yang, Shashank Gupta, Bodhisattwa~Prasad Majumder, Katherine Hermann, Sean Welleck, Amir Yazdanbakhsh, and Peter Clark.
\newblock Self-refine: Iterative refinement with self-feedback, 2023.
\newblock URL \url{https://arxiv.org/abs/2303.17651}.

\bibitem[Mahan et~al.(2024)Mahan, Phung, Rafailov, Blagden, Lile, Castricato, Fränken, Finn, and Albalak]{mahan2024generativerewardmodels}
Dakota Mahan, Duy~Van Phung, Rafael Rafailov, Chase Blagden, Nathan Lile, Louis Castricato, Jan-Philipp Fränken, Chelsea Finn, and Alon Albalak.
\newblock Generative reward models, 2024.
\newblock URL \url{https://arxiv.org/abs/2410.12832}.

\bibitem[Maini et~al.(2024)Maini, Seto, Bai, Grangier, Zhang, and Jaitly]{maini2024rephrasingwebrecipecompute}
Pratyush Maini, Skyler Seto, He~Bai, David Grangier, Yizhe Zhang, and Navdeep Jaitly.
\newblock Rephrasing the web: A recipe for compute and data-efficient language modeling, 2024.
\newblock URL \url{https://arxiv.org/abs/2401.16380}.

\bibitem[Mandlekar et~al.(2023)Mandlekar, Nasiriany, Wen, Akinola, Narang, Fan, Zhu, and Fox]{mandlekar2023mimicgen}
Ajay Mandlekar, Soroush Nasiriany, Bowen Wen, Iretiayo Akinola, Yashraj~S. Narang, Linxi Fan, Yuke Zhu, and Dieter Fox.
\newblock Mimicgen: {A} data generation system for scalable robot learning using human demonstrations.
\newblock In \emph{Proceedings of the Conference on Robot Learning, CoRL}, 2023.
\newblock URL \url{https://proceedings.mlr.press/v229/mandlekar23a.html}.

\bibitem[McCarthy \& Jarvis(2009)McCarthy and Jarvis]{mtld}
Philip~M. McCarthy and Scott Jarvis.
\newblock Mtld, vocd-d, and hd-d: A validation study of sophisticated approaches to lexical diversity assessment, 2009.

\bibitem[Meyerson et~al.(2023)Meyerson, Nelson, Bradley, Moradi, Hoover, and Lehman]{meyerson2023language}
Elliot Meyerson, Mark~J Nelson, Herbie Bradley, Arash Moradi, Amy~K Hoover, and Joel Lehman.
\newblock Language model crossover: Variation through few-shot prompting.
\newblock \emph{arXiv preprint arXiv:2302.12170}, 2023.

\bibitem[Miao et~al.(2023)Miao, Teh, and Rainforth]{miao2023selfcheckusingllmszeroshot}
Ning Miao, Yee~Whye Teh, and Tom Rainforth.
\newblock Selfcheck: Using llms to zero-shot check their own step-by-step reasoning, 2023.
\newblock URL \url{https://arxiv.org/abs/2308.00436}.

\bibitem[Miller et~al.(2021)Miller, Taori, Raghunathan, Sagawa, Koh, Shankar, Liang, Carmon, and Schmidt]{miller2021accuracylinestrongcorrelation}
John Miller, Rohan Taori, Aditi Raghunathan, Shiori Sagawa, Pang~Wei Koh, Vaishaal Shankar, Percy Liang, Yair Carmon, and Ludwig Schmidt.
\newblock Accuracy on the line: On the strong correlation between out-of-distribution and in-distribution generalization, 2021.
\newblock URL \url{https://arxiv.org/abs/2107.04649}.

\bibitem[Mindermann et~al.(2022)Mindermann, Brauner, Razzak, Sharma, Kirsch, Xu, Höltgen, Gomez, Morisot, Farquhar, and Gal]{mindermann2022prioritizedtrainingpointslearnable}
Sören Mindermann, Jan Brauner, Muhammed Razzak, Mrinank Sharma, Andreas Kirsch, Winnie Xu, Benedikt Höltgen, Aidan~N. Gomez, Adrien Morisot, Sebastian Farquhar, and Yarin Gal.
\newblock Prioritized training on points that are learnable, worth learning, and not yet learnt, 2022.
\newblock URL \url{https://arxiv.org/abs/2206.07137}.

\bibitem[Mitchell(2009)]{mitchell_complexity}
Melanie Mitchell.
\newblock \emph{Complexity: A Guided Tour}.
\newblock Oxford University Press, Inc., USA, 2009.
\newblock ISBN 0195124413.

\bibitem[Mouret \& Clune(2015)Mouret and Clune]{mouret2015illuminatingsearchspacesmapping}
Jean-Baptiste Mouret and Jeff Clune.
\newblock Illuminating search spaces by mapping elites, 2015.
\newblock URL \url{https://arxiv.org/abs/1504.04909}.

\bibitem[Mukherjee et~al.(2023)Mukherjee, Mitra, Jawahar, Agarwal, Palangi, and Awadallah]{mukherjee2023orcaprogressivelearningcomplex}
Subhabrata Mukherjee, Arindam Mitra, Ganesh Jawahar, Sahaj Agarwal, Hamid Palangi, and Ahmed Awadallah.
\newblock Orca: Progressive learning from complex explanation traces of gpt-4, 2023.
\newblock URL \url{https://arxiv.org/abs/2306.02707}.

\bibitem[Naik et~al.(2024)Naik, Chandrasekaran, Yuksekgonul, Palangi, and Nushi]{naik2024diversitythoughtimprovesreasoning}
Ranjita Naik, Varun Chandrasekaran, Mert Yuksekgonul, Hamid Palangi, and Besmira Nushi.
\newblock Diversity of thought improves reasoning abilities of llms, 2024.
\newblock URL \url{https://arxiv.org/abs/2310.07088}.

\bibitem[Nasir et~al.(2023)Nasir, Earle, Togelius, James, and Cleghorn]{nasir2023llmatic}
Muhammad~U Nasir, Sam Earle, Julian Togelius, Steven James, and Christopher Cleghorn.
\newblock Llmatic: Neural architecture search via large language models and quality-diversity optimization.
\newblock \emph{arXiv preprint arXiv:2306.01102}, 2023.

\bibitem[Ni et~al.(2024)Ni, Allamanis, Cohan, Deng, Shi, Sutton, and Yin]{ni2024nextteachinglargelanguage}
Ansong Ni, Miltiadis Allamanis, Arman Cohan, Yinlin Deng, Kensen Shi, Charles Sutton, and Pengcheng Yin.
\newblock Next: Teaching large language models to reason about code execution, 2024.
\newblock URL \url{https://arxiv.org/abs/2404.14662}.

\bibitem[Omura et~al.(2024)Omura, Fujita, and Kataoka]{omura2024entropycontrollabledirectpreference}
Motoki Omura, Yasuhiro Fujita, and Toshiki Kataoka.
\newblock Entropy controllable direct preference optimization, 2024.
\newblock URL \url{https://arxiv.org/abs/2411.07595}.

\bibitem[OpenAI et~al.(2024)OpenAI, Achiam, Adler, Agarwal, Ahmad, Akkaya, Aleman, Almeida, Altenschmidt, Altman, Anadkat, Avila, Babuschkin, Balaji, Balcom, Baltescu, Bao, Bavarian, Belgum, Bello, Berdine, Bernadett-Shapiro, Berner, Bogdonoff, Boiko, Boyd, Brakman, Brockman, Brooks, Brundage, Button, Cai, Campbell, Cann, Carey, Carlson, Carmichael, Chan, Chang, Chantzis, Chen, Chen, Chen, Chen, Chen, Chess, Cho, Chu, Chung, Cummings, Currier, Dai, Decareaux, Degry, Deutsch, Deville, Dhar, Dohan, Dowling, Dunning, Ecoffet, Eleti, Eloundou, Farhi, Fedus, Felix, Fishman, Forte, Fulford, Gao, Georges, Gibson, Goel, Gogineni, Goh, Gontijo-Lopes, Gordon, Grafstein, Gray, Greene, Gross, Gu, Guo, Hallacy, Han, Harris, He, Heaton, Heidecke, Hesse, Hickey, Hickey, Hoeschele, Houghton, Hsu, Hu, Hu, Huizinga, Jain, Jain, Jang, Jiang, Jiang, Jin, Jin, Jomoto, Jonn, Jun, Kaftan, Łukasz Kaiser, Kamali, Kanitscheider, Keskar, Khan, Kilpatrick, Kim, Kim, Kim, Kirchner, Kiros, Knight, Kokotajlo, Łukasz Kondraciuk, Kondrich,
  Konstantinidis, Kosic, Krueger, Kuo, Lampe, Lan, Lee, Leike, Leung, Levy, Li, Lim, Lin, Lin, Litwin, Lopez, Lowe, Lue, Makanju, Malfacini, Manning, Markov, Markovski, Martin, Mayer, Mayne, McGrew, McKinney, McLeavey, McMillan, McNeil, Medina, Mehta, Menick, Metz, Mishchenko, Mishkin, Monaco, Morikawa, Mossing, Mu, Murati, Murk, Mély, Nair, Nakano, Nayak, Neelakantan, Ngo, Noh, Ouyang, O'Keefe, Pachocki, Paino, Palermo, Pantuliano, Parascandolo, Parish, Parparita, Passos, Pavlov, Peng, Perelman, de~Avila Belbute~Peres, Petrov, de~Oliveira~Pinto, Michael, Pokorny, Pokrass, Pong, Powell, Power, Power, Proehl, Puri, Radford, Rae, Ramesh, Raymond, Real, Rimbach, Ross, Rotsted, Roussez, Ryder, Saltarelli, Sanders, Santurkar, Sastry, Schmidt, Schnurr, Schulman, Selsam, Sheppard, Sherbakov, Shieh, Shoker, Shyam, Sidor, Sigler, Simens, Sitkin, Slama, Sohl, Sokolowsky, Song, Staudacher, Such, Summers, Sutskever, Tang, Tezak, Thompson, Tillet, Tootoonchian, Tseng, Tuggle, Turley, Tworek, Uribe, Vallone, Vijayvergiya,
  Voss, Wainwright, Wang, Wang, Wang, Ward, Wei, Weinmann, Welihinda, Welinder, Weng, Weng, Wiethoff, Willner, Winter, Wolrich, Wong, Workman, Wu, Wu, Wu, Xiao, Xu, Yoo, Yu, Yuan, Zaremba, Zellers, Zhang, Zhang, Zhao, Zheng, Zhuang, Zhuk, and Zoph]{openai2024gpt4technicalreport}
OpenAI, Josh Achiam, Steven Adler, Sandhini Agarwal, Lama Ahmad, Ilge Akkaya, Florencia~Leoni Aleman, Diogo Almeida, Janko Altenschmidt, Sam Altman, Shyamal Anadkat, Red Avila, Igor Babuschkin, Suchir Balaji, Valerie Balcom, Paul Baltescu, Haiming Bao, Mohammad Bavarian, Jeff Belgum, Irwan Bello, Jake Berdine, Gabriel Bernadett-Shapiro, Christopher Berner, Lenny Bogdonoff, Oleg Boiko, Madelaine Boyd, Anna-Luisa Brakman, Greg Brockman, Tim Brooks, Miles Brundage, Kevin Button, Trevor Cai, Rosie Campbell, Andrew Cann, Brittany Carey, Chelsea Carlson, Rory Carmichael, Brooke Chan, Che Chang, Fotis Chantzis, Derek Chen, Sully Chen, Ruby Chen, Jason Chen, Mark Chen, Ben Chess, Chester Cho, Casey Chu, Hyung~Won Chung, Dave Cummings, Jeremiah Currier, Yunxing Dai, Cory Decareaux, Thomas Degry, Noah Deutsch, Damien Deville, Arka Dhar, David Dohan, Steve Dowling, Sheila Dunning, Adrien Ecoffet, Atty Eleti, Tyna Eloundou, David Farhi, Liam Fedus, Niko Felix, Simón~Posada Fishman, Juston Forte, Isabella Fulford, Leo
  Gao, Elie Georges, Christian Gibson, Vik Goel, Tarun Gogineni, Gabriel Goh, Rapha Gontijo-Lopes, Jonathan Gordon, Morgan Grafstein, Scott Gray, Ryan Greene, Joshua Gross, Shixiang~Shane Gu, Yufei Guo, Chris Hallacy, Jesse Han, Jeff Harris, Yuchen He, Mike Heaton, Johannes Heidecke, Chris Hesse, Alan Hickey, Wade Hickey, Peter Hoeschele, Brandon Houghton, Kenny Hsu, Shengli Hu, Xin Hu, Joost Huizinga, Shantanu Jain, Shawn Jain, Joanne Jang, Angela Jiang, Roger Jiang, Haozhun Jin, Denny Jin, Shino Jomoto, Billie Jonn, Heewoo Jun, Tomer Kaftan, Łukasz Kaiser, Ali Kamali, Ingmar Kanitscheider, Nitish~Shirish Keskar, Tabarak Khan, Logan Kilpatrick, Jong~Wook Kim, Christina Kim, Yongjik Kim, Jan~Hendrik Kirchner, Jamie Kiros, Matt Knight, Daniel Kokotajlo, Łukasz Kondraciuk, Andrew Kondrich, Aris Konstantinidis, Kyle Kosic, Gretchen Krueger, Vishal Kuo, Michael Lampe, Ikai Lan, Teddy Lee, Jan Leike, Jade Leung, Daniel Levy, Chak~Ming Li, Rachel Lim, Molly Lin, Stephanie Lin, Mateusz Litwin, Theresa Lopez, Ryan
  Lowe, Patricia Lue, Anna Makanju, Kim Malfacini, Sam Manning, Todor Markov, Yaniv Markovski, Bianca Martin, Katie Mayer, Andrew Mayne, Bob McGrew, Scott~Mayer McKinney, Christine McLeavey, Paul McMillan, Jake McNeil, David Medina, Aalok Mehta, Jacob Menick, Luke Metz, Andrey Mishchenko, Pamela Mishkin, Vinnie Monaco, Evan Morikawa, Daniel Mossing, Tong Mu, Mira Murati, Oleg Murk, David Mély, Ashvin Nair, Reiichiro Nakano, Rajeev Nayak, Arvind Neelakantan, Richard Ngo, Hyeonwoo Noh, Long Ouyang, Cullen O'Keefe, Jakub Pachocki, Alex Paino, Joe Palermo, Ashley Pantuliano, Giambattista Parascandolo, Joel Parish, Emy Parparita, Alex Passos, Mikhail Pavlov, Andrew Peng, Adam Perelman, Filipe de~Avila Belbute~Peres, Michael Petrov, Henrique~Ponde de~Oliveira~Pinto, Michael, Pokorny, Michelle Pokrass, Vitchyr~H. Pong, Tolly Powell, Alethea Power, Boris Power, Elizabeth Proehl, Raul Puri, Alec Radford, Jack Rae, Aditya Ramesh, Cameron Raymond, Francis Real, Kendra Rimbach, Carl Ross, Bob Rotsted, Henri Roussez,
  Nick Ryder, Mario Saltarelli, Ted Sanders, Shibani Santurkar, Girish Sastry, Heather Schmidt, David Schnurr, John Schulman, Daniel Selsam, Kyla Sheppard, Toki Sherbakov, Jessica Shieh, Sarah Shoker, Pranav Shyam, Szymon Sidor, Eric Sigler, Maddie Simens, Jordan Sitkin, Katarina Slama, Ian Sohl, Benjamin Sokolowsky, Yang Song, Natalie Staudacher, Felipe~Petroski Such, Natalie Summers, Ilya Sutskever, Jie Tang, Nikolas Tezak, Madeleine~B. Thompson, Phil Tillet, Amin Tootoonchian, Elizabeth Tseng, Preston Tuggle, Nick Turley, Jerry Tworek, Juan Felipe~Cerón Uribe, Andrea Vallone, Arun Vijayvergiya, Chelsea Voss, Carroll Wainwright, Justin~Jay Wang, Alvin Wang, Ben Wang, Jonathan Ward, Jason Wei, CJ~Weinmann, Akila Welihinda, Peter Welinder, Jiayi Weng, Lilian Weng, Matt Wiethoff, Dave Willner, Clemens Winter, Samuel Wolrich, Hannah Wong, Lauren Workman, Sherwin Wu, Jeff Wu, Michael Wu, Kai Xiao, Tao Xu, Sarah Yoo, Kevin Yu, Qiming Yuan, Wojciech Zaremba, Rowan Zellers, Chong Zhang, Marvin Zhang, Shengjia
  Zhao, Tianhao Zheng, Juntang Zhuang, William Zhuk, and Barret Zoph.
\newblock Gpt-4 technical report, 2024.
\newblock URL \url{https://arxiv.org/abs/2303.08774}.

\bibitem[{Ortiz Su{\'a}rez} et~al.(2019){Ortiz Su{\'a}rez}, Sagot, and Romary]{OrtizSuarezSagotRomary2019}
Pedro~Javier {Ortiz Su{\'a}rez}, Beno{\^i}t Sagot, and Laurent Romary.
\newblock Asynchronous pipelines for processing huge corpora on medium to low resource infrastructures.
\newblock In Piotr Bański, Adrien Barbaresi, Hanno Biber, Evelyn Breiteneder, Simon Clematide, Marc Kupietz, Harald L{\"u}ngen, and Caroline Iliadi (eds.), \emph{Proceedings of the Workshop on Challenges in the Management of Large Corpora}, Proceedings of the Workshop on Challenges in the Management of Large Corpora (CMLC-7) 2019. Cardiff, 22nd July 2019, pp.\  9 -- 16, Mannheim, 2019. Leibniz-Institut f{\"u}r Deutsche Sprache.
\newblock \doi{10.14618/ids-pub-9021}.
\newblock URL \url{http://nbn-resolving.de/urn:nbn:de:bsz:mh39-90215}.

\bibitem[Ouyang et~al.(2022)Ouyang, Wu, Jiang, Almeida, Wainwright, Mishkin, Zhang, Agarwal, Slama, Ray, Schulman, Hilton, Kelton, Miller, Simens, Askell, Welinder, Christiano, Leike, and Lowe]{ouyang2022traininglanguagemodelsfollow}
Long Ouyang, Jeff Wu, Xu~Jiang, Diogo Almeida, Carroll~L. Wainwright, Pamela Mishkin, Chong Zhang, Sandhini Agarwal, Katarina Slama, Alex Ray, John Schulman, Jacob Hilton, Fraser Kelton, Luke Miller, Maddie Simens, Amanda Askell, Peter Welinder, Paul Christiano, Jan Leike, and Ryan Lowe.
\newblock Training language models to follow instructions with human feedback, 2022.
\newblock URL \url{https://arxiv.org/abs/2203.02155}.

\bibitem[O’Mahony et~al.(2024)O’Mahony, Grinsztajn, Schoelkopf, and Biderman]{oattributing}
Laura O’Mahony, Leo Grinsztajn, Hailey Schoelkopf, and Stella Biderman.
\newblock Attributing mode collapse in the fine-tuning of large language models.
\newblock In \emph{ICLR 2024 Mathematical and Empirical Understanding of Foundation Models (ME-FoMo) workshop}, 2024.

\bibitem[Pace et~al.(2024)Pace, Mallinson, Malmi, Krause, and Severyn]{pace2024westofnsyntheticpreferencegeneration}
Alizée Pace, Jonathan Mallinson, Eric Malmi, Sebastian Krause, and Aliaksei Severyn.
\newblock West-of-n: Synthetic preference generation for improved reward modeling, 2024.
\newblock URL \url{https://arxiv.org/abs/2401.12086}.

\bibitem[Packard et~al.(2019)Packard, Bedau, Channon, Ikegami, Rasmussen, Stanley, and Taylor]{packard2019overview}
Norman Packard, Mark~A Bedau, Alastair Channon, Takashi Ikegami, Steen Rasmussen, Kenneth~O Stanley, and Tim Taylor.
\newblock An overview of open-ended evolution: Editorial introduction to the open-ended evolution ii special issue.
\newblock \emph{Artificial life}, 25\penalty0 (2):\penalty0 93--103, 2019.

\bibitem[Pandey(2024)]{pandey2024gzippredictsdatadependentscaling}
Rohan Pandey.
\newblock gzip predicts data-dependent scaling laws, 2024.
\newblock URL \url{https://arxiv.org/abs/2405.16684}.

\bibitem[Pang et~al.(2024)Pang, Yuan, Cho, He, Sukhbaatar, and Weston]{pang2024iterativereasoningpreferenceoptimization}
Richard~Yuanzhe Pang, Weizhe Yuan, Kyunghyun Cho, He~He, Sainbayar Sukhbaatar, and Jason Weston.
\newblock Iterative reasoning preference optimization, 2024.
\newblock URL \url{https://arxiv.org/abs/2404.19733}.

\bibitem[Papineni et~al.(2002{\natexlab{a}})Papineni, Roukos, Ward, and Zhu]{Papineni2002BleuAM}
Kishore Papineni, Salim Roukos, Todd Ward, and Wei-Jing Zhu.
\newblock Bleu: a method for automatic evaluation of machine translation.
\newblock In \emph{Annual Meeting of the Association for Computational Linguistics}, 2002{\natexlab{a}}.
\newblock URL \url{https://api.semanticscholar.org/CorpusID:11080756}.

\bibitem[Papineni et~al.(2002{\natexlab{b}})Papineni, Roukos, Ward, and Zhu]{papineni2002bleu}
Kishore Papineni, Salim Roukos, Todd Ward, and Wei-Jing Zhu.
\newblock Bleu: a method for automatic evaluation of machine translation.
\newblock In \emph{Proceedings of the 40th annual meeting of the Association for Computational Linguistics}, pp.\  311--318, 2002{\natexlab{b}}.

\bibitem[Penedo et~al.(2024)Penedo, Kydlíček, allal, Lozhkov, Mitchell, Raffel, Werra, and Wolf]{penedo2024finewebdatasetsdecantingweb}
Guilherme Penedo, Hynek Kydlíček, Loubna~Ben allal, Anton Lozhkov, Margaret Mitchell, Colin Raffel, Leandro~Von Werra, and Thomas Wolf.
\newblock The fineweb datasets: Decanting the web for the finest text data at scale, 2024.
\newblock URL \url{https://arxiv.org/abs/2406.17557}.

\bibitem[Pillutla et~al.(2021)Pillutla, Swayamdipta, Zellers, Thickstun, Welleck, Choi, and Harchaoui]{pillutla2021mauve}
Krishna Pillutla, Swabha Swayamdipta, Rowan Zellers, John Thickstun, Sean Welleck, Yejin Choi, and Zaid Harchaoui.
\newblock Mauve: Measuring the gap between neural text and human text using divergence frontiers.
\newblock \emph{Advances in Neural Information Processing Systems}, 34:\penalty0 4816--4828, 2021.

\bibitem[Polo et~al.(2024)Polo, Weber, Choshen, Sun, Xu, and Yurochkin]{polo2024tinybenchmarks}
Felipe~Maia Polo, Lucas Weber, Leshem Choshen, Yuekai Sun, Gongjun Xu, and Mikhail Yurochkin.
\newblock tinybenchmarks: evaluating llms with fewer examples.
\newblock \emph{arXiv preprint arXiv:2402.14992}, 2024.

\bibitem[Pourcel et~al.(2024)Pourcel, Colas, Molinaro, Oudeyer, and Teodorescu]{pourcel2024acesgeneratingdiverseprogramming}
Julien Pourcel, Cédric Colas, Gaia Molinaro, Pierre-Yves Oudeyer, and Laetitia Teodorescu.
\newblock Aces: Generating diverse programming puzzles with with autotelic generative models, 2024.
\newblock URL \url{https://arxiv.org/abs/2310.10692}.

\bibitem[Pugh et~al.(2016)Pugh, Soros, and Stanley]{pugh2016quality}
Justin~K Pugh, Lisa~B Soros, and Kenneth~O Stanley.
\newblock Quality diversity: A new frontier for evolutionary computation.
\newblock \emph{Frontiers in Robotics and AI}, 3:\penalty0 40, 2016.

\bibitem[Qin et~al.(2024)Qin, Yang, Guo, Li, Shao, Shi, Xu, Gu, Li, and Sun]{qin2024unleashingpowerdatatsunami}
Yulei Qin, Yuncheng Yang, Pengcheng Guo, Gang Li, Hang Shao, Yuchen Shi, Zihan Xu, Yun Gu, Ke~Li, and Xing Sun.
\newblock Unleashing the power of data tsunami: A comprehensive survey on data assessment and selection for instruction tuning of language models, 2024.
\newblock URL \url{https://arxiv.org/abs/2408.02085}.

\bibitem[Qu et~al.(2024)Qu, Zhang, Garg, and Kumar]{qu2024recursive}
Yuxiao Qu, Tianjun Zhang, Naman Garg, and Aviral Kumar.
\newblock Recursive introspection: Teaching {LLM} agents how to self-improve.
\newblock In \emph{ICML 2024 Workshop on LLMs and Cognition}, 2024.
\newblock URL \url{https://openreview.net/forum?id=ze0nodITam}.

\bibitem[Rafailov et~al.(2024)Rafailov, Sharma, Mitchell, Ermon, Manning, and Finn]{rafailov2024directpreferenceoptimizationlanguage}
Rafael Rafailov, Archit Sharma, Eric Mitchell, Stefano Ermon, Christopher~D. Manning, and Chelsea Finn.
\newblock Direct preference optimization: Your language model is secretly a reward model, 2024.
\newblock URL \url{https://arxiv.org/abs/2305.18290}.

\bibitem[Raffel et~al.(2020)Raffel, Shazeer, Roberts, Lee, Narang, Matena, Zhou, Li, and Liu]{Raffel2019ExploringTL}
Colin Raffel, Noam Shazeer, Adam Roberts, Katherine Lee, Sharan Narang, Michael Matena, Yanqi Zhou, Wei Li, and Peter~J. Liu.
\newblock Exploring the limits of transfer learning with a unified text-to-text transformer.
\newblock \emph{J. Mach. Learn. Res.}, 21\penalty0 (1), jan 2020.
\newblock ISSN 1532-4435.
\newblock URL \url{https://jmlr.org/papers/volume21/20-074/20-074.pdf}.

\bibitem[Rasch(1993)]{rasch1993probabilistic}
Georg Rasch.
\newblock \emph{Probabilistic models for some intelligence and attainment tests.}
\newblock ERIC, 1993.

\bibitem[Reimers \& Gurevych(2019)Reimers and Gurevych]{reimers2019sentencebertsentenceembeddingsusing}
Nils Reimers and Iryna Gurevych.
\newblock Sentence-bert: Sentence embeddings using siamese bert-networks, 2019.
\newblock URL \url{https://arxiv.org/abs/1908.10084}.

\bibitem[Rein et~al.(2023)Rein, Hou, Stickland, Petty, Pang, Dirani, Michael, and Bowman]{rein2023gpqagraduatelevelgoogleproofqa}
David Rein, Betty~Li Hou, Asa~Cooper Stickland, Jackson Petty, Richard~Yuanzhe Pang, Julien Dirani, Julian Michael, and Samuel~R. Bowman.
\newblock Gpqa: A graduate-level google-proof qa benchmark, 2023.
\newblock URL \url{https://arxiv.org/abs/2311.12022}.

\bibitem[Reynolds \& McDonell(2021)Reynolds and McDonell]{reynolds2021promptprogramminglargelanguage}
Laria Reynolds and Kyle McDonell.
\newblock Prompt programming for large language models: Beyond the few-shot paradigm, 2021.
\newblock URL \url{https://arxiv.org/abs/2102.07350}.

\bibitem[Romera-Paredes et~al.(2024)Romera-Paredes, Barekatain, Novikov, Balog, Kumar, Dupont, Ruiz, Ellenberg, Wang, Fawzi, Kohli, and Fawzi]{Romera-Paredes2024}
Bernardino Romera-Paredes, Mohammadamin Barekatain, Alexander Novikov, Matej Balog, M.~Pawan Kumar, Emilien Dupont, Francisco J.~R. Ruiz, 1~Jordan~S. Ellenberg, Pengming Wang, Omar Fawzi, Pushmeet Kohli, and Alhussein Fawzi.
\newblock Mathematical discoveries from program search with large language models.
\newblock \emph{Nature}, 625\penalty0 (7995):\penalty0 468--475, 2024.
\newblock \doi{10.1038/s41586-023-06924-6}.
\newblock URL \url{https://doi.org/10.1038/s41586-023-06924-6}.

\bibitem[Ruan et~al.(2024)Ruan, Maddison, and Hashimoto]{ruan2024observationalscalinglawspredictability}
Yangjun Ruan, Chris~J. Maddison, and Tatsunori Hashimoto.
\newblock Observational scaling laws and the predictability of language model performance, 2024.
\newblock URL \url{https://arxiv.org/abs/2405.10938}.

\bibitem[Russakovsky et~al.(2015)Russakovsky, Deng, Su, Krause, Satheesh, Ma, Huang, Karpathy, Khosla, Bernstein, Berg, and Fei-Fei]{russakovsky2015imagenetlargescalevisual}
Olga Russakovsky, Jia Deng, Hao Su, Jonathan Krause, Sanjeev Satheesh, Sean Ma, Zhiheng Huang, Andrej Karpathy, Aditya Khosla, Michael Bernstein, Alexander~C. Berg, and Li~Fei-Fei.
\newblock Imagenet large scale visual recognition challenge, 2015.
\newblock URL \url{https://arxiv.org/abs/1409.0575}.

\bibitem[Samvelyan et~al.(2024{\natexlab{a}})Samvelyan, Paglieri, Jiang, Parker-Holder, and Rockt{\"a}schel]{samvelyan2024multi}
Mikayel Samvelyan, Davide Paglieri, Minqi Jiang, Jack Parker-Holder, and Tim Rockt{\"a}schel.
\newblock Multi-agent diagnostics for robustness via illuminated diversity.
\newblock \emph{arXiv preprint arXiv:2401.13460}, 2024{\natexlab{a}}.

\bibitem[Samvelyan et~al.(2024{\natexlab{b}})Samvelyan, Raparthy, Lupu, Hambro, Markosyan, Bhatt, Mao, Jiang, Parker-Holder, Foerster, Rocktäschel, and Raileanu]{samvelyan2024rainbowteamingopenendedgeneration}
Mikayel Samvelyan, Sharath~Chandra Raparthy, Andrei Lupu, Eric Hambro, Aram~H. Markosyan, Manish Bhatt, Yuning Mao, Minqi Jiang, Jack Parker-Holder, Jakob Foerster, Tim Rocktäschel, and Roberta Raileanu.
\newblock Rainbow teaming: Open-ended generation of diverse adversarial prompts, 2024{\natexlab{b}}.
\newblock URL \url{https://arxiv.org/abs/2402.16822}.

\bibitem[Sawada et~al.(2023)Sawada, Paleka, Havrilla, Tadepalli, Vidas, Kranias, Nay, Gupta, and Komatsuzaki]{sawada2023arbadvancedreasoningbenchmark}
Tomohiro Sawada, Daniel Paleka, Alexander Havrilla, Pranav Tadepalli, Paula Vidas, Alexander Kranias, John~J. Nay, Kshitij Gupta, and Aran Komatsuzaki.
\newblock Arb: Advanced reasoning benchmark for large language models, 2023.
\newblock URL \url{https://arxiv.org/abs/2307.13692}.

\bibitem[Schick et~al.(2023)Schick, Dwivedi-Yu, Dessì, Raileanu, Lomeli, Zettlemoyer, Cancedda, and Scialom]{schick2023toolformerlanguagemodelsteach}
Timo Schick, Jane Dwivedi-Yu, Roberto Dessì, Roberta Raileanu, Maria Lomeli, Luke Zettlemoyer, Nicola Cancedda, and Thomas Scialom.
\newblock Toolformer: Language models can teach themselves to use tools, 2023.
\newblock URL \url{https://arxiv.org/abs/2302.04761}.

\bibitem[Schulman et~al.(2017)Schulman, Wolski, Dhariwal, Radford, and Klimov]{schulman2017proximalpolicyoptimizationalgorithms}
John Schulman, Filip Wolski, Prafulla Dhariwal, Alec Radford, and Oleg Klimov.
\newblock Proximal policy optimization algorithms, 2017.
\newblock URL \url{https://arxiv.org/abs/1707.06347}.

\bibitem[Sennrich et~al.(2016)Sennrich, Haddow, and Birch]{sennrich-etal-2016-improving}
Rico Sennrich, Barry Haddow, and Alexandra Birch.
\newblock Improving neural machine translation models with monolingual data.
\newblock In Katrin Erk and Noah~A. Smith (eds.), \emph{Proceedings of the 54th Annual Meeting of the Association for Computational Linguistics (Volume 1: Long Papers)}, pp.\  86--96, Berlin, Germany, August 2016. Association for Computational Linguistics.
\newblock \doi{10.18653/v1/P16-1009}.
\newblock URL \url{https://aclanthology.org/P16-1009}.

\bibitem[Setlur et~al.(2024)Setlur, Garg, Geng, Garg, Smith, and Kumar]{setlur2024rlincorrectsyntheticdata}
Amrith Setlur, Saurabh Garg, Xinyang Geng, Naman Garg, Virginia Smith, and Aviral Kumar.
\newblock Rl on incorrect synthetic data scales the efficiency of llm math reasoning by eight-fold, 2024.
\newblock URL \url{https://arxiv.org/abs/2406.14532}.

\bibitem[Shao et~al.(2024)Shao, Wang, Zhu, Xu, Song, Bi, Zhang, Zhang, Li, Wu, and Guo]{shao2024deepseekmathpushinglimitsmathematical}
Zhihong Shao, Peiyi Wang, Qihao Zhu, Runxin Xu, Junxiao Song, Xiao Bi, Haowei Zhang, Mingchuan Zhang, Y.~K. Li, Y.~Wu, and Daya Guo.
\newblock Deepseekmath: Pushing the limits of mathematical reasoning in open language models, 2024.
\newblock URL \url{https://arxiv.org/abs/2402.03300}.

\bibitem[Sharma \& Kaplan(2020)Sharma and Kaplan]{sharma2020neuralscalinglawdimension}
Utkarsh Sharma and Jared Kaplan.
\newblock A neural scaling law from the dimension of the data manifold, 2020.
\newblock URL \url{https://arxiv.org/abs/2004.10802}.

\bibitem[Sharma et~al.(2024)Sharma, Padthe, Ardalani, Tirumala, Howes, Xu, Huang, Li, Aghajanyan, Ghosh, and Zettlemoyer]{sharma2024textqualitybasedpruningefficient}
Vasu Sharma, Karthik Padthe, Newsha Ardalani, Kushal Tirumala, Russell Howes, Hu~Xu, Po-Yao Huang, Shang-Wen Li, Armen Aghajanyan, Gargi Ghosh, and Luke Zettlemoyer.
\newblock Text quality-based pruning for efficient training of language models, 2024.
\newblock URL \url{https://arxiv.org/abs/2405.01582}.

\bibitem[Shinn et~al.(2023)Shinn, Cassano, Berman, Gopinath, Narasimhan, and Yao]{shinn2023reflexionlanguageagentsverbal}
Noah Shinn, Federico Cassano, Edward Berman, Ashwin Gopinath, Karthik Narasimhan, and Shunyu Yao.
\newblock Reflexion: Language agents with verbal reinforcement learning, 2023.
\newblock URL \url{https://arxiv.org/abs/2303.11366}.

\bibitem[Sigaud et~al.(2023)Sigaud, Baldassarre, Colas, Doncieux, Duro, Perrin-Gilbert, and Santucci]{sigaud2023definition}
Olivier Sigaud, Gianluca Baldassarre, Cedric Colas, Stephane Doncieux, Richard Duro, Nicolas Perrin-Gilbert, and Vieri-Giuliano Santucci.
\newblock A definition of open-ended learning problems for goal-conditioned agents.
\newblock \emph{arXiv preprint arXiv:2311.00344}, 2023.

\bibitem[Singh et~al.(2024)Singh, Co-Reyes, Agarwal, Anand, Patil, Garcia, Liu, Harrison, Lee, Xu, Parisi, Kumar, Alemi, Rizkowsky, Nova, Adlam, Bohnet, Elsayed, Sedghi, Mordatch, Simpson, Gur, Snoek, Pennington, Hron, Kenealy, Swersky, Mahajan, Culp, Xiao, Bileschi, Constant, Novak, Liu, Warkentin, Qian, Bansal, Dyer, Neyshabur, Sohl-Dickstein, and Fiedel]{singh2024humandatascalingselftraining}
Avi Singh, John~D. Co-Reyes, Rishabh Agarwal, Ankesh Anand, Piyush Patil, Xavier Garcia, Peter~J. Liu, James Harrison, Jaehoon Lee, Kelvin Xu, Aaron Parisi, Abhishek Kumar, Alex Alemi, Alex Rizkowsky, Azade Nova, Ben Adlam, Bernd Bohnet, Gamaleldin Elsayed, Hanie Sedghi, Igor Mordatch, Isabelle Simpson, Izzeddin Gur, Jasper Snoek, Jeffrey Pennington, Jiri Hron, Kathleen Kenealy, Kevin Swersky, Kshiteej Mahajan, Laura Culp, Lechao Xiao, Maxwell~L. Bileschi, Noah Constant, Roman Novak, Rosanne Liu, Tris Warkentin, Yundi Qian, Yamini Bansal, Ethan Dyer, Behnam Neyshabur, Jascha Sohl-Dickstein, and Noah Fiedel.
\newblock Beyond human data: Scaling self-training for problem-solving with language models, 2024.
\newblock URL \url{https://arxiv.org/abs/2312.06585}.

\bibitem[Snell et~al.(2024)Snell, Lee, Xu, and Kumar]{snell2024scalingllmtesttimecompute}
Charlie Snell, Jaehoon Lee, Kelvin Xu, and Aviral Kumar.
\newblock Scaling llm test-time compute optimally can be more effective than scaling model parameters, 2024.
\newblock URL \url{https://arxiv.org/abs/2408.03314}.

\bibitem[Song(2022)]{song2022little}
Asiiah Song.
\newblock A little taxonomy of open-endedness.
\newblock In \emph{ICLR Workshop on Agent Learning in Open-Endedness}, 2022.

\bibitem[Soros et~al.(2017)Soros, Lehman, and Stanley]{soros2017open}
Lisa~B Soros, Joel Lehman, and Kenneth~O Stanley.
\newblock Open-endedness: The last grand challenge you’ve never heard of, 2017.

\bibitem[Stiennon et~al.(2022)Stiennon, Ouyang, Wu, Ziegler, Lowe, Voss, Radford, Amodei, and Christiano]{stiennon2022learningsummarizehumanfeedback}
Nisan Stiennon, Long Ouyang, Jeff Wu, Daniel~M. Ziegler, Ryan Lowe, Chelsea Voss, Alec Radford, Dario Amodei, and Paul Christiano.
\newblock Learning to summarize from human feedback, 2022.
\newblock URL \url{https://arxiv.org/abs/2009.01325}.

\bibitem[Sudhakaran et~al.(2024)Sudhakaran, Gonz{\'a}lez-Duque, Freiberger, Glanois, Najarro, and Risi]{sudhakaran2024mariogpt}
Shyam Sudhakaran, Miguel Gonz{\'a}lez-Duque, Matthias Freiberger, Claire Glanois, Elias Najarro, and Sebastian Risi.
\newblock Mariogpt: Open-ended text2level generation through large language models.
\newblock \emph{Advances in Neural Information Processing Systems}, 36, 2024.

\bibitem[Sun et~al.(2024)Sun, Yu, Shen, Liu, Yang, Welleck, and Gan]{sun2024easytohardgeneralizationscalablealignment}
Zhiqing Sun, Longhui Yu, Yikang Shen, Weiyang Liu, Yiming Yang, Sean Welleck, and Chuang Gan.
\newblock Easy-to-hard generalization: Scalable alignment beyond human supervision, 2024.
\newblock URL \url{https://arxiv.org/abs/2403.09472}.

\bibitem[Team et~al.(2021)Team, Stooke, Mahajan, Barros, Deck, Bauer, Sygnowski, Trebacz, Jaderberg, Mathieu, McAleese, Bradley-Schmieg, Wong, Porcel, Raileanu, Hughes-Fitt, Dalibard, and Czarnecki]{openendedlearningteam2021openendedlearningleadsgenerally}
Open Ended~Learning Team, Adam Stooke, Anuj Mahajan, Catarina Barros, Charlie Deck, Jakob Bauer, Jakub Sygnowski, Maja Trebacz, Max Jaderberg, Michael Mathieu, Nat McAleese, Nathalie Bradley-Schmieg, Nathaniel Wong, Nicolas Porcel, Roberta Raileanu, Steph Hughes-Fitt, Valentin Dalibard, and Wojciech~Marian Czarnecki.
\newblock Open-ended learning leads to generally capable agents, 2021.
\newblock URL \url{https://arxiv.org/abs/2107.12808}.

\bibitem[Thissen(1982)]{thissen1982marginal}
David Thissen.
\newblock Marginal maximum likelihood estimation for the one-parameter logistic model.
\newblock \emph{Psychometrika}, 47:\penalty0 175--186, 1982.

\bibitem[Tian et~al.(2024)Tian, Peng, Song, Jin, Yu, Mi, and Yu]{tian2024selfimprovementllmsimaginationsearching}
Ye~Tian, Baolin Peng, Linfeng Song, Lifeng Jin, Dian Yu, Haitao Mi, and Dong Yu.
\newblock Toward self-improvement of llms via imagination, searching, and criticizing, 2024.
\newblock URL \url{https://arxiv.org/abs/2404.12253}.

\bibitem[Tirumala et~al.(2023)Tirumala, Simig, Aghajanyan, and Morcos]{tirumala2023d4improvingllmpretraining}
Kushal Tirumala, Daniel Simig, Armen Aghajanyan, and Ari~S. Morcos.
\newblock D4: Improving llm pretraining via document de-duplication and diversification, 2023.
\newblock URL \url{https://arxiv.org/abs/2308.12284}.

\bibitem[Tjanaka(2022)]{tjanaka2022quantifying}
Bryon Tjanaka.
\newblock Quantifying efficiency in quality diversity optimization.
\newblock In \emph{Workshop on Benchmarks for Quality-Diversity Algorithms at GECCO}, 2022.

\bibitem[Toshniwal et~al.(2024)Toshniwal, Moshkov, Narenthiran, Gitman, Jia, and Gitman]{toshniwal2024openmathinstruct118millionmath}
Shubham Toshniwal, Ivan Moshkov, Sean Narenthiran, Daria Gitman, Fei Jia, and Igor Gitman.
\newblock Openmathinstruct-1: A 1.8 million math instruction tuning dataset, 2024.
\newblock URL \url{https://arxiv.org/abs/2402.10176}.

\bibitem[Touvron et~al.(2023{\natexlab{a}})Touvron, Lavril, Izacard, Martinet, Lachaux, Lacroix, Rozière, Goyal, Hambro, Azhar, Rodriguez, Joulin, Grave, and Lample]{touvron2023llamaopenefficientfoundation}
Hugo Touvron, Thibaut Lavril, Gautier Izacard, Xavier Martinet, Marie-Anne Lachaux, Timothée Lacroix, Baptiste Rozière, Naman Goyal, Eric Hambro, Faisal Azhar, Aurelien Rodriguez, Armand Joulin, Edouard Grave, and Guillaume Lample.
\newblock Llama: Open and efficient foundation language models, 2023{\natexlab{a}}.
\newblock URL \url{https://arxiv.org/abs/2302.13971}.

\bibitem[Touvron et~al.(2023{\natexlab{b}})Touvron, Martin, Stone, Albert, Almahairi, Babaei, Bashlykov, Batra, Bhargava, Bhosale, Bikel, Blecher, Ferrer, Chen, Cucurull, Esiobu, Fernandes, Fu, Fu, Fuller, Gao, Goswami, Goyal, Hartshorn, Hosseini, Hou, Inan, Kardas, Kerkez, Khabsa, Kloumann, Korenev, Koura, Lachaux, Lavril, Lee, Liskovich, Lu, Mao, Martinet, Mihaylov, Mishra, Molybog, Nie, Poulton, Reizenstein, Rungta, Saladi, Schelten, Silva, Smith, Subramanian, Tan, Tang, Taylor, Williams, Kuan, Xu, Yan, Zarov, Zhang, Fan, Kambadur, Narang, Rodriguez, Stojnic, Edunov, and Scialom]{touvron2023llama2openfoundation}
Hugo Touvron, Louis Martin, Kevin Stone, Peter Albert, Amjad Almahairi, Yasmine Babaei, Nikolay Bashlykov, Soumya Batra, Prajjwal Bhargava, Shruti Bhosale, Dan Bikel, Lukas Blecher, Cristian~Canton Ferrer, Moya Chen, Guillem Cucurull, David Esiobu, Jude Fernandes, Jeremy Fu, Wenyin Fu, Brian Fuller, Cynthia Gao, Vedanuj Goswami, Naman Goyal, Anthony Hartshorn, Saghar Hosseini, Rui Hou, Hakan Inan, Marcin Kardas, Viktor Kerkez, Madian Khabsa, Isabel Kloumann, Artem Korenev, Punit~Singh Koura, Marie-Anne Lachaux, Thibaut Lavril, Jenya Lee, Diana Liskovich, Yinghai Lu, Yuning Mao, Xavier Martinet, Todor Mihaylov, Pushkar Mishra, Igor Molybog, Yixin Nie, Andrew Poulton, Jeremy Reizenstein, Rashi Rungta, Kalyan Saladi, Alan Schelten, Ruan Silva, Eric~Michael Smith, Ranjan Subramanian, Xiaoqing~Ellen Tan, Binh Tang, Ross Taylor, Adina Williams, Jian~Xiang Kuan, Puxin Xu, Zheng Yan, Iliyan Zarov, Yuchen Zhang, Angela Fan, Melanie Kambadur, Sharan Narang, Aurelien Rodriguez, Robert Stojnic, Sergey Edunov, and Thomas
  Scialom.
\newblock Llama 2: Open foundation and fine-tuned chat models, 2023{\natexlab{b}}.
\newblock URL \url{https://arxiv.org/abs/2307.09288}.

\bibitem[Uesato et~al.(2022)Uesato, Kushman, Kumar, Song, Siegel, Wang, Creswell, Irving, and Higgins]{uesato2022solvingmathwordproblems}
Jonathan Uesato, Nate Kushman, Ramana Kumar, Francis Song, Noah Siegel, Lisa Wang, Antonia Creswell, Geoffrey Irving, and Irina Higgins.
\newblock Solving math word problems with process- and outcome-based feedback, 2022.
\newblock URL \url{https://arxiv.org/abs/2211.14275}.

\bibitem[Viswanathan et~al.(2023)Viswanathan, Zhao, Bertsch, Wu, and Neubig]{viswanathan2023prompt2modelgeneratingdeployablemodels}
Vijay Viswanathan, Chenyang Zhao, Amanda Bertsch, Tongshuang Wu, and Graham Neubig.
\newblock Prompt2model: Generating deployable models from natural language instructions, 2023.
\newblock URL \url{https://arxiv.org/abs/2308.12261}.

\bibitem[Wang \& Komatsuzaki(2021)Wang and Komatsuzaki]{wang2021gpt}
Ben Wang and Aran Komatsuzaki.
\newblock Gpt-j-6b: A 6 billion parameter autoregressive language model, 2021.

\bibitem[Wang et~al.(2024{\natexlab{a}})Wang, Cassano, Wu, Bai, Song, Nath, Han, Hendryx, Yue, and Zhang]{wang2024planningnaturallanguageimproves}
Evan Wang, Federico Cassano, Catherine Wu, Yunfeng Bai, Will Song, Vaskar Nath, Ziwen Han, Sean Hendryx, Summer Yue, and Hugh Zhang.
\newblock Planning in natural language improves llm search for code generation, 2024{\natexlab{a}}.
\newblock URL \url{https://arxiv.org/abs/2409.03733}.

\bibitem[Wang et~al.(2024{\natexlab{b}})Wang, Mehrabi, Goyal, Gupta, Chang, and Galstyan]{wang2024dataadvisor}
Fei Wang, Ninareh Mehrabi, Palash Goyal, Rahul Gupta, Kai-Wei Chang, and Aram Galstyan.
\newblock Data advisor: Dynamic data curation for safety alignment of large language models, 2024{\natexlab{b}}.
\newblock URL \url{https://arxiv.org/abs/2410.05269}.

\bibitem[Wang et~al.(2023{\natexlab{a}})Wang, Xie, Jiang, Mandlekar, Xiao, Zhu, Fan, and Anandkumar]{wang2023voyageropenendedembodiedagent}
Guanzhi Wang, Yuqi Xie, Yunfan Jiang, Ajay Mandlekar, Chaowei Xiao, Yuke Zhu, Linxi Fan, and Anima Anandkumar.
\newblock Voyager: An open-ended embodied agent with large language models, 2023{\natexlab{a}}.
\newblock URL \url{https://arxiv.org/abs/2305.16291}.

\bibitem[Wang et~al.(2024{\natexlab{c}})Wang, Zhang, Du, Zhang, and Chu]{wang2024surveydataselectionllm}
Jiahao Wang, Bolin Zhang, Qianlong Du, Jiajun Zhang, and Dianhui Chu.
\newblock A survey on data selection for llm instruction tuning, 2024{\natexlab{c}}.
\newblock URL \url{https://arxiv.org/abs/2402.05123}.

\bibitem[Wang et~al.(2024{\natexlab{d}})Wang, Li, Shao, Xu, Dai, Li, Chen, Wu, and Sui]{wang2024mathshepherdverifyreinforcellms}
Peiyi Wang, Lei Li, Zhihong Shao, R.~X. Xu, Damai Dai, Yifei Li, Deli Chen, Y.~Wu, and Zhifang Sui.
\newblock Math-shepherd: Verify and reinforce llms step-by-step without human annotations, 2024{\natexlab{d}}.
\newblock URL \url{https://arxiv.org/abs/2312.08935}.

\bibitem[Wang et~al.(2023{\natexlab{b}})Wang, Yu, Tan, O'Brien, Pasunuru, Dwivedi-Yu, Golovneva, Zettlemoyer, Fazel-Zarandi, and Celikyilmaz]{wang2023shepherdcriticlanguagemodel}
Tianlu Wang, Ping Yu, Xiaoqing~Ellen Tan, Sean O'Brien, Ramakanth Pasunuru, Jane Dwivedi-Yu, Olga Golovneva, Luke Zettlemoyer, Maryam Fazel-Zarandi, and Asli Celikyilmaz.
\newblock Shepherd: A critic for language model generation, 2023{\natexlab{b}}.
\newblock URL \url{https://arxiv.org/abs/2308.04592}.

\bibitem[Wang et~al.(2022)Wang, Mishra, Alipoormolabashi, Kordi, Mirzaei, Arunkumar, Ashok, Dhanasekaran, Naik, Stap, Pathak, Karamanolakis, Lai, Purohit, Mondal, Anderson, Kuznia, Doshi, Patel, Pal, Moradshahi, Parmar, Purohit, Varshney, Kaza, Verma, Puri, Karia, Sampat, Doshi, Mishra, Reddy, Patro, Dixit, Shen, Baral, Choi, Smith, Hajishirzi, and Khashabi]{wang2022supernaturalinstructionsgeneralizationdeclarativeinstructions}
Yizhong Wang, Swaroop Mishra, Pegah Alipoormolabashi, Yeganeh Kordi, Amirreza Mirzaei, Anjana Arunkumar, Arjun Ashok, Arut~Selvan Dhanasekaran, Atharva Naik, David Stap, Eshaan Pathak, Giannis Karamanolakis, Haizhi~Gary Lai, Ishan Purohit, Ishani Mondal, Jacob Anderson, Kirby Kuznia, Krima Doshi, Maitreya Patel, Kuntal~Kumar Pal, Mehrad Moradshahi, Mihir Parmar, Mirali Purohit, Neeraj Varshney, Phani~Rohitha Kaza, Pulkit Verma, Ravsehaj~Singh Puri, Rushang Karia, Shailaja~Keyur Sampat, Savan Doshi, Siddhartha Mishra, Sujan Reddy, Sumanta Patro, Tanay Dixit, Xudong Shen, Chitta Baral, Yejin Choi, Noah~A. Smith, Hannaneh Hajishirzi, and Daniel Khashabi.
\newblock Super-naturalinstructions: Generalization via declarative instructions on 1600+ nlp tasks, 2022.
\newblock URL \url{https://arxiv.org/abs/2204.07705}.

\bibitem[Wang et~al.(2023{\natexlab{c}})Wang, Kordi, Mishra, Liu, Smith, Khashabi, and Hajishirzi]{wang2023selfinstructaligninglanguagemodels}
Yizhong Wang, Yeganeh Kordi, Swaroop Mishra, Alisa Liu, Noah~A. Smith, Daniel Khashabi, and Hannaneh Hajishirzi.
\newblock Self-instruct: Aligning language models with self-generated instructions, 2023{\natexlab{c}}.
\newblock URL \url{https://arxiv.org/abs/2212.10560}.

\bibitem[Wei et~al.(2022)Wei, Bosma, Zhao, Guu, Yu, Lester, Du, Dai, and Le]{wei2022finetunedlanguagemodelszeroshot}
Jason Wei, Maarten Bosma, Vincent~Y. Zhao, Kelvin Guu, Adams~Wei Yu, Brian Lester, Nan Du, Andrew~M. Dai, and Quoc~V. Le.
\newblock Finetuned language models are zero-shot learners, 2022.
\newblock URL \url{https://arxiv.org/abs/2109.01652}.

\bibitem[Wei et~al.(2023)Wei, Wang, Schuurmans, Bosma, Ichter, Xia, Chi, Le, and Zhou]{wei2023chainofthoughtpromptingelicitsreasoning}
Jason Wei, Xuezhi Wang, Dale Schuurmans, Maarten Bosma, Brian Ichter, Fei Xia, Ed~Chi, Quoc Le, and Denny Zhou.
\newblock Chain-of-thought prompting elicits reasoning in large language models, 2023.
\newblock URL \url{https://arxiv.org/abs/2201.11903}.

\bibitem[Wei et~al.(2024)Wei, Huang, Lu, Zhou, and Le]{wei2024simplesyntheticdatareduces}
Jerry Wei, Da~Huang, Yifeng Lu, Denny Zhou, and Quoc~V. Le.
\newblock Simple synthetic data reduces sycophancy in large language models, 2024.
\newblock URL \url{https://arxiv.org/abs/2308.03958}.

\bibitem[Welleck et~al.(2020)Welleck, Kulikov, Roller, Dinan, Cho, and Weston]{Welleck2020Neural}
Sean Welleck, Ilia Kulikov, Stephen Roller, Emily Dinan, Kyunghyun Cho, and Jason Weston.
\newblock Neural text generation with unlikelihood training.
\newblock In \emph{International Conference on Learning Representations}, 2020.
\newblock URL \url{https://openreview.net/forum?id=SJeYe0NtvH}.

\bibitem[Wenzek et~al.(2019)Wenzek, Lachaux, Conneau, Chaudhary, Guzm{\'a}n, Joulin, and Grave]{wenzek2019ccnet}
Guillaume Wenzek, Marie-Anne Lachaux, Alexis Conneau, Vishrav Chaudhary, Francisco Guzm{\'a}n, Armand Joulin, and Edouard Grave.
\newblock Ccnet: Extracting high quality monolingual datasets from web crawl data.
\newblock \emph{arXiv preprint arXiv:1911.00359}, 2019.

\bibitem[Wenzek et~al.(2020)Wenzek, Lachaux, Conneau, Chaudhary, Guzm{\'a}n, Joulin, and Grave]{wenzek-etal-2020-ccnet}
Guillaume Wenzek, Marie-Anne Lachaux, Alexis Conneau, Vishrav Chaudhary, Francisco Guzm{\'a}n, Armand Joulin, and Edouard Grave.
\newblock {CCN}et: Extracting high quality monolingual datasets from web crawl data.
\newblock In Nicoletta Calzolari, Fr{\'e}d{\'e}ric B{\'e}chet, Philippe Blache, Khalid Choukri, Christopher Cieri, Thierry Declerck, Sara Goggi, Hitoshi Isahara, Bente Maegaard, Joseph Mariani, H{\'e}l{\`e}ne Mazo, Asuncion Moreno, Jan Odijk, and Stelios Piperidis (eds.), \emph{Proceedings of the Twelfth Language Resources and Evaluation Conference}, pp.\  4003--4012, Marseille, France, May 2020. European Language Resources Association.
\newblock ISBN 979-10-95546-34-4.
\newblock URL \url{https://aclanthology.org/2020.lrec-1.494}.

\bibitem[Wettig et~al.(2024)Wettig, Gupta, Malik, and Chen]{wettig2024quratingselectinghighqualitydata}
Alexander Wettig, Aatmik Gupta, Saumya Malik, and Danqi Chen.
\newblock Qurating: Selecting high-quality data for training language models, 2024.
\newblock URL \url{https://arxiv.org/abs/2402.09739}.

\bibitem[Williams(1992)]{Williams1992}
Ronald~J. Williams.
\newblock Simple statistical gradient-following algorithms for connectionist reinforcement learning.
\newblock \emph{Machine Learning}, 8\penalty0 (3):\penalty0 229--256, may 1992.
\newblock ISSN 1573-0565.
\newblock \doi{10.1007/BF00992696}.
\newblock URL \url{https://doi.org/10.1007/BF00992696}.

\bibitem[Wu et~al.(2024{\natexlab{a}})Wu, hao Wu, Wu, Feng, and Tan]{wu2024evolutionarycomputationeralarge}
Xingyu Wu, Sheng hao Wu, Jibin Wu, Liang Feng, and Kay~Chen Tan.
\newblock Evolutionary computation in the era of large language model: Survey and roadmap, 2024{\natexlab{a}}.
\newblock URL \url{https://arxiv.org/abs/2401.10034}.

\bibitem[Wu et~al.(2024{\natexlab{b}})Wu, Sun, Li, Welleck, and Yang]{wu2024inferencescalinglawsempirical}
Yangzhen Wu, Zhiqing Sun, Shanda Li, Sean Welleck, and Yiming Yang.
\newblock Inference scaling laws: An empirical analysis of compute-optimal inference for problem-solving with language models, 2024{\natexlab{b}}.
\newblock URL \url{https://arxiv.org/abs/2408.00724}.

\bibitem[Xie et~al.(2023)Xie, Santurkar, Ma, and Liang]{xie2023dataselectionlanguagemodels}
Sang~Michael Xie, Shibani Santurkar, Tengyu Ma, and Percy Liang.
\newblock Data selection for language models via importance resampling, 2023.
\newblock URL \url{https://arxiv.org/abs/2302.03169}.

\bibitem[Xie et~al.(2024)Xie, Pham, Dong, Du, Liu, Lu, Liang, Le, Ma, and Yu]{xie2024doremi}
Sang~Michael Xie, Hieu Pham, Xuanyi Dong, Nan Du, Hanxiao Liu, Yifeng Lu, Percy~S Liang, Quoc~V Le, Tengyu Ma, and Adams~Wei Yu.
\newblock Doremi: Optimizing data mixtures speeds up language model pretraining.
\newblock \emph{Advances in Neural Information Processing Systems}, 36, 2024.

\bibitem[Xu et~al.(2023)Xu, Sun, Zheng, Geng, Zhao, Feng, Tao, and Jiang]{xu2023wizardlmempoweringlargelanguage}
Can Xu, Qingfeng Sun, Kai Zheng, Xiubo Geng, Pu~Zhao, Jiazhan Feng, Chongyang Tao, and Daxin Jiang.
\newblock Wizardlm: Empowering large language models to follow complex instructions, 2023.
\newblock URL \url{https://arxiv.org/abs/2304.12244}.

\bibitem[Yang et~al.(2024{\natexlab{a}})Yang, Salamatian, Sun, Suresh, and Beirami]{yang2024asymptoticslanguagemodelalignment}
Joy~Qiping Yang, Salman Salamatian, Ziteng Sun, Ananda~Theertha Suresh, and Ahmad Beirami.
\newblock Asymptotics of language model alignment, 2024{\natexlab{a}}.
\newblock URL \url{https://arxiv.org/abs/2404.01730}.

\bibitem[Yang et~al.(2024{\natexlab{b}})Yang, Klein, Celikyilmaz, Peng, and Tian]{yang2024rlcdreinforcementlearningcontrastive}
Kevin Yang, Dan Klein, Asli Celikyilmaz, Nanyun Peng, and Yuandong Tian.
\newblock Rlcd: Reinforcement learning from contrastive distillation for language model alignment, 2024{\natexlab{b}}.
\newblock URL \url{https://arxiv.org/abs/2307.12950}.

\bibitem[Yao et~al.(2023)Yao, Yu, Zhao, Shafran, Griffiths, Cao, and Narasimhan]{yao2023treethoughtsdeliberateproblem}
Shunyu Yao, Dian Yu, Jeffrey Zhao, Izhak Shafran, Thomas~L. Griffiths, Yuan Cao, and Karthik Narasimhan.
\newblock Tree of thoughts: Deliberate problem solving with large language models, 2023.
\newblock URL \url{https://arxiv.org/abs/2305.10601}.

\bibitem[Ye et~al.(2022{\natexlab{a}})Ye, Gao, Feng, Wu, Yu, and Kong]{ye2022progenprogressivezeroshotdataset}
Jiacheng Ye, Jiahui Gao, Jiangtao Feng, Zhiyong Wu, Tao Yu, and Lingpeng Kong.
\newblock Progen: Progressive zero-shot dataset generation via in-context feedback, 2022{\natexlab{a}}.
\newblock URL \url{https://arxiv.org/abs/2210.12329}.

\bibitem[Ye et~al.(2022{\natexlab{b}})Ye, Gao, Li, Xu, Feng, Wu, Yu, and Kong]{ye2022zerogenefficientzeroshotlearning}
Jiacheng Ye, Jiahui Gao, Qintong Li, Hang Xu, Jiangtao Feng, Zhiyong Wu, Tao Yu, and Lingpeng Kong.
\newblock Zerogen: Efficient zero-shot learning via dataset generation, 2022{\natexlab{b}}.
\newblock URL \url{https://arxiv.org/abs/2202.07922}.

\bibitem[Ye et~al.(2024{\natexlab{a}})Ye, Liu, Sun, Zhou, Zhan, and Qiu]{ye2024datamixinglawsoptimizing}
Jiasheng Ye, Peiju Liu, Tianxiang Sun, Yunhua Zhou, Jun Zhan, and Xipeng Qiu.
\newblock Data mixing laws: Optimizing data mixtures by predicting language modeling performance, 2024{\natexlab{a}}.
\newblock URL \url{https://arxiv.org/abs/2403.16952}.

\bibitem[Ye et~al.(2024{\natexlab{b}})Ye, Greenlee-Scott, Bartolo, Blunsom, Campos, and Gallé]{ye2024improvingrewardmodelssynthetic}
Zihuiwen Ye, Fraser Greenlee-Scott, Max Bartolo, Phil Blunsom, Jon~Ander Campos, and Matthias Gallé.
\newblock Improving reward models with synthetic critiques, 2024{\natexlab{b}}.
\newblock URL \url{https://arxiv.org/abs/2405.20850}.

\bibitem[Yu et~al.(2024)Yu, Jiang, Shi, Yu, Liu, Zhang, Kwok, Li, Weller, and Liu]{yu2024metamathbootstrapmathematicalquestions}
Longhui Yu, Weisen Jiang, Han Shi, Jincheng Yu, Zhengying Liu, Yu~Zhang, James~T. Kwok, Zhenguo Li, Adrian Weller, and Weiyang Liu.
\newblock Metamath: Bootstrap your own mathematical questions for large language models, 2024.
\newblock URL \url{https://arxiv.org/abs/2309.12284}.

\bibitem[Yu et~al.(2023{\natexlab{a}})Yu, Xiao, Stone, Tompson, Brohan, Wang, Singh, Tan, Peralta, Ichter, et~al.]{yu2023scaling}
Tianhe Yu, Ted Xiao, Austin Stone, Jonathan Tompson, Anthony Brohan, Su~Wang, Jaspiar Singh, Clayton Tan, Jodilyn Peralta, Brian Ichter, et~al.
\newblock Scaling robot learning with semantically imagined experience.
\newblock \emph{arXiv preprint arXiv:2302.11550}, 2023{\natexlab{a}}.

\bibitem[Yu et~al.(2023{\natexlab{b}})Yu, Zhuang, Zhang, Meng, Ratner, Krishna, Shen, and Zhang]{yu2023largelanguagemodelattributed}
Yue Yu, Yuchen Zhuang, Jieyu Zhang, Yu~Meng, Alexander Ratner, Ranjay Krishna, Jiaming Shen, and Chao Zhang.
\newblock Large language model as attributed training data generator: A tale of diversity and bias, 2023{\natexlab{b}}.
\newblock URL \url{https://arxiv.org/abs/2306.15895}.

\bibitem[Yuan et~al.(2023)Yuan, Yuan, Li, Dong, Lu, Tan, Zhou, and Zhou]{yuan2023scalingrelationshiplearningmathematical}
Zheng Yuan, Hongyi Yuan, Chengpeng Li, Guanting Dong, Keming Lu, Chuanqi Tan, Chang Zhou, and Jingren Zhou.
\newblock Scaling relationship on learning mathematical reasoning with large language models, 2023.
\newblock URL \url{https://arxiv.org/abs/2308.01825}.

\bibitem[Yue et~al.(2023)Yue, Qu, Zhang, Fu, Huang, Sun, Su, and Chen]{yue2023mammothbuildingmathgeneralist}
Xiang Yue, Xingwei Qu, Ge~Zhang, Yao Fu, Wenhao Huang, Huan Sun, Yu~Su, and Wenhu Chen.
\newblock Mammoth: Building math generalist models through hybrid instruction tuning, 2023.
\newblock URL \url{https://arxiv.org/abs/2309.05653}.

\bibitem[Yue et~al.(2024)Yue, Zheng, Zhang, and Chen]{yue2024mammoth2scalinginstructionsweb}
Xiang Yue, Tuney Zheng, Ge~Zhang, and Wenhu Chen.
\newblock Mammoth2: Scaling instructions from the web, 2024.
\newblock URL \url{https://arxiv.org/abs/2405.03548}.

\bibitem[Zeng et~al.(2024)Zeng, Xu, Zhao, Lou, and Chen]{zeng2024automaticinstructionevolvinglarge}
Weihao Zeng, Can Xu, Yingxiu Zhao, Jian-Guang Lou, and Weizhu Chen.
\newblock Automatic instruction evolving for large language models, 2024.
\newblock URL \url{https://arxiv.org/abs/2406.00770}.

\bibitem[Zhang et~al.(2024{\natexlab{a}})Zhang, Wang, and Charton]{zhang2024instructiondiversitydrivesgeneralization}
Dylan Zhang, Justin Wang, and Francois Charton.
\newblock Instruction diversity drives generalization to unseen tasks, 2024{\natexlab{a}}.
\newblock URL \url{https://arxiv.org/abs/2402.10891}.

\bibitem[Zhang et~al.(2024{\natexlab{b}})Zhang, Wang, and Charton]{zhang2024textbfonlyif}
Dylan Zhang, Justin Wang, and Francois Charton.
\newblock $\textbf{Only-IF}$:revealing the decisive effect of instruction diversity on generalization, 2024{\natexlab{b}}.
\newblock URL \url{https://arxiv.org/abs/2410.04717}.

\bibitem[Zhang et~al.(2023)Zhang, Lehman, Stanley, and Clune]{zhang2023omni}
Jenny Zhang, Joel Lehman, Kenneth Stanley, and Jeff Clune.
\newblock Omni: Open-endedness via models of human notions of interestingness.
\newblock \emph{arXiv preprint arXiv:2306.01711}, 2023.

\bibitem[Zhang et~al.(2024{\natexlab{c}})Zhang, Hosseini, Bansal, Kazemi, Kumar, and Agarwal]{zhang2024generativeverifiersrewardmodeling}
Lunjun Zhang, Arian Hosseini, Hritik Bansal, Mehran Kazemi, Aviral Kumar, and Rishabh Agarwal.
\newblock Generative verifiers: Reward modeling as next-token prediction, 2024{\natexlab{c}}.
\newblock URL \url{https://arxiv.org/abs/2408.15240}.

\bibitem[Zhang et~al.(2024{\natexlab{d}})Zhang, Luo, Yuan, and Yao]{zhang2024autonomous}
Yifan Zhang, Yifan Luo, Yang Yuan, and Andrew~C Yao.
\newblock Autonomous data selection with language models for mathematical texts.
\newblock In \emph{ICLR 2024 Workshop on Navigating and Addressing Data Problems for Foundation Models}, 2024{\natexlab{d}}.
\newblock URL \url{https://openreview.net/forum?id=bBF077z8LF}.

\bibitem[Zhang et~al.(2024{\natexlab{e}})Zhang, Schwarzschild, Carlini, Kolter, and Ippolito]{zhang2024forcingdiffusedistributionslanguage}
Yiming Zhang, Avi Schwarzschild, Nicholas Carlini, Zico Kolter, and Daphne Ippolito.
\newblock Forcing diffuse distributions out of language models, 2024{\natexlab{e}}.
\newblock URL \url{https://arxiv.org/abs/2404.10859}.

\bibitem[Zhang et~al.(2024{\natexlab{f}})Zhang, Fontaine, Bhatt, Nikolaidis, and Li]{zhang2024arbitrarily}
Yulun Zhang, Matthew Fontaine, Varun Bhatt, Stefanos Nikolaidis, and Jiaoyang Li.
\newblock Arbitrarily scalable environment generators via neural cellular automata.
\newblock \emph{Advances in Neural Information Processing Systems}, 36, 2024{\natexlab{f}}.

\bibitem[Zhao et~al.(2024{\natexlab{a}})Zhao, Andrews, Papakyriakopoulos, and Xiang]{zhao2024positionmeasuredatasetdiversity}
Dora Zhao, Jerone T.~A. Andrews, Orestis Papakyriakopoulos, and Alice Xiang.
\newblock Position: Measure dataset diversity, don't just claim it, 2024{\natexlab{a}}.
\newblock URL \url{https://arxiv.org/abs/2407.08188}.

\bibitem[Zhao et~al.(2019)Zhao, Peyrard, Liu, Gao, Meyer, and Eger]{zhao-etal-2019-moverscore}
Wei Zhao, Maxime Peyrard, Fei Liu, Yang Gao, Christian~M. Meyer, and Steffen Eger.
\newblock {M}over{S}core: Text generation evaluating with contextualized embeddings and earth mover distance.
\newblock In Kentaro Inui, Jing Jiang, Vincent Ng, and Xiaojun Wan (eds.), \emph{Proceedings of the 2019 Conference on Empirical Methods in Natural Language Processing and the 9th International Joint Conference on Natural Language Processing (EMNLP-IJCNLP)}, pp.\  563--578, Hong Kong, China, November 2019. Association for Computational Linguistics.
\newblock \doi{10.18653/v1/D19-1053}.
\newblock URL \url{https://aclanthology.org/D19-1053}.

\bibitem[Zhao et~al.(2024{\natexlab{b}})Zhao, Yin, and Durrett]{zhao2024understandingsyntheticcontextextension}
Xinyu Zhao, Fangcong Yin, and Greg Durrett.
\newblock Understanding synthetic context extension via retrieval heads, 2024{\natexlab{b}}.
\newblock URL \url{https://arxiv.org/abs/2410.22316}.

\bibitem[Zhao et~al.(2024{\natexlab{c}})Zhao, Yu, Hui, Yu, Huang, Li, and Zhang]{zhao2024preliminarystudyintrinsicrelationship}
Yingxiu Zhao, Bowen Yu, Binyuan Hui, Haiyang Yu, Fei Huang, Yongbin Li, and Nevin~L. Zhang.
\newblock A preliminary study of the intrinsic relationship between complexity and alignment, 2024{\natexlab{c}}.
\newblock URL \url{https://arxiv.org/abs/2308.05696}.

\bibitem[Zheng et~al.(2022)Zheng, Han, and Polu]{zheng2022minif2fcrosssystembenchmarkformal}
Kunhao Zheng, Jesse~Michael Han, and Stanislas Polu.
\newblock Minif2f: a cross-system benchmark for formal olympiad-level mathematics, 2022.
\newblock URL \url{https://arxiv.org/abs/2109.00110}.

\bibitem[Zheng et~al.(2023)Zheng, Chiang, Sheng, Zhuang, Wu, Zhuang, Lin, Li, Li, Xing, Zhang, Gonzalez, and Stoica]{zheng2023judgingllmasajudgemtbenchchatbot}
Lianmin Zheng, Wei-Lin Chiang, Ying Sheng, Siyuan Zhuang, Zhanghao Wu, Yonghao Zhuang, Zi~Lin, Zhuohan Li, Dacheng Li, Eric~P. Xing, Hao Zhang, Joseph~E. Gonzalez, and Ion Stoica.
\newblock Judging llm-as-a-judge with mt-bench and chatbot arena, 2023.
\newblock URL \url{https://arxiv.org/abs/2306.05685}.

\bibitem[Zhong et~al.(2022)Zhong, Liu, Yin, Mao, Jiao, Liu, Zhu, Ji, and Han]{zhong2022towards}
Ming Zhong, Yang Liu, Da~Yin, Yuning Mao, Yizhu Jiao, Pengfei Liu, Chenguang Zhu, Heng Ji, and Jiawei Han.
\newblock Towards a unified multi-dimensional evaluator for text generation.
\newblock \emph{arXiv preprint arXiv:2210.07197}, 2022.

\bibitem[Zhou et~al.(2023{\natexlab{a}})Zhou, Liu, Xu, Iyer, Sun, Mao, Ma, Efrat, Yu, Yu, Zhang, Ghosh, Lewis, Zettlemoyer, and Levy]{zhou2023limaalignment}
Chunting Zhou, Pengfei Liu, Puxin Xu, Srini Iyer, Jiao Sun, Yuning Mao, Xuezhe Ma, Avia Efrat, Ping Yu, Lili Yu, Susan Zhang, Gargi Ghosh, Mike Lewis, Luke Zettlemoyer, and Omer Levy.
\newblock Lima: Less is more for alignment, 2023{\natexlab{a}}.
\newblock URL \url{https://arxiv.org/abs/2305.11206}.

\bibitem[Zhou et~al.(2023{\natexlab{b}})Zhou, Muresanu, Han, Paster, Pitis, Chan, and Ba]{zhou2023largelanguagemodelshumanlevel}
Yongchao Zhou, Andrei~Ioan Muresanu, Ziwen Han, Keiran Paster, Silviu Pitis, Harris Chan, and Jimmy Ba.
\newblock Large language models are human-level prompt engineers, 2023{\natexlab{b}}.
\newblock URL \url{https://arxiv.org/abs/2211.01910}.

\bibitem[Zhou et~al.(2024)Zhou, Keuper, and Fritz]{zhou2024balancingdiversityriskllm}
Yuxuan Zhou, Margret Keuper, and Mario Fritz.
\newblock Balancing diversity and risk in llm sampling: How to select your method and parameter for open-ended text generation, 2024.
\newblock URL \url{https://arxiv.org/abs/2408.13586}.

\bibitem[Zhu et~al.(2018)Zhu, Lu, Zheng, Guo, Zhang, Wang, and Yu]{zhu2018texygenbenchmarkingplatformtext}
Yaoming Zhu, Sidi Lu, Lei Zheng, Jiaxian Guo, Weinan Zhang, Jun Wang, and Yong Yu.
\newblock Texygen: A benchmarking platform for text generation models, 2018.
\newblock URL \url{https://arxiv.org/abs/1802.01886}.

\bibitem[Zhuang et~al.(2023)Zhuang, Liu, Ning, Huang, Lv, Huang, Zhao, Zhang, Mao, Wang, et~al.]{zhuang2023efficiently}
Yan Zhuang, Qi~Liu, Yuting Ning, Weizhe Huang, Rui Lv, Zhenya Huang, Guanhao Zhao, Zheng Zhang, Qingyang Mao, Shijin Wang, et~al.
\newblock Efficiently measuring the cognitive ability of llms: An adaptive testing perspective.
\newblock \emph{arXiv preprint arXiv:2306.10512}, 2023.

\bibitem[Ziegler et~al.(2020)Ziegler, Stiennon, Wu, Brown, Radford, Amodei, Christiano, and Irving]{ziegler2020finetuninglanguagemodelshuman}
Daniel~M. Ziegler, Nisan Stiennon, Jeffrey Wu, Tom~B. Brown, Alec Radford, Dario Amodei, Paul Christiano, and Geoffrey Irving.
\newblock Fine-tuning language models from human preferences, 2020.
\newblock URL \url{https://arxiv.org/abs/1909.08593}.

\end{thebibliography}
